\documentclass[10pt]{article} 
\usepackage[accepted]{rlc}

\usepackage{amssymb}            
\usepackage{mathtools}          
\usepackage{mathrsfs}           
\usepackage{graphicx}           
\usepackage{subcaption}         
\usepackage[space]{grffile}     
\usepackage{url}                

\usepackage{amsmath,amsfonts,bm}

\def\eqref#1{equation~\ref{#1}}

\def\1{\bm{1}}

\def\va{{\bm{a}}}
\def\vb{{\bm{b}}}

\def\vw{{\bm{w}}}
\def\vx{{\bm{x}}}
\def\vy{{\bm{y}}}

\def\mW{{\bm{W}}}

\DeclareMathAlphabet{\mathsfit}{\encodingdefault}{\sfdefault}{m}{sl}
\SetMathAlphabet{\mathsfit}{bold}{\encodingdefault}{\sfdefault}{bx}{n}

\usepackage{algorithm}
\usepackage[noend]{algpseudocode}
\newcommand{\hangin}{\goodbreak\hangindent=.35cm \noindent}
\algrenewcommand\algorithmicindent{0.7em}
\usepackage{xfrac}
\usepackage{wrapfig}
\usepackage{multirow}
\usepackage{amsthm}
\newtheorem{theorem}{Theorem}

\newtheorem{corollary}{Corollary}
\usepackage{xcolor}
\usepackage{lipsum}  

\title{Weight Clipping for Deep Continual and\\Reinforcement Learning}

\author{
\vspace{-0.4cm}\\
\makebox[.24\linewidth]{\begin{tabular}{c}
        \textbf{Mohamed Elsayed} \\
        University of Alberta \\
        mohamedelsayed@ualberta.ca\\
        \vphantom{} 
    \end{tabular}}
    \hspace{0.15cm}
    \makebox[.24\linewidth]{\begin{tabular}{c}
        \textbf{Qingfeng Lan} \\
        University of Alberta \\
        qlan3@ualberta.ca\\
        \vphantom{} 
    \end{tabular}}
    \hspace{-0.2cm}
    \makebox[.24\linewidth]{\begin{tabular}{c}
        \textbf{Clare Lyle} \\
        Google DeepMind \\
        clarelyle@google.com\\
        \vphantom{} 
    \end{tabular}}
    \hspace{0cm}
    \makebox[.24\linewidth]{\begin{tabular}{c}
        \textbf{A.\ Rupam Mahmood} \\
        University of Alberta \\
        CIFAR AI Chair \\
        armahmood@ualberta.ca
    \end{tabular}}
    }

\begin{document}

\maketitle
\vspace{-0.1cm}
\begin{abstract}
Many failures in deep continual and reinforcement learning are associated with increasing magnitudes of the weights, making them hard to change and potentially causing overfitting.
While many methods address these learning failures, they often change the optimizer or the architecture, a complexity that hinders widespread adoption in various systems.
In this paper, we focus on learning failures that are associated with increasing weight norm and we propose a simple technique that can be easily added on top of existing learning systems: clipping neural network weights to limit them to a specific range.
We study the effectiveness of weight clipping in a series of supervised and reinforcement learning experiments.
Our empirical results highlight the benefits of weight clipping for generalization, addressing loss of plasticity and policy collapse, and facilitating learning with a large replay ratio.
\footnote{Code is available at \url{https://github.com/mohmdelsayed/weight-clipping}}
\end{abstract}

\section{Introduction}
\label{sec:introduction}
\begin{wrapfigure}{r}{0.345\textwidth}
    \centering
    \vspace{-0.4cm}
    \includegraphics[width=0.345\textwidth]{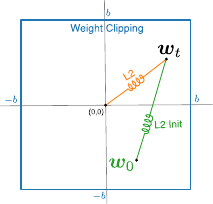}
    \caption{Weight clipping confines the weights in restricted space while \emph{L2 Init} pulls the current weight $\vw_t$ to the weight at initialization $\vw_0$ and \emph{L2} pulls the current weight $\vw_t$ to the zero vector.}
    \label{fig:diagram}
\vspace{-0.4cm}
\end{wrapfigure}
Deep learning and reinforcement learning methods face many challenges when learning online or continually. These challenges include loss of plasticity (Lyle et al.\ 2023, Dohare et al.\ 2023a), failure to achieve further improvement (e.g., Sokar et al.\ 2023, Lyle et al.\ 2023, 2021), gradual performance decreases (e.g., Dohare et al.\ 2023a, Abbas et al.\ 2023, Elsayed \& Mahmood 2024), and even loss of generalization (Ash \& Adams 2019).
One common feature among many instances of learning failures is their association with increasing weight magnitudes. 
An unbounded weight growth can make it increasingly harder for the learners to adjust the weight further (Lyle et al.\ 2024), causing loss of plasticity (Dohare et al.\ 2023a) or policy collapse (Dohare et al.\ 2023b). 
Moreover, large weight magnitudes can be harmful to the optimization dynamics (see Lyle et al.\ 2023 and Wortsman et al.\ 2023) and are often associated with overfitting (Zhang et al.\ 2021), leading to performance decrease and potentially explaining the learning difficulties.

While many methods exist to address these learning difficulties, these methods are predominantly complex or require significant changes to the learning systems. For example, several methods require a change of the optimization method (Dohare et al.\ 2023a, Elsayed \& Mahmood 2024, Sokar et al.\ 2023), use an auxiliary objective (Lan et al.\ 2023), change the architecture (e.g., Lyle et al.\ 2023, Nikishin et al.\ 2023, Lan \& Mahmood 2023), or even require a new reinforcement learning estimator that maintains exploration (Garg et al.\ 2022).
There are simpler techniques of weight regularization that directly promote small weight magnitudes, such as weight decay (\emph{L2})  (Krogh \& Hertz 1991, Ash \& Adams 2019, Lyle et al.\ 2024) and more recently, regularization toward initial weights (\emph{L2 Init}) (Kumar et al.\ 2023a). 
In Fig.\ \ref{fig:diagram}, we illustrate the difference between L2 and L2 Init, showing that L2 results in biasing the weights towards the zero weights while L2 Init results in biasing the weights towards the initial weights. 
The issue of biasing toward a particular point is that it applies to all weights regardless of their usefulness, resulting in overwriting and re-learning of previously useful weights, a primary cause of catastrophic forgetting (McCloskey \& Cohen 1989, French 1999, Elsayed \& Mahmood 2024). 
Lewandowski et al.\ (2024) proposed to use the empirical Wasserstein distance instead of L2 to alleviate the issue of L2 Init. However, the Wasserstein distance requires sorting the parameters, which is computationally expensive. It is desirable to address the issue of increasing weight magnitude in a computationally cheap manner without biasing the weight vector towards any point, allowing the learner to continually build on previously useful weights and overcome continual learning difficulties.

We use a simple remedy that can reduce the norm of the weights by clipping any large weight magnitude. This technique is distinct from gradient or update clipping (e.g., Brock et al.\ 2021, Badia et al.\ 2020), which aims to clip large updates or gradients, not weights. Constraining the weights prevents large weight magnitudes without biasing them toward a certain point, as depicted in Fig.\ \ref{fig:diagram}. In this paper, we study the role of weight clipping in 1) improving generalization, 2) addressing loss of plasticity and policy collapse, and 3) facilitating learning with a large replay ratio.


\section{Problem Formulation}
\label{sec:problem-formulation}
In this section, we describe the two problem formulations we use in this paper and what metrics we use to evaluate learners in each of them.

\subsection{Streaming Supervised Learning}
\label{sec:streaming-learning}
In streaming supervised learning, the data samples are presented to the learner as they come, one sample at each time step. Each sample is processed \emph{once} by the agent, then the learner is evaluated based on some evaluation metric, and after that, the sample is discarded immediately (Hayes et al.\ 2019). This setup mirrors animal learning (Hayes \& Kanan 2022) and is important for various applications such as on-device learning. The target function $f_t$ generating these data samples is typically non-stationary, producing non-independently and identically distributed (non-i.i.d.) samples such that $\vy_t=f_t(\vx_t)$. For simplicity, we assume that the target function is locally stationary in time, not arbitrarily non-stationary, but instead changes frequently (e.g., due to task change). The learner is expected to process the input $\vx_t \in\mathbb{R}^d$ and produce a prediction $\hat{\vy}_t\in\mathbb{R}^n$, after which it is evaluated based on the metric $E(\vy_t, \hat{\vy}_t)$. The goal of the learner is to maximize the average online metric (see Kumar et al.\ 2023b, Elsayed \& Mahmood 2024) given by $\frac{1}{T} \sum_{t=1}^T E(\vy_t, \hat{\vy}_t)$, where $T$ is the total number of time steps.

\subsection{Reinforcement Learning} 
\label{sec:reinforcement-learning}
The sequential decision process of the agent and the interaction with the environment is modeled as a Markov decision process (MDP). In this paper, we consider episodic interactions between the agent and the environment in which its episodic MDP is denoted by the tuple $(\mathcal{S}, \mathcal{A}, \mathcal{P}, \mathcal{R}, \gamma, d_0, \mathcal{H})$, where $\mathcal{S}$ is the set of states, $\mathcal{A}$ is the set of actions, $\mathcal{R} \subset \mathbb{R} $ denotes the set of reward signals, $\mathcal{P}:\mathcal{S}\times\mathcal{A}\rightarrow \Delta(\mathcal{S}\times \mathcal{R})$ denotes the transition dynamics model in which $\Delta(X)$ is a distribution over the set $X$, $d_0$ is the starting state distribution, $\gamma\in[0,1]$ is the discount factor, and $\mathcal{H}$ is the set of terminal states. The agent interacts with the environments using a policy $\pi: \mathcal{S}\rightarrow\Delta(\mathcal{A})$ that outputs a distribution over actions conditioned on the state (Sutton \& Barto 2018). Each episode of interaction begins after the environment samples a state from the starting state distribution $S_0\sim d_0$. At each time step $t$, the policy receives the state $S_t$ and produces the action $A_t\sim \pi(.|S_t)$, and then the environment samples a new state and reward signal using the transition dynamics as follows: $S_{t+1}, R_{t+1}\sim p(.,.|S_{t}, A_{t})$. The interaction continues until the agent ends up in one of the terminal states $S_T \sim \mathcal{H}$, where $T$ is the termination time step. The amount of the discounted sum of rewards collected by the agent during the episode at time step $t$ is known as the episodic return and is given by $G_t\doteq\sum_{k=t+1}^T \gamma^{k-t-1}R_{k}$. The goal of the agent is to maximize the expected return $\mathbb{E}_\pi[G_t]$ produced by following the policy $\pi$.


\section{Method}
\label{sec:method}
In this section, we introduce weight clipping and show how it can be used with existing optimization methods. We propose a clipping scheme that uses the boundaries given by a uniform distribution at initialization (e.g., He et al.\ 2015). Consider a neural network $f$ parameterized by the set of weights $\mathcal{W}=\{\mW_1,\mW_2,\dots,\mW_L\}$ and the set of biases $\{ \vb_1, \vb_2,\dots,\vb_L\}$. Specifically, given that the entries of the weight matrix $\mW_l, \forall l$ and the bias vector $\vb_l,\forall l$ are initialized from the uniform distribution $U[-s_l, s_l], s_l\in \mathbb{R}^+,\forall l$, we propose clipping any value outside the range $[-\kappa s_l, \kappa s_l]$, where $\kappa\in\mathbb{R}^+$ is a hyper-parameter defining the size of the constraint weight space. Algorithm \ref{alg:wc} shows how weight clipping can be integrated into optimization methods. 

\begin{algorithm}[ht]
\caption{Weight-constrained Stochastic Optimization}\label{alg:wc}
\begin{algorithmic}
\State Given a stream of data $\mathcal{D}$, a neural network $f$ with weights $\{\mW_1,...,\mW_L\}$ and biases $\{\vb_1,\dots,\vb_L\}$
\State Set step size $\alpha$ and initialization bounds $\{s_l\}_{l=1}^L$ (e.g., $s_l=1/\sqrt{n_l})$, where $n_l$ is \texttt{fan\_in}
\State Set clipping parameter $\kappa$ (e.g., $\kappa=2$)
\State Initialize weights $W_{l,i,j} \sim U[-s_l, s_l], \forall i,j,l$
\State
\For{$S$ in $\mathcal{D}$}
\For{$l$ in $\{L,L-1,...,1\}$}
\State $\mW_l \leftarrow \mW_l - \alpha \Delta \mW_l$ \Comment{Weight Update (e.g., $\Delta \mW_l=\nabla_{\mW_l} \mathcal{L}(S)$ in SGD)}
\State $\vb_l \leftarrow \vb_l - \alpha \Delta \vb_l$ \Comment{Bias Update (e.g., $\Delta \vb_l=\nabla_{\vb_l} \mathcal{L}(S)$ in SGD)}
\State $\mW_l \leftarrow \text{Clip}(\mW_l, \text{min}=-\kappa s_l, \text{max}=\kappa s_l)$
\State $\vb_l \leftarrow \text{Clip}(\vb_l, \text{min}=-\kappa s_l, \text{max}=\kappa s_l)$
\EndFor
\EndFor
\end{algorithmic}
\end{algorithm}

One natural question is whether we reduce the expressivity of the network by weight clipping. Typically, the larger the neural network, the smaller the weight change is to reduce the loss (Ghorbani et al.\ 2019, Geiger et al.\ 2020). For example, in over-parameterized networks, the weight change is usually tiny to represent any function, which is referred to as lazy training (see Chizat et al.\ 2019). Thus, we suspect weight clipping would not reduce much of the expressivity of over-parameterized networks due to tiny weight changes and a relatively large clipping range. We leave studying the effect of weight clipping on neural network expressivity to future work.

Next, we investigate the effect of weight clipping on smoothing the neural network. Liu et al.\ (2022) showed that reducing the Lipschitz constant (e.g., via Lipschitz regularization) increases the smoothness of the functions represented by neural networks and, hence, improves generalization (Yoshida et al.\ 2017), stabilizes Wasserstein generative adversarial networks (Arjovsky et al.\ 2017), and protects against adversarial attacks (Li et al.\ 2019). We start our analysis by showing that weight clipping leads to bounding the Lipschitz constant. In Theorem \ref{thm:smoothness}, we show that weight clipping bounds the Lipschitz constant (see proof in Appendix \ref{appendix:proofs}). For simplicity, we consider fully connected networks and $1$-Lipschitz activation functions (e.g., ReLU, Leaky ReLU, Tanh), but the results can be extended to other networks (see Gouk et al.\ 2021) and other activation functions. 
\begin{theorem}
\label{thm:smoothness}
\textbf{Smoothness of Clipped Networks.} Consider a fully-connected neural network $f_{\mathcal{W}}: \mathcal{X} \rightarrow \mathcal{Y}$ parametrized by the set of augmented weight matrices (include biases) $\mathcal{W}_{\text{Aug}}=\{\mW_1,\dots,\mW_L \}$. If the activation function $\sigma$ used is $1$-Lipschitz (e.g., ReLU), then the clipped network $f_{\mathcal{W}}^\text{Clipped}$ is Lipschitz continuous. That is, $\exists k\geq 0$ such that $\| f^\text{Clipped}_{\mathcal{W}}(\vx_1) - f^\text{Clipped}_{\mathcal{W}}(\vx_2) \|_1 \leq k \| \vx_1 -\vx_2 \|_1, \forall \vx_1,\vx_2\in \mathcal{X}$.
\end{theorem}

This result suggests that weight clipping adds an upper bound to the sharpness level of the network. In contrast, while other regularization methods can also improve smoothness (e.g., Liu et al.\ 2022), they do not have any guarantees on the bounds of the Lipschitz constant, and, hence, they can sometimes still reach sharp solutions.

Finally, we show that weight clipping makes the function change bounded. Update boundedness is a desired property of reinforcement learning methods, especially for the on-policy ones such as PPO (Schulman et al.\ 2017), to prevent the policy from changing too much from its old policy, achieving more learning stability. In corollary \ref{corollary:boundedness}, we show how weight clipping leads to update boundedness using the fact that clipped networks are Lipschitz continuous (see proof in Appendix \ref{appendix:proofs}).

\begin{corollary}
\label{corollary:boundedness}
\textbf{Update Boundedness of Clipped Networks.} Consider a clipped fully-connected neural network $f_\mathcal{W}^\text{Clipped}$ parameterized by the set of augmented weight matrices (include biases) $\mathcal{W}_{\text{Aug}}=\{\mW_1,\dots,\mW_L \}$. If the activation function $\sigma$ used is $1$-Lipschitz (e.g., ReLU), then any function update $\|\Delta f_\mathcal{W}^\text{Clipped}\|_1$ is bounded.
\end{corollary}


\section{Experiments}
\label{sec:experiments}
In this section, we study the effectiveness of weight clipping in 1) improving generalization, 2) maintaining network plasticity, 3) mitigating policy collapse, and 4) facilitating learning with large replay ratios. We start by considering the warm start setup (see Ash \& Adams 2019) and show that weight clipping can reduce loss of generalization. We then evaluate weight clipping using non-stationary streaming learning problems introduced by Elsayed and Mahmood (2024). Specifically, we use the Input-Permuted MNIST, Label-Permuted EMNIST, and Label-Permuted mini-ImageNet problems. Finally, we evaluate the effectiveness of weight clipping in addressing policy collapse in PPO (Schulman et al.\ 2017) and improving the performance of DQN (Mnih et al.\ 2015) and Rainbow (Hessel et al.\ 2018) with a large replay ratio.

The performance is evaluated based on the test accuracy in the warm-starting problem, the average online accuracy for the streaming learning problems, and the non-discounted episodic return for the reinforcement learning problems. We perform a hyper-parameter search (see Appendix \ref{appendix:hp-search}) for each method and use the best-performing configuration to plot in the following experiments. Our criterion for the best-performing configuration is the one that maximizes the area under the accuracy curve in supervised learning problems and the area under the non-discounted episodic return for the reinforcement learning problems.

\subsection{Weight Clipping for Improved Generalization}
\begin{wrapfigure}{r}{0.4\textwidth}
    \vspace{-0.6cm}
    \centering
    \includegraphics[width=0.4\textwidth]{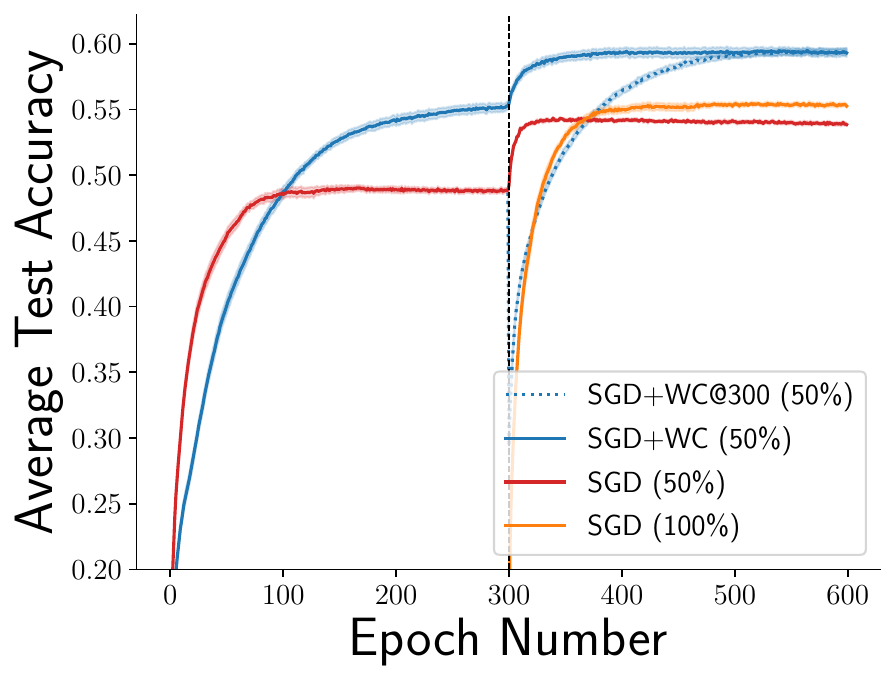}
    \caption{Performance when training on the data sequentially against when data is aggregated. The shaded region represents the standard error.}
    \label{fig:warm-starting}
    \vspace{-0.3cm}
\end{wrapfigure}
Here, we study the role of weight clipping (WC) in improving generalization. We use the warm starting setup proposed by Ash and Adams (2019) with the CIFAR-10 dataset (Krizhevsky 2009). Two variations of SGD are trained from scratch using ResNet18 (He et al.\ 2016), one using $100\%$ of the training and another in two stages: on $50\%$ of the data, followed by the other $50\%$. In Fig.\ \ref{fig:warm-starting}, when SGD is trained in two stages, we observe that its test accuracy is lower than if it was trained on the full data in one stage, which gives the same loss of generalization phenomenon demonstrated by Ash and Adams (2019). We introduce weight clipping in two ways: 1) clip once after the training on the first half of the data and 2) clip every time step. We observe that weight clipping only once removes the generalization gap and improves the test accuracy significantly. On the other hand, weight clipping every time step improves generalization in both stages of learning. These results show that large weights can cause overfitting and reduction in performance, which is alleviated by weight clipping that can improve generalization. The results are averaged over $10$ independent runs.

\subsection{Weight Clipping in Streaming Learning}
Now, we study weight clipping in addressing loss of plasticity using the Input-permuted MNIST problem. The input-permuted MNIST problem is a standard problem for studying loss of plasticity in neural networks since the learned features become irrelevant to the next task after each permutation. We permute the inputs every $5000$ time step. A new task starts when a permutation is performed. Next, we use the label-permuted EMNIST and label-permuted mini-ImageNet problems where loss of plasticity is intertwined with catastrophic forgetting (Elsayed \& Mahmood 2024). In both label-permuted problems, we permuted the labels each $2500$ time step.

We compare SGD and Adam (Kingma \& Ba 2014) along with their variations with two regularization methods, Shirnk \& Perturb (S\&P) and L2 Init, against weight clipping with SGD and Adam. In previous works (Kumar et al.\ 2023a, Dohare et al.\ 2023a), L2-Init and S\&P have been shown to be effective in maintaining plasticity when combined with SGD but have not been studied when combined with Adam. In addition, we compare against the Madam optimizer (Bernstein et al.\ 2020), which uses weight clipping to reduce its exponential weight growth but has not been studied in non-stationary settings before. In the streaming classification problems, the learner is required to maximize the online average accuracy using a stream of data, one sample at a time. Each learner is presented with $1$M samples and uses a multi-layer ($300\times150$) network with \emph{leaky-relu} units.

\begin{figure}[t]
    \centering
    \begin{subfigure}[b]{0.32\textwidth}
        \centering
    \includegraphics[width=\textwidth]{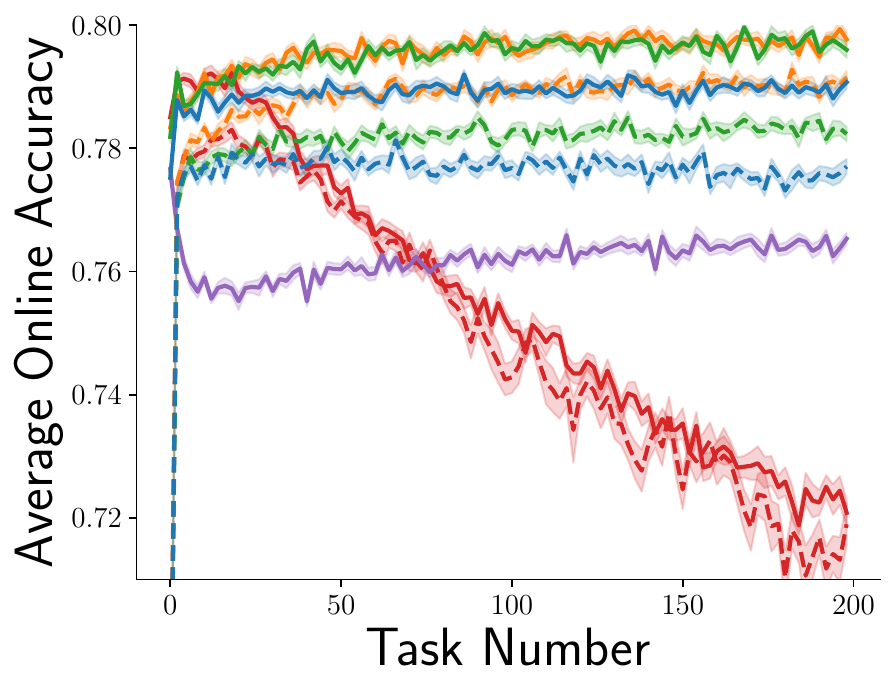}
            \caption{I/P MNIST}
        \label{fig:ip-mnist}
    \end{subfigure}
    \hfill
    \begin{subfigure}[b]{0.32\textwidth}
        \centering
    \includegraphics[width=\textwidth]{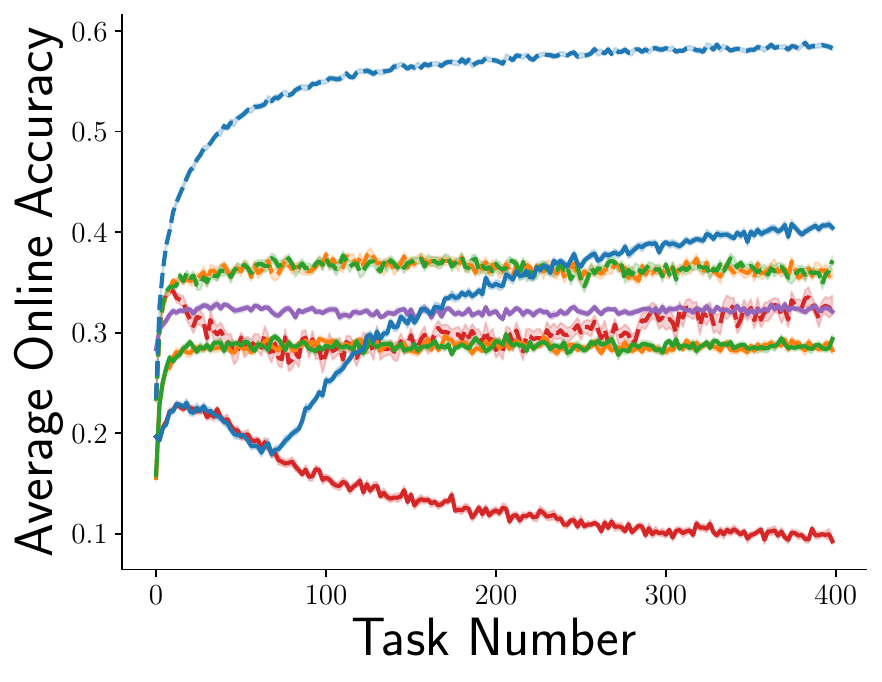}
            \caption{L/P EMNIST}
        \label{fig:lp-emnist}
    \end{subfigure}
    \hfill
    \begin{subfigure}[b]{0.32\textwidth}
        \centering
    \includegraphics[width=\textwidth]{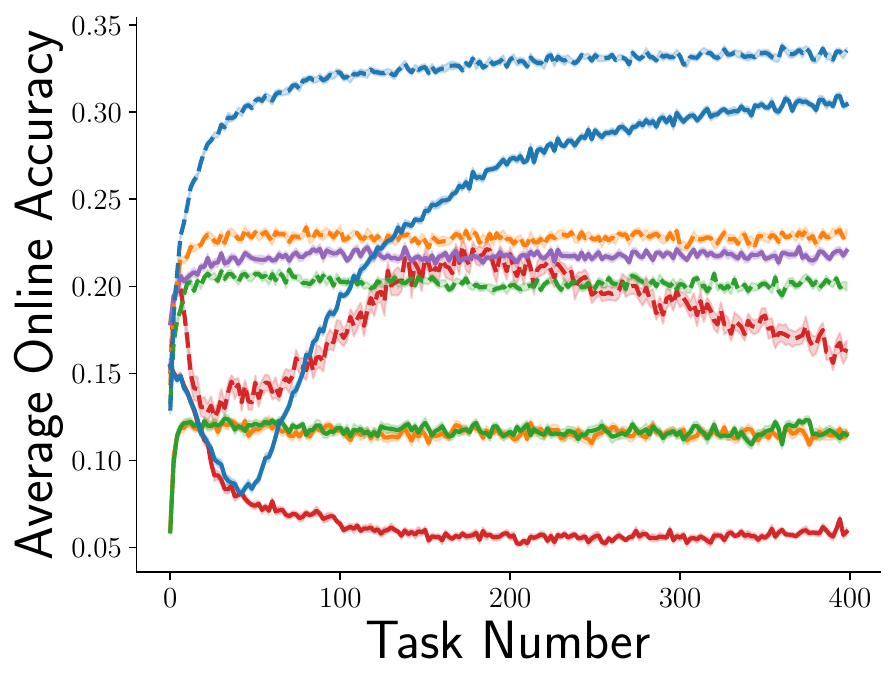}
            \caption{L/P \emph{mini}-ImageNet}
        \label{fig:lp-imagenet}
    \end{subfigure}
    \includegraphics[width=\textwidth]{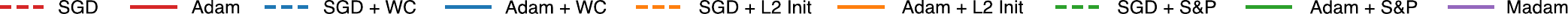}
    \caption{Performance of Adam and SGD with Weight Clipping on Input-permuted MNIST, Label-Permuted EMNIST, and Label-Permuted \emph{mini}-ImageNet. All curves are averaged over $20$ independent runs. The shaded area represents the standard error
    }
    \label{fig:sl-experiments}
\end{figure}

Fig.\ \ref{fig:ip-mnist} shows that almost all methods except for SGD and Adam maintain their performance throughout, but by different degrees. We plot the average online accuracy against the number of tasks, where each point in the figure represents the percentage of correct predictions within the task since the sample online prediction is $1$ for correct predictions and $0$ for incorrect ones.

\begin{figure}[ht]
    \centering
    \includegraphics[width=0.24\textwidth]{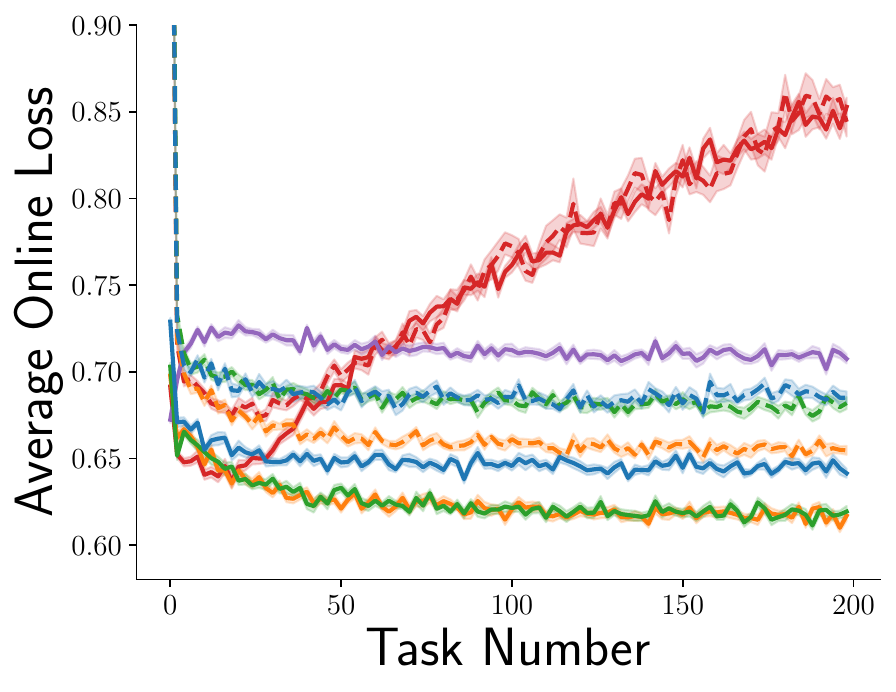}
    \includegraphics[width=0.244\textwidth]{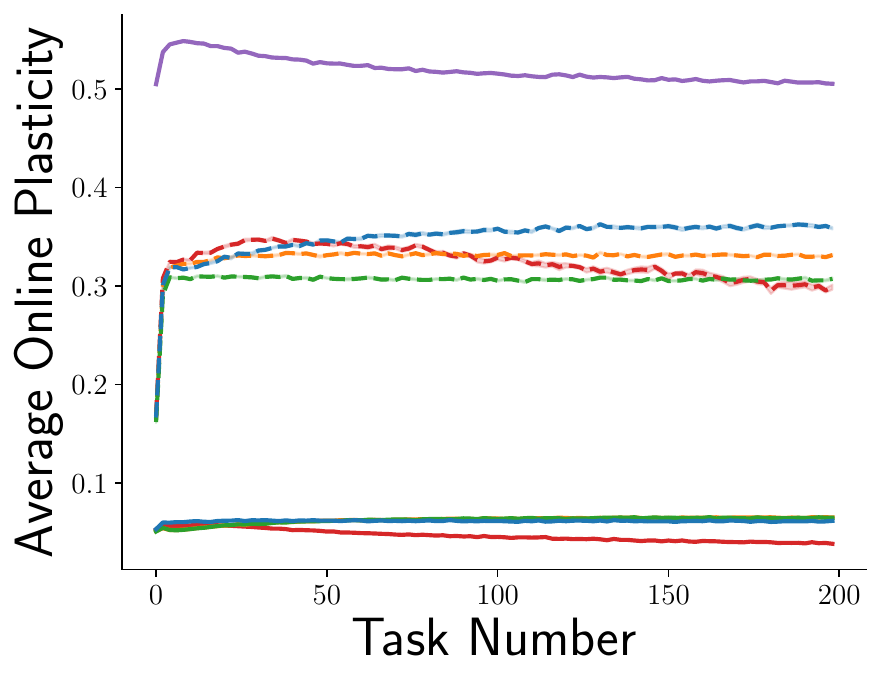}
    \includegraphics[width=0.244\textwidth]{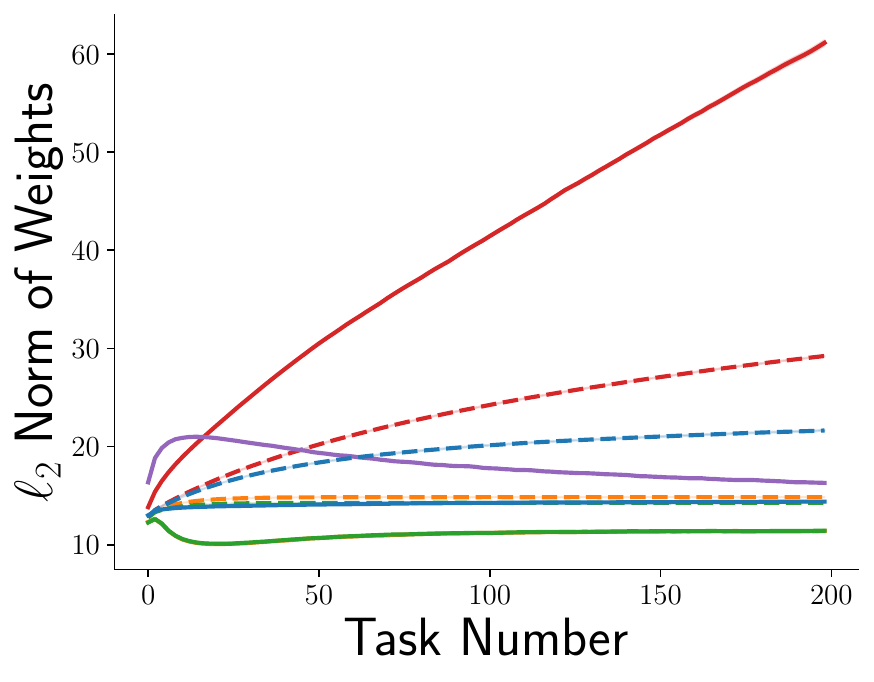}
    \includegraphics[width=0.244\textwidth]{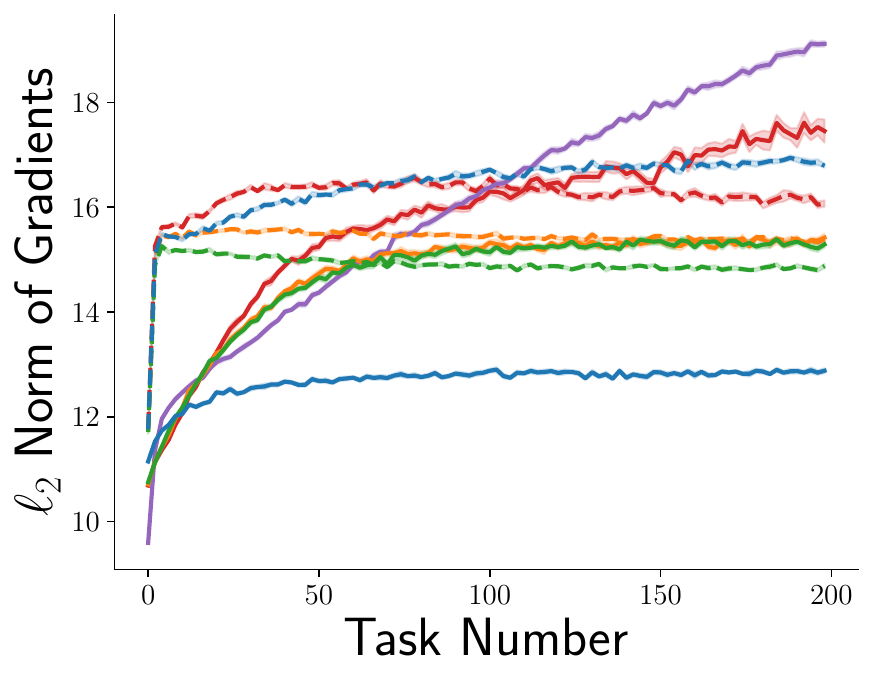}
    \includegraphics[width=\textwidth]{figures/SL/wide-legend.pdf}
    \caption{Diagnostic Statistics of different methods in Input-permuted MNIST. We show the online loss, the online plasticity, the $\ell_2$-norm of gradients, and the $\ell_2$-norm of weights.}
    \label{fig:input-permuted-mnist-stats}
    \vspace{-0.25cm}
\end{figure}

In Fig.\ \ref{fig:input-permuted-mnist-stats}, we characterize the solutions of each method using diagnostic statistics. Specifically, we show the online loss, $\ell_2$-norm of weights, and $\ell_2$-norm of gradients. In addition, we show the average online plasticity of each method using the sample plasticity metric (Elsayed \& Mahmood 2024), which is given by $p(Z)= \text{max}\left(1-\frac{\mathcal{L}(\mathcal{W}^\dagger, Z)}{\max(\mathcal{L}(\mathcal{W}, Z), \epsilon)},0\right)\in [0,1]$, where $Z$ is the sample, $\mathcal{W}^\dagger$ is the set of weights after performing the update and $\epsilon$ is a small number for numerical stability. We observe a gradual increase of the $\ell_2$ norm of the weights of SGD and Adam compared to other methods. 

Next, we use the label-permuted problems to evaluate the role of weight clipping in not biasing the weights towards some point. The label-permuted problems involve label permutation, which means the learned representation by the learner does not need to change, and the learner can continually improve upon them instead of overwriting and then re-learning. Fig.\ \ref{fig:lp-emnist} and \ref{fig:lp-imagenet} show that while all methods addressing loss of plasticity can maintain their performance, only weight clipping can keep improving its performance, likely due to not biasing the weights toward a specific point. We defer the diagnostic statistics on these two problems to Appendix \ref{appendix:experimental-details}.

\subsection{Weight Clipping Against Policy Collapse}
\begin{figure}[ht]
    \centering
    \includegraphics[width=0.244\textwidth]{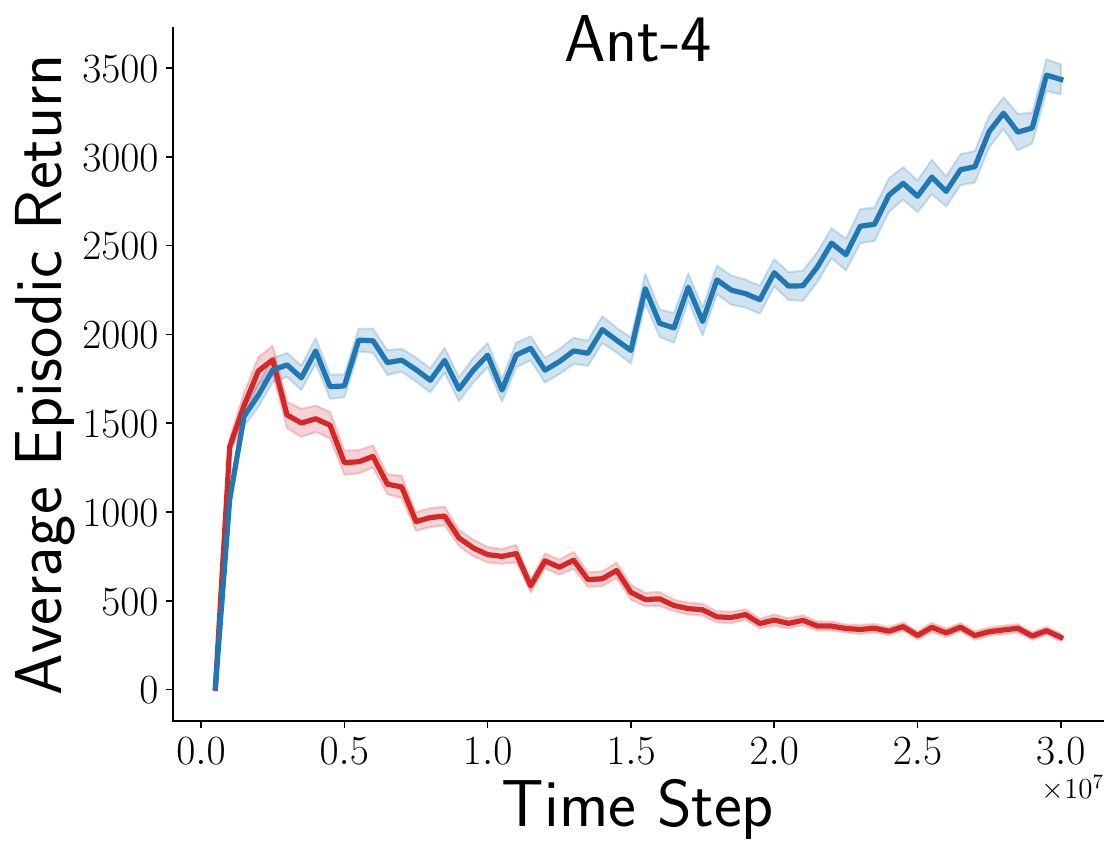}
    \includegraphics[width=0.244\textwidth]{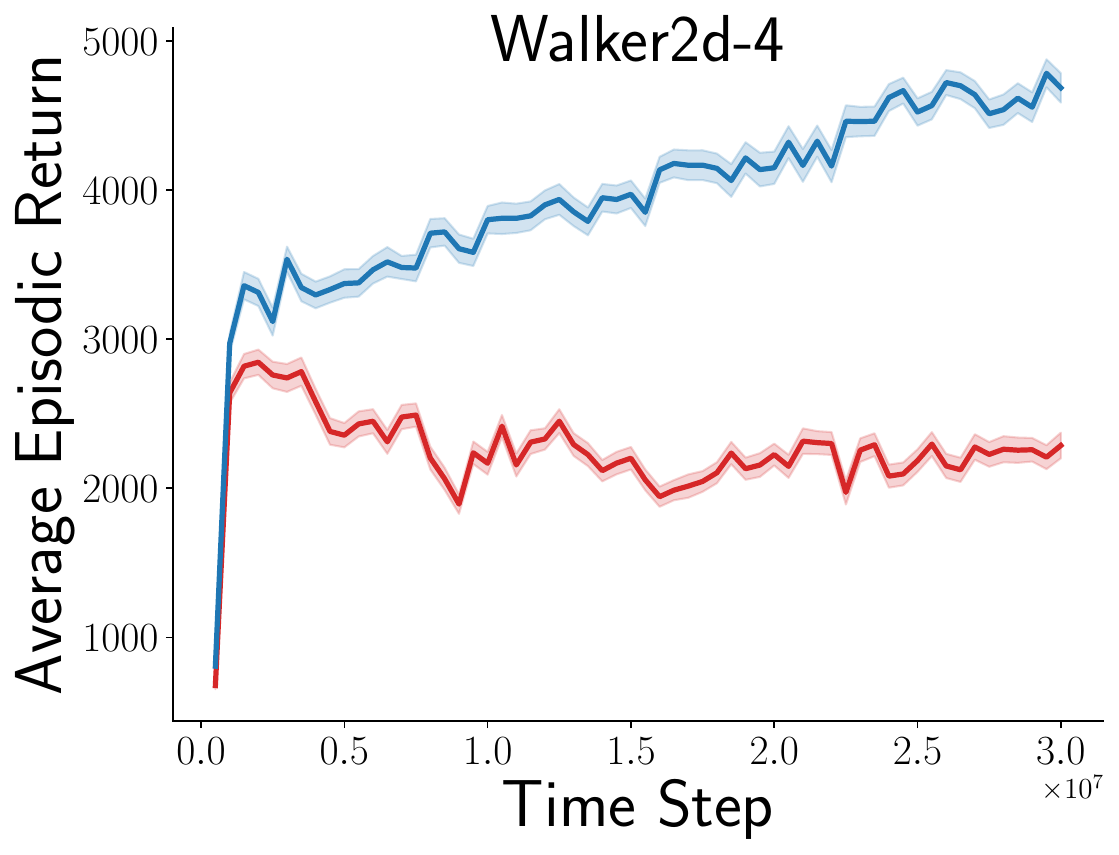}
    \includegraphics[width=0.244\textwidth]{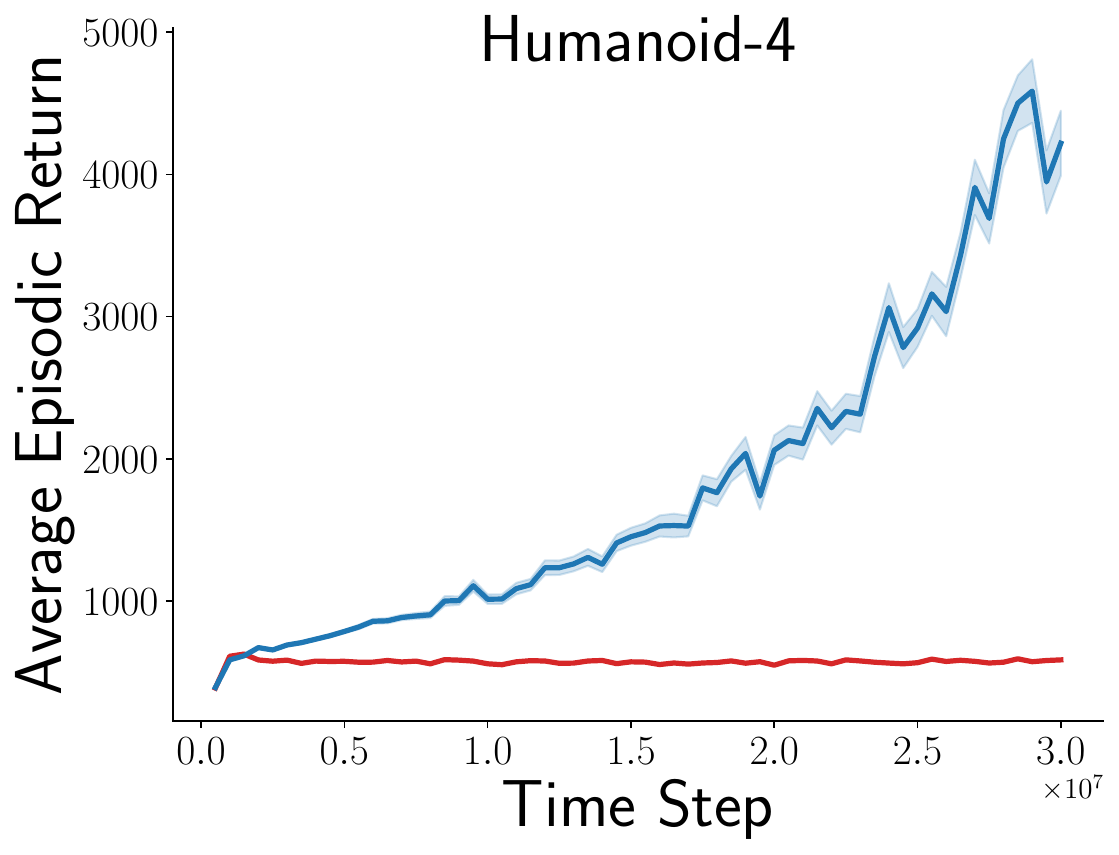}
    \includegraphics[width=0.244\textwidth]{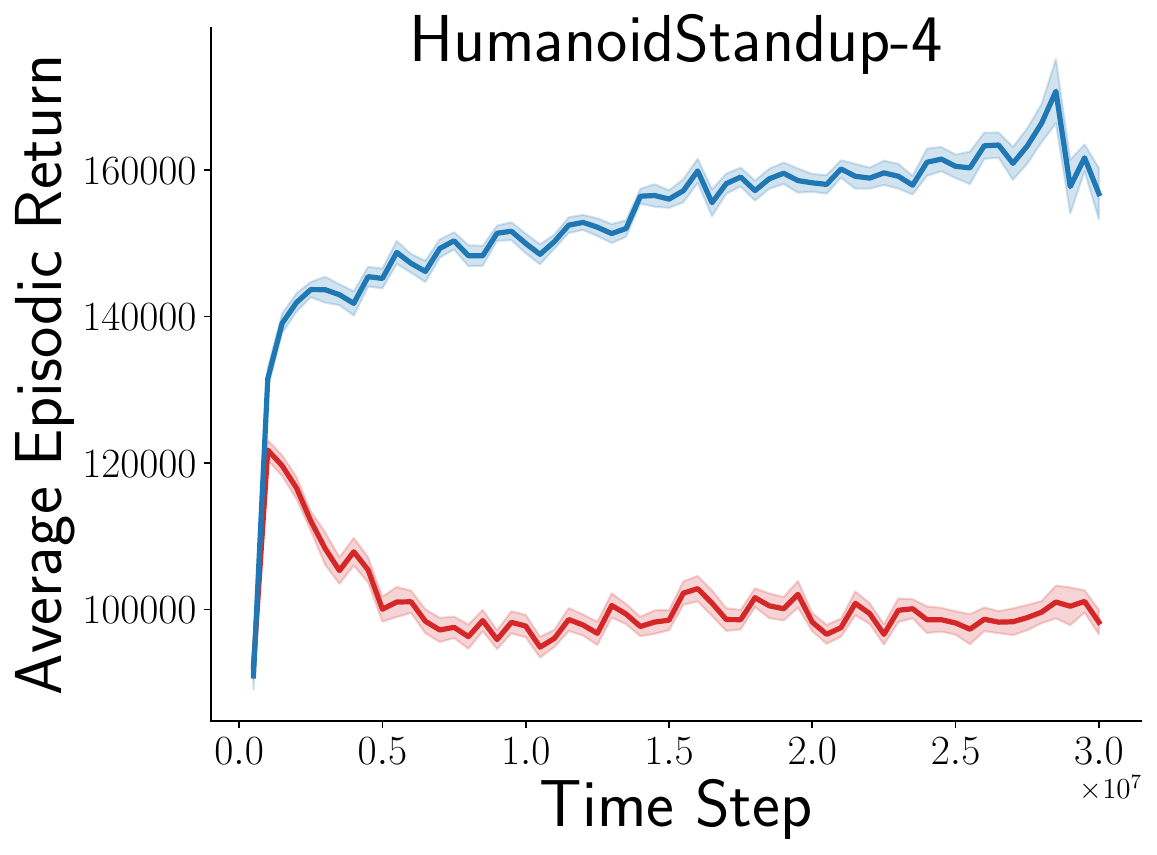}
    \vspace{-0.1cm}
    \includegraphics[width=0.25\textwidth]{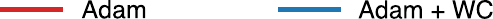}
    \vspace{-0.1cm}
    \caption{Policy Collapse in PPO. The performance of PPO with Adam drops when trained for longer in contrast to Adam+WC, which can keep improving its performance. All curves are averaged over $30$ independent runs. The shaded area represents the standard error.}
    \label{fig:ppo-policy-collapse}
    \vspace{-0.25cm}
\end{figure}

\begin{figure}[ht]
    \centering
    \includegraphics[width=0.244\textwidth]{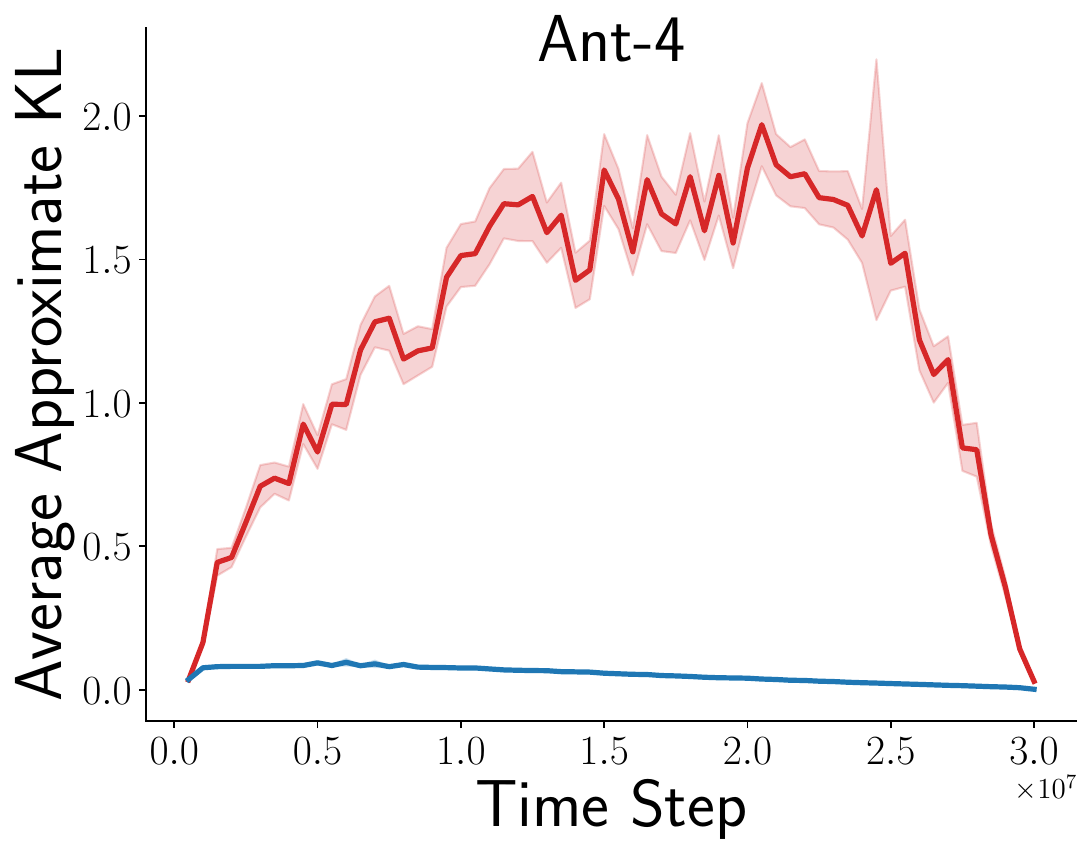}
    \includegraphics[width=0.244\textwidth]{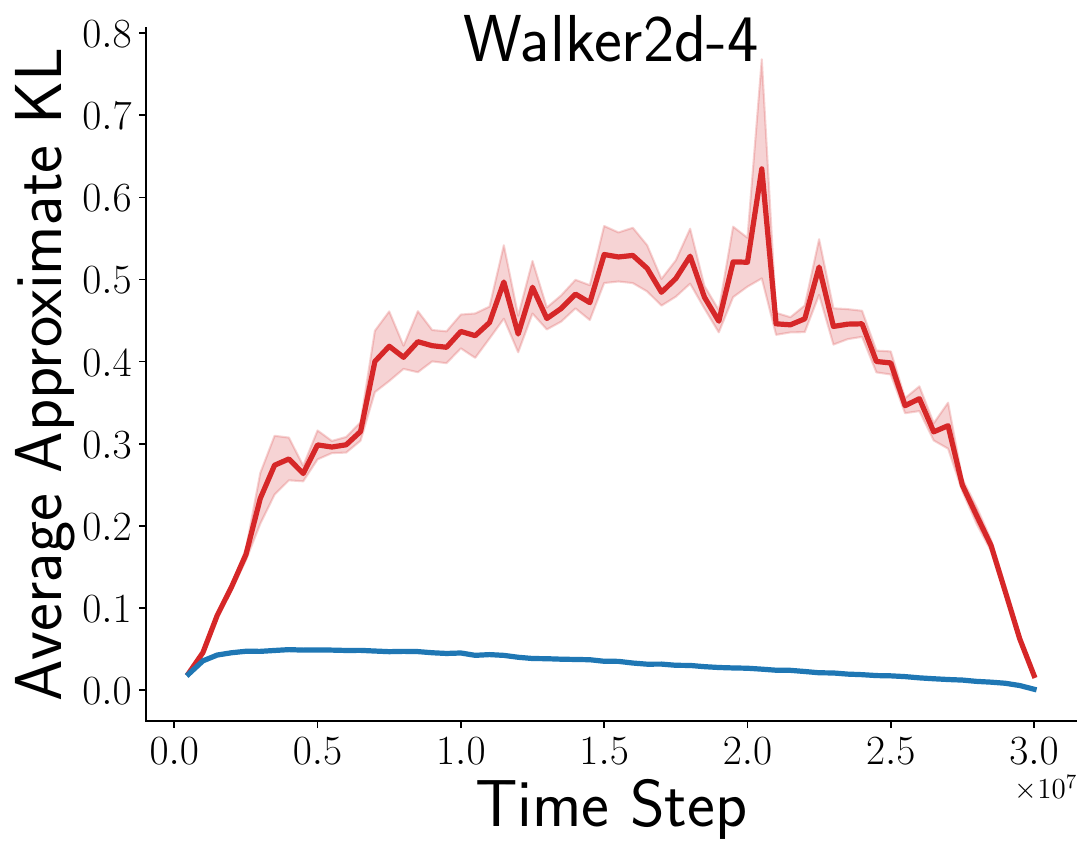}
    \includegraphics[width=0.244\textwidth]{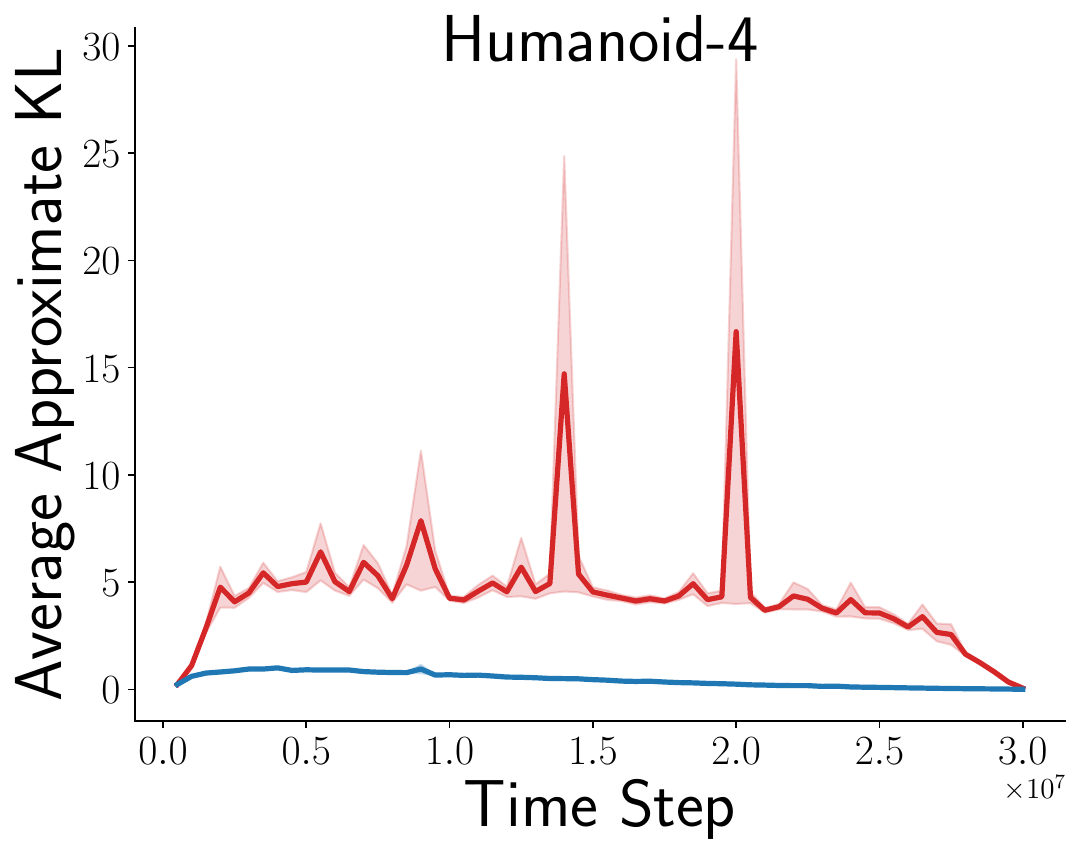}
    \includegraphics[width=0.244\textwidth]{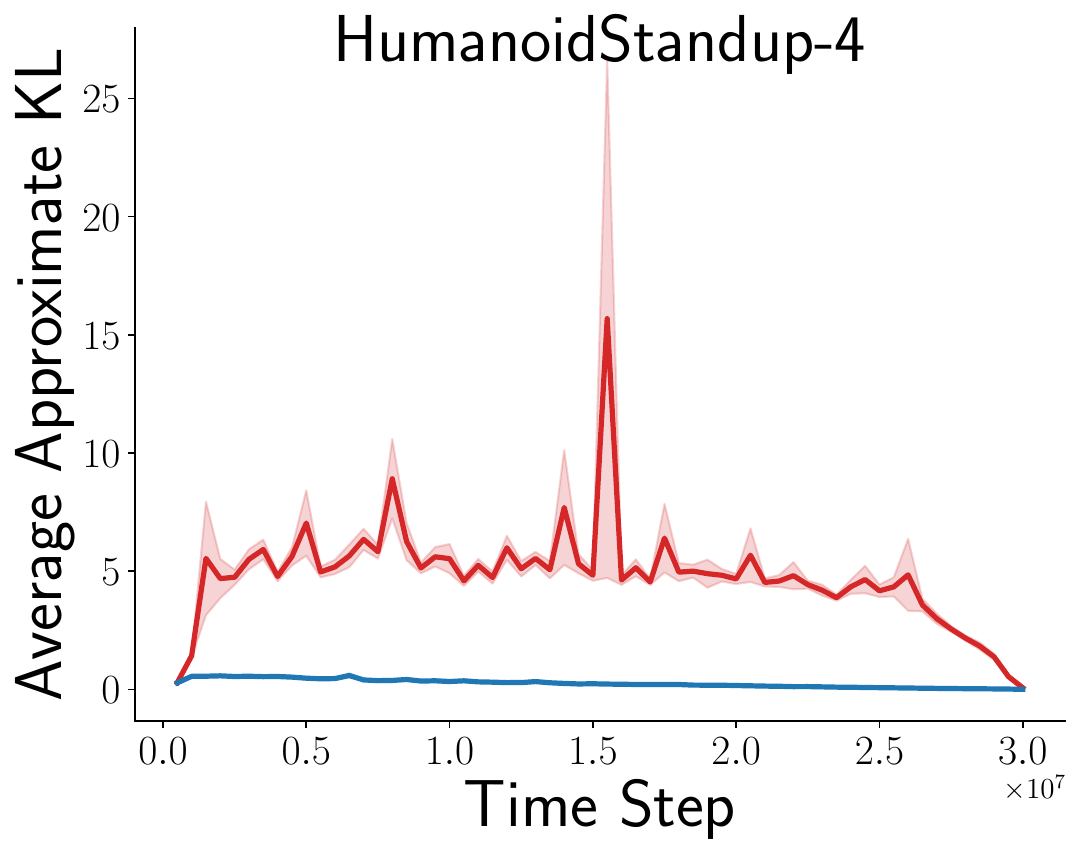}
    \vspace{-0.1cm}
    \includegraphics[width=0.25\textwidth]{figures/ppo/adam-wc.pdf}
    \vspace{-0.1cm}
    \caption{Approximate KL between the current and old policy. The values reach zero at the end since learning rate annealing is used.}
    \label{fig:ppo-kl}
\end{figure}

\begin{figure}[ht]
    \centering
    \begin{subfigure}[b]{0.49\textwidth}
        \centering
    \includegraphics[width=0.49\textwidth]{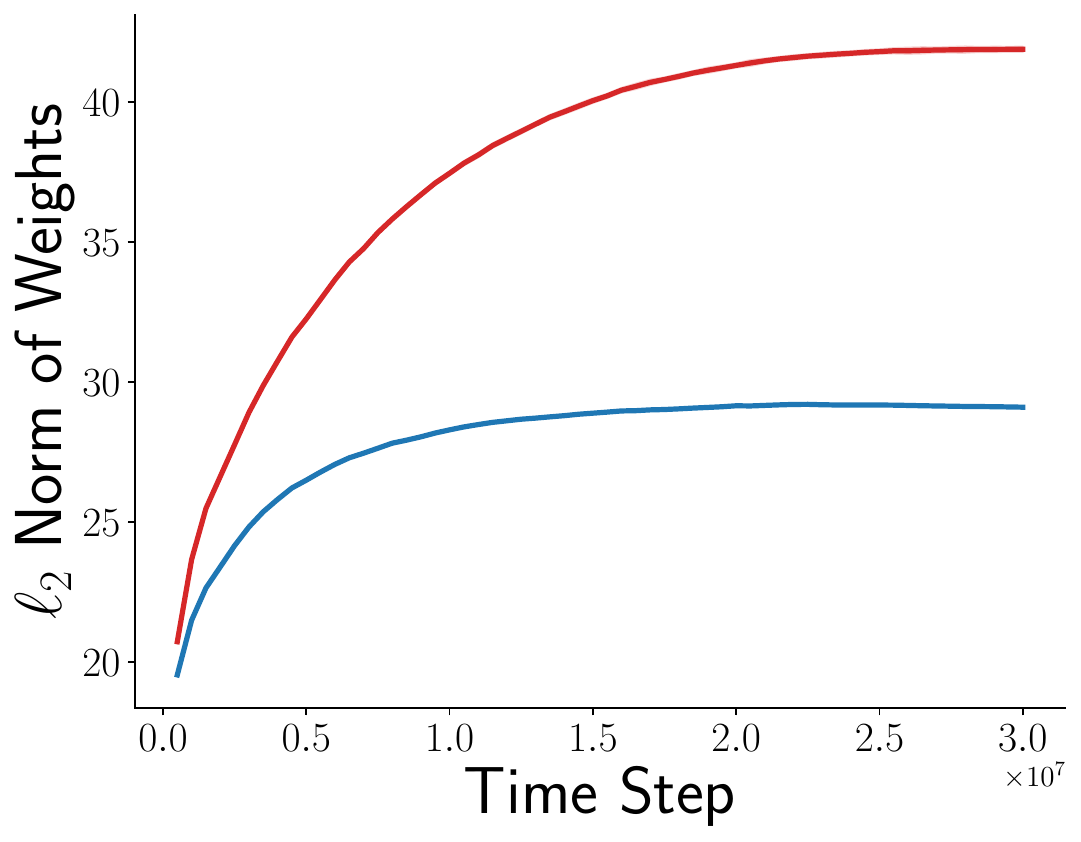}
    \includegraphics[width=0.49\textwidth]{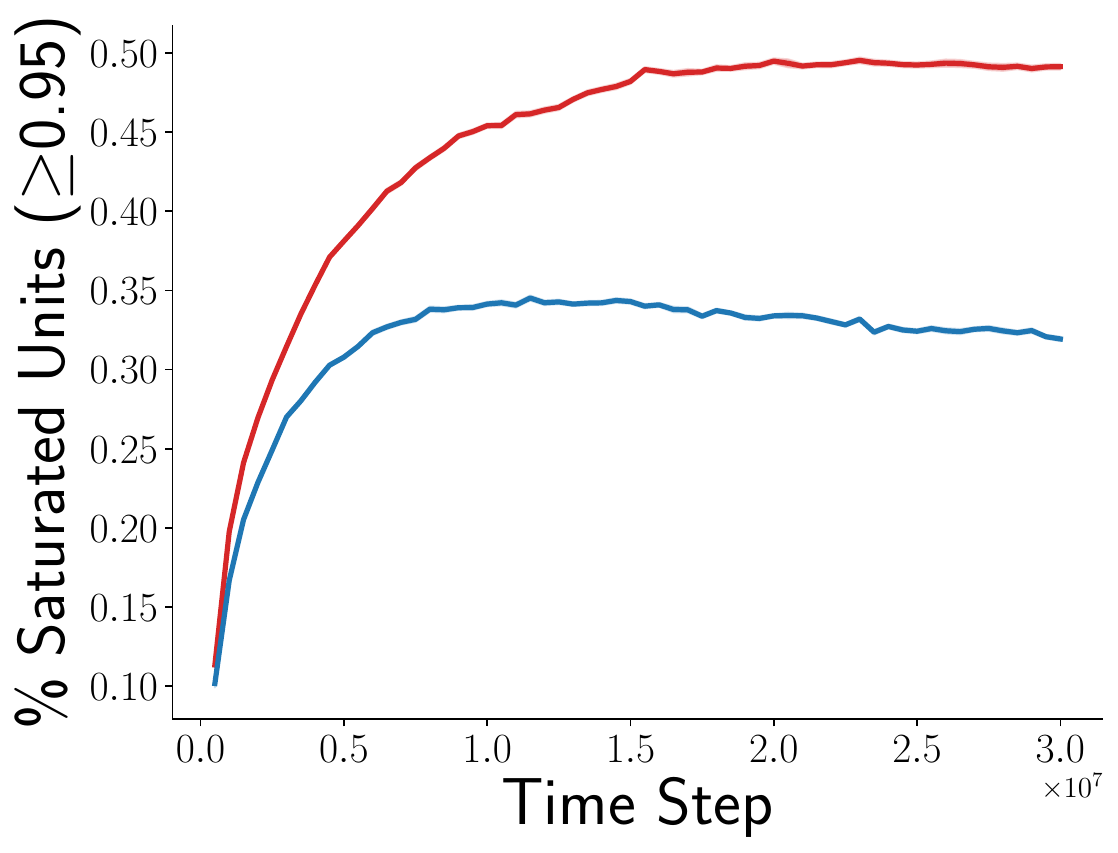}
            \caption{Walker2d-v4}
        \label{fig:diagnostic-walker2d}
    \end{subfigure}
    \hfill
    \begin{subfigure}[b]{0.49\textwidth}
        \centering
    \includegraphics[width=0.49\textwidth]{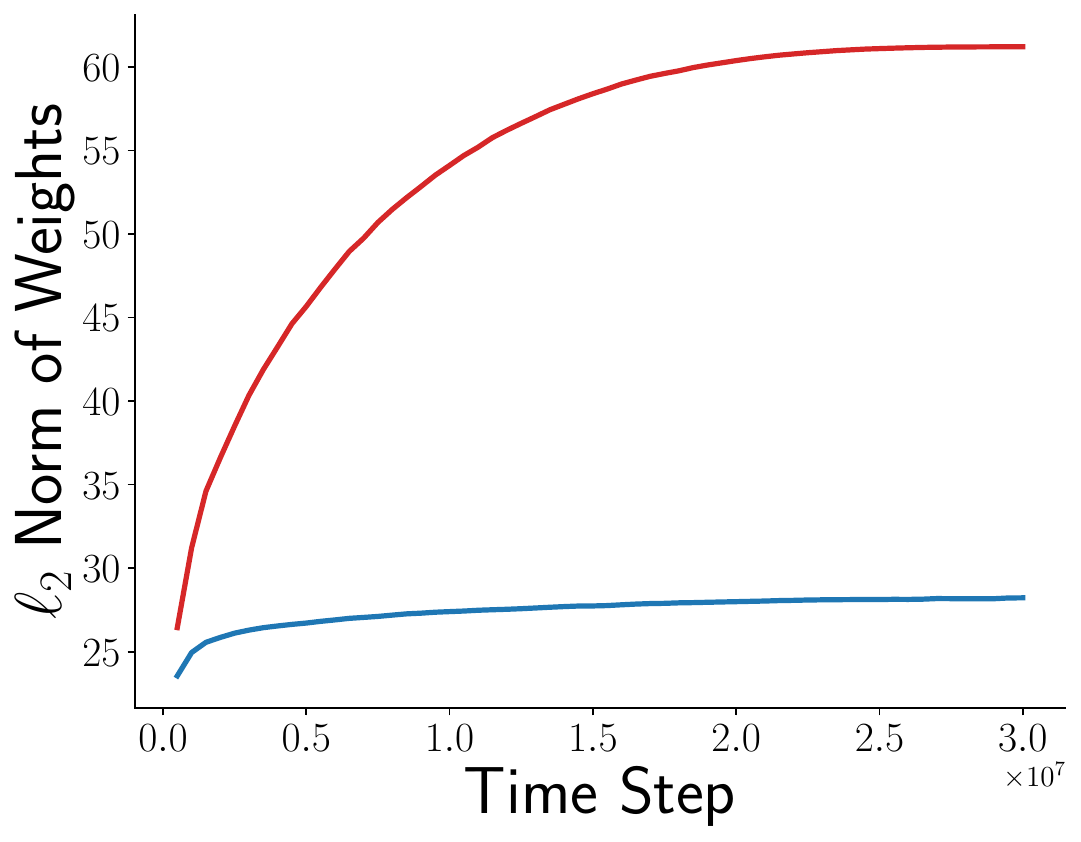}
    \includegraphics[width=0.45\textwidth]{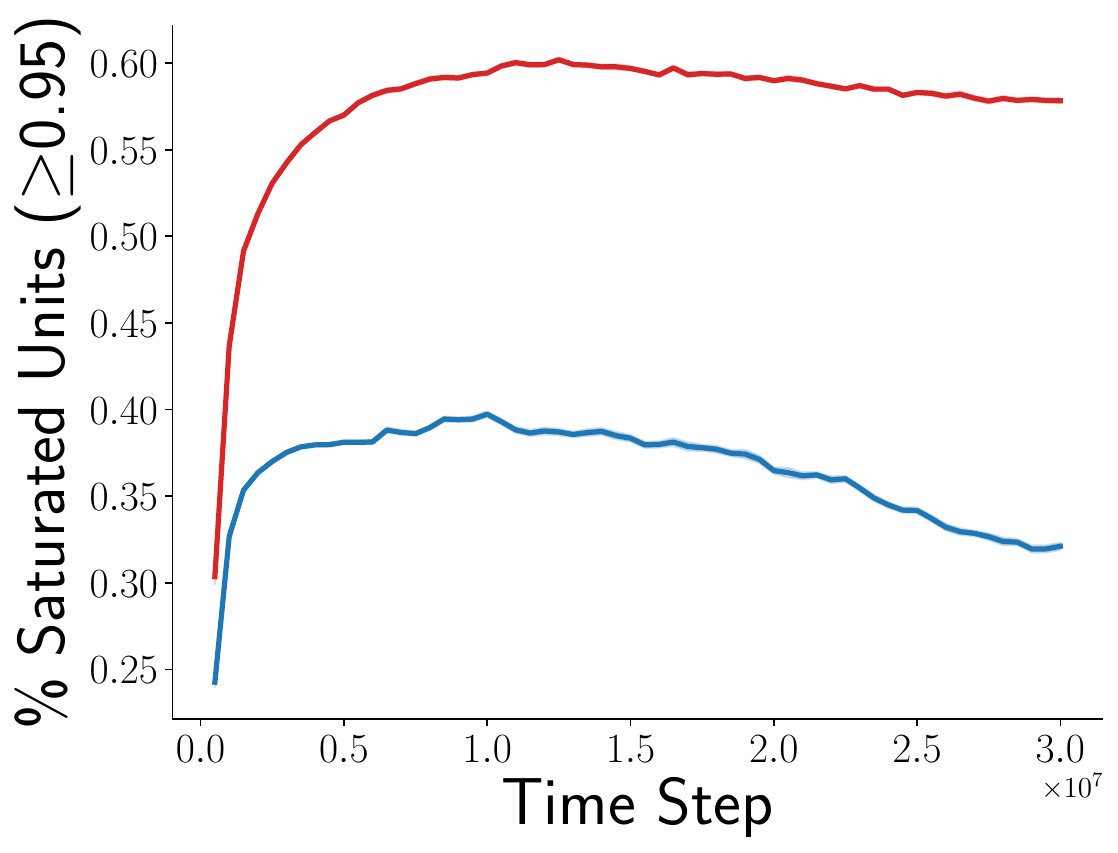}
            \caption{Humanoid-v4}
        \label{fig:diagnostic-humanoid}
    \end{subfigure}
    \vspace{-0.1cm}
    \includegraphics[width=0.25\textwidth]{figures/ppo/adam-wc.pdf}
    \vspace{-0.1cm}
    \caption{Diagnostic Statistics for policy collapse. The $\ell_2$ norm of the weights and percentage of saturated units are shown. A tanh unit is considered saturated if $|x|\geq 0.95$, where $x$ is its output.}
    \label{fig:ppo-diagnostics}
    \vspace{-0.2cm}
\end{figure}

In this section, we study the role of weight clipping in mitigating policy collapse (Dohare et al.\ 2023b). Dohare et al.\ (2023b) demonstrated that the performance of PPO can collapse if trained for longer periods of time. We use CleanRL's implementation (Huang et al.\ 2022) with the default hyper-parameters as suggested by Dohare et al.\ (2023b). The network used is multi-layered ($64\times 64$) with \emph{tanh} activations. Fig.\ \ref{fig:ppo-policy-collapse} illustrates the phenomenon of policy collapse where the performance of PPO with Adam drops gradually with time in a number of MuJoCo (Todorov et al.\ 2012) environments. Weight clipping is effective in mitigating policy collapse, allowing for continual improvement. Here, we only show the performance in the MuJoCo environments with which policy collapse happens and exclude other environments where PPO with Adam experiences no policy collapse. We further investigate why policy collapse happens using the approximate KL given by $(1-r) - \log r$ as our diagnostic metric, where $r$ is the ratio in PPO between the current policy and the old policy. Fig.\ \ref{fig:ppo-kl} shows the approximate KL throughout learning using Adam against Adam+WC.
We observe that the approximate KL with Adam increases, indicating that the current policy deviates a lot from the old policy, in contrast to Adam+WC, which maintains small values, stabilizing learning in PPO.
Next, we show additional diagnostic statistics in Fig.\ \ref{fig:ppo-diagnostics} to characterize the solutions found by Adam and Adam+WC. Notably, we observe that weight clipping reduces the $\ell_2$ norm of the weights and reduces the percentage of saturated units. We defer diagnostic statistics on the rest of the environment in Appendix \ref{appendix:experimental-details}.

\subsection{Weight Clipping with Large Replay Ratios}

The replay ratio (RR) is the number of gradient updates performed per environment step.
An increase of RR helps a learning agent extract as much information as possible from transitions, resulting in a higher sample efficiency.
However, in practice, when the RR is too high, the learning agent would overfit to a small amount of data and lose the plasticity to learn new information due to aggressive parameter updates, thus reducing the learning performance (Nikishin et al.\ 2022).

\begin{figure}[ht]
\centering
\includegraphics[width=0.85\textwidth]{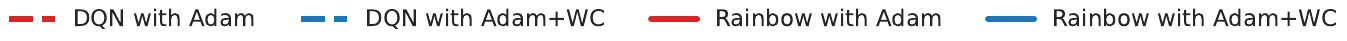}
\includegraphics[width=\textwidth]{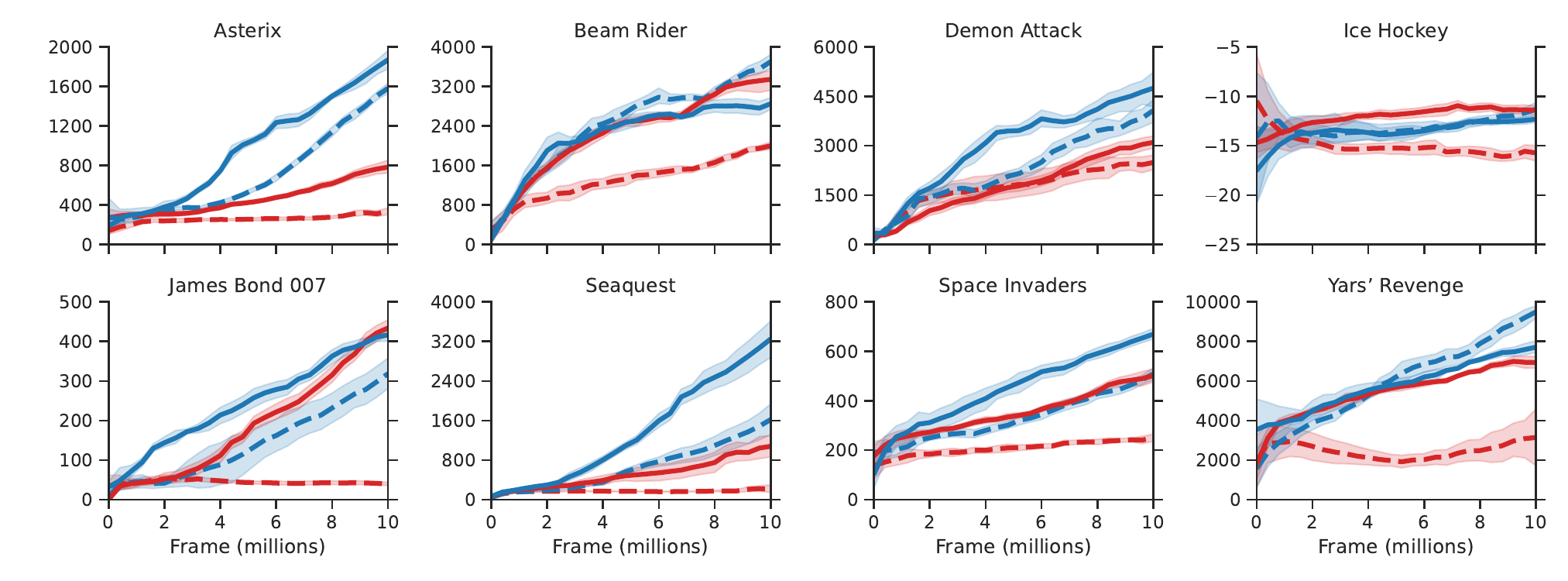}
\caption{The performance of DQN and Rainbow trained with $RR=1$ in Atari tasks, optimized by Adam and Adam+WC. Solid lines correspond to the mean performance over 5 random seeds, and the shaded areas correspond to $90\%$ confidence interval.}
\label{fig:atari}
\vspace{-0.5cm}
\end{figure}

In this section, we show that by incorporating weight clipping, we can prevent the agent from aggressive parameter updates and improve sample efficiency by a large amount under a high RR setting.
Specifically, we train DQN (Mnih et al.\ 2015) and Rainbow (Hessel et al.\ 2018) with $RR=1$ in several Atari (Bellemare et al.\ 2013) tasks, optimized by Adam or Adam+WC.
All our experiments are conducted using the Tianshou framework (Weng et al.\ 2022) and the default RR is $0.1$.
Note that due to the high computation cost of using a high RR, we train both agents for 10M frames and summarize all results across over 5 random seeds, as shown in Figure~\ref{fig:atari}.
Overall, without weight clipping, both DQN and Rainbow perform poorly and have low sample efficiency.
Once weight clipping is applied, we observe a great improvement in sample efficiency and learning performance, demonstrating the effectiveness of our method.

To further investigate the influence of weight clipping in terms of optimization and generalization, we visualize the gradient covariance matrices of training DQN in Space Invaders, optimized by Adam and Adam+WC in Fig.~\ref{fig:atari_grad}.
The heatmaps for all Atari tasks are included in Appendix~\ref{app:grad}.
Specifically, the gradient covariance matrix is known to be strongly related to optimization and generalization (Fort et al.\ 2019, Lyle et al.\ 2022; Lyle et al.\ 2023).
Formally, we estimate it by sampling $k$ training points and compute it entrywise as 
\begin{equation*}
C_k[i, j]=\frac{\langle\nabla_\theta \ell(\theta, \mathbf{x}_i), \nabla_\theta \ell(\theta, \mathbf{x}_j)\rangle}{\|\nabla_\theta \ell(\theta, \mathbf{x}_i)\|\|\nabla_\theta \ell(\theta, \mathbf{x}_j)\|},
\end{equation*}
where $\mathbf{x}_1, \cdots, \mathbf{x}_k$ are randomly sampled training points and $\ell$ is the loss function. In Fig.~\ref{fig:atari_grad}, we observe that when weight clipping is applied, the gradient covariance matrix has smaller off-diagonal values, indicating a smoother loss landscape and less interference.

\section{Related Works}
\label{sec:related-works}
\textbf{Biologically plausible NNs}. Physical and biological systems usually have bounded outputs, as components or elements within these systems often have inherent limits or constraints. For example, in neurobiological processes, synaptic weights are assumed to have a maximum value (Michiels van Kessenich et al.\ 2016). Weight clipping can be viewed as a biologically plausible mechanism that improves plasticity in artificial neural networks.

\begin{wrapfigure}{r}{0.5\textwidth}
\vspace{0cm}
\centering
\includegraphics[width=0.5\textwidth]{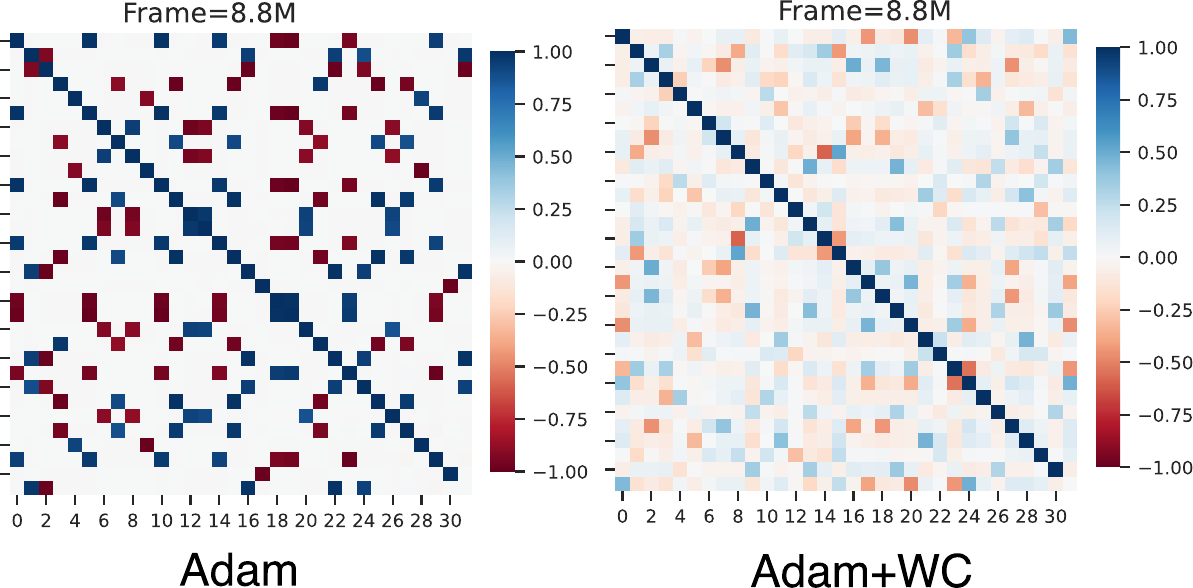}
\caption{The gradient covariance heatmap with DQN in Space Invaders, optimized by Adam and Adam+WC, respectively. Optimizing with Adam+WC results in smaller off-diagonal values in the heatmap compared to Adam.}
\label{fig:atari_grad}
\vspace{-0.4cm}
\end{wrapfigure}
\textbf{Gradient Clipping}. Gradient clipping has been used to stabilize learning which avoids instability by preventing large updates to the network. Barczewski \& Ramon (2023) showed that weight clipping improves optimization performance compared to gradient clipping, likely due to the fact that weight clipping is not biased. Nevertheless, gradient clipping is often an essential component of some reinforcement learning algorithms to stabilize learning and control the maximum size of the update (e.g., Badia et al.\ 2020, Hafner et al.\ 2023, Mnih et al.\ 2016).

\textbf{Weight Clipping in the Literature.} 
Weight clipping has appeared in previous works to achieve different desiderata. For example, Bernstein et al.\ (2020) introduced the Madam optimizer, which uses weight clipping to limit the exponential growth of the weights when using a multiplicative weight update rule. Arjovsky et al. (2017) used weight clipping to stabilize Wasserstein generative adversarial networks. Moreover, weight clipping has been used in binary neural networks to help binarize the weights (Courbariaux et al.\ 2015, Alizadeh et al.\ 2018).

\textbf{Wasserstein Regularization}.
Wasserstein regularization (Lewandowski et al.\ 2024) aims to address the loss of plasticity while allowing parameters to deviate from initialization. However, the Wasserstein metric usually requires sorting of the parameters, which can be expensive. Weight Clipping allows the weights to deviate instead of biasing towards a specific point, achieving a similar goal as Wasserstein regularization but having substantially lower computational requirements.

\section{Conclusion}
\label{sec:conclusion}
In this paper, we introduced weight clipping as a simple mechanism that helps with learning under non-stationarity. Weight clipping can be used besides existing methods without any major change to the optimizer or the network used. Our results show that weight clipping can help mitigate loss of plasticity in streaming learning, alleviate policy collapse, and improve performance when learning with large replay ratios. Future work can perform adaptive weight clipping that does not require any hyper-parameter tuning or develop Lipschitz regularization methods that guarantee smoothness.

\subsubsection*{Acknowledgments}
We gratefully acknowledge funding from the Canada CIFAR AI Chairs program, the Reinforcement Learning and Artificial Intelligence (RLAI) laboratory, the Alberta Machine Intelligence Institute (Amii), and the Natural Sciences and Engineering Research Council (NSERC) of Canada. We would also like to thank the Digital Research Alliance of Canada for providing the computational resources needed. We thank Mark Rowland for the valuable feedback that helped improve this paper. Qingfeng Lan would also like to acknowledge Alberta Innovates for the support they provided him for his research.

\section*{References}
\hangin Abbas, Z., Zhao, R., Modayil, J., White, A., \& Machado, M.\ C.\ (2023). Loss of plasticity in continual deep reinforcement learning. \emph{arXiv preprint arXiv:2303.07507}.\\
\hangin Arjovsky, M., Chintala, S., \& Bottou, L.\ (2017). Wasserstein generative adversarial networks. \emph{International Conference on Machine Learning} (pp. 214-223). 
\hangin Alizadeh, M., Fernández-Marqués, J., Lane, N.\ D., \& Gal, Y.\ (2018). An empirical study of binary neural networks' optimisation. \textit{International Conference on Learning Representations}.\\
\hangin Ash, J., \& Adams, R.\ P.\ (2020). On warm-starting neural network training. \emph{Advances in Neural Information Processing Systems}, \emph{33}, 3884-3894.\\
\hangin Badia, A.\ P., Piot, B., Kapturowski, S., Sprechmann, P., Vitvitskyi, A., Guo, Z.\ D., \& Blundell, C.\ (2020). Agent57: Outperforming the Atari human benchmark. \emph{International Conference on Machine Learning} (pp.\ 507-517).\\
\hangin Bellemare, M.\ G., Naddaf, Y., Veness, J., \& Bowling, M.\ (2013). The arcade learning environment: An evaluation platform for general agents. \emph{Journal of Artificial Intelligence Research}, \emph{47}, 253-279.\\
\hangin Brock, A., De, S., Smith, S.\ L.\ \& Simonyan, K.\ (2021). High-Performance Large-Scale Image Recognition Without Normalization. \emph{International Conference on Machine Learning} (pp.\ 1059-1071).\\
\hangin Bernstein, J., Zhao, J., Meister, M., Liu, M.\ Y., Anandkumar, A., \& Yue, Y.\ (2020). Learning compositional functions via multiplicative weight updates. \emph{Advances in neural information processing systems}, \emph{33}, 13319-13330.\\
\hangin Barczewski, A., \& Ramon, J.\ (2023). DP-SGD with weight clipping. \emph{arXiv preprint arXiv:2310.18001}.\\
\hangin Courbariaux, M., Bengio, Y., \& David, J.\ P.\ (2015). Binaryconnect: Training deep neural networks with binary weights during propagations. Advances in neural information processing systems, 28.\\
\hangin Chizat, L., Oyallon, E., \& Bach, F.\ (2019). On lazy training in differentiable programming. \textit{Advances in Neural Information Processing Systems}, \emph{32}.\\
\hangin Dohare, S., Sutton, R.\ S., \& Mahmood, A.\ R.\ (2021). Continual backprop: Stochastic gradient descent with persistent randomness. \emph{arXiv preprint arXiv:2108.06325}.\\
\hangin Dohare, S., Hernandez-Garcia, J.\ F., Rahman, P., Sutton, R.\ S., \& Mahmood, A. R. (2023a). Loss of plasticity in deep continual learning. \emph{arXiv preprint arXiv:2306.13812}.\\
\hangin Dohare, S., Lan, Q., \& Mahmood, A.\ R.\ (2023b). Overcoming policy collapse in deep reinforcement learning. \emph{European Workshop on Reinforcement Learning}.\\
\hangin Elsayed, M., \& Mahmood, A.\ R.\ (2024). Addressing loss of plasticity and catastrophic forgetting in continual learning. \emph{International Conference on Learning Representations}.\\
\hangin Fort, S., Nowak, P.\ K., Jastrzebski, S., \& Narayanan, S.\ (2019). Stiffness: A new perspective on generalization in neural networks. \emph{arXiv preprint arXiv:1901.09491}.\\
\hangin Garg, S., Tosatto, S., Pan, Y., White, M., \& Mahmood, A.\ R.\ (2022). An alternate policy gradient estimator for softmax policies. \emph{International Conference on Artificial Intelligence and Statistics}.\\
\hangin Gouk, H., Frank, E., Pfahringer, B., \& Cree, M.\ J.\ (2021). Regularisation of neural networks by enforcing Lipschitz continuity. \emph{Machine Learning}, \emph{110}, 393-416.\\
\hangin Geiger, M., Spigler, S., Jacot, A., \& Wyart, M.\ (2020). Disentangling feature and lazy training in deep neural networks. \textit{Journal of Statistical Mechanics: Theory and Experiment}, \textit{2020}(11), 113301.\\
\hangin Ghorbani, B., Mei, S., Misiakiewicz, T., \& Montanari, A.\ (2019). Limitations of lazy training of two-layers neural network. \emph{Advances in Neural Information Processing Systems}, \emph{32}.\\
\hangin Hafner, D., Pasukonis, J., Ba, J., \& Lillicrap, T.\ (2023). Mastering diverse domains through world models. \emph{arXiv preprint arXiv:2301.04104}.\\
\hangin He, K., Zhang, X., Ren, S., \& Sun, J.\ (2015). Delving deep into rectifiers: Surpassing human-level performance on imagenet classification. \emph{IEEE International Conference on Computer Vision} (pp. 1026-1034).\\
\hangin He, K., Zhang, X., Ren, S., \& Sun, J.\ (2016). Deep residual learning for image recognition. \emph{Conference on Computer Vision and Pattern Recognition} (pp. 770-778).\\
\hangin Hessel, M., Modayil, J., Van Hasselt, H., Schaul, T., Ostrovski, G., Dabney, W., ... \& Silver, D.\ (2018). Rainbow: Combining improvements in deep reinforcement learning. \emph{AAAI Conference on Artificial Intelligence} (Vol. 32, No. 1).\\
\hangin Hayes, T. L., \& Kanan, C.\ (2022). Online Continual Learning for Embedded Devices. \emph{Conference on Lifelong Learning Agents} (pp.\ 744-766).\\
\hangin Hayes, T.\ L., Cahill, N.\ D., \& Kanan, C.\ (2019). Memory efficient experience replay for streaming learning. \emph{International Conference on Robotics and Automation} (pp. 9769-9776).\\
\hangin Huang, S., Dossa, R.\ F.\ J., Ye, C., Braga, J., Chakraborty, D., Mehta, K., \& AraÃšjo, J. G. (2022). Cleanrl: High-quality single-file implementations of deep reinforcement learning algorithms. \emph{Journal of Machine Learning Research}, \emph{23}(274), 1-18.\\
\hangin Kumar, S., Marklund, H., \& Van Roy, B.\ (2023a). Maintaining plasticity via regenerative regularization. \emph{arXiv preprint arXiv:2308.11958}.\\
\hangin Kumar, S., Marklund, H., Rao, A., Zhu, Y., Jeon, H.\ J., Liu, Y., \& Van Roy, B.\ (2023b). Continual learning as computationally constrained reinforcement learning. \emph{arXiv preprint arXiv:2307.04345}.\\
\hangin Kingma, D.\ P., \& Ba, J.\ (2014). Adam: A method for stochastic optimization. \emph{arXiv preprint arXiv:1412.6980}.\\
\hangin Krizhevsky, A.\ (2009) \emph{Learning Multiple Layers of Features from Tiny Images}. Ph.D. dissertation, University of Toronto.\\
\hangin Krogh, A., \& Hertz, J.\ (1991). A simple weight decay can improve generalization. \emph{Advances in Neural Information Processing Systems}, \emph{4}.\\
\hangin Lan, Q., \& Mahmood, A.\ R.\ (2023). Elephant neural networks: Born to be a continual learner. \emph{arXiv preprint arXiv:2310.01365}.\\
\hangin Lan, Q., Pan, Y., Luo, J., \& Mahmood, A.\ R.\ (2023). Memory-efficient Reinforcement Learning with Value-based Knowledge Consolidation. \emph{Transactions on Machine Learning Research}.\\
\hangin Li, Q., Haque, S., Anil, C., Lucas, J., Grosse, R.\ B., \& Jacobsen, J.\ H.\ (2019). Preventing gradient attenuation in Lipschitz constrained convolutional networks. \emph{Advances in Neural Information Processing Systems}, \emph{32}.\\
\hangin Liu, H.\ T.\ D., Williams, F., Jacobson, A., Fidler, S., \& Litany, O.\ (2022). Learning smooth neural functions via Lipschitz regularization. \emph{ACM SIGGRAPH 2022 Conference Proceedings} (pp.\ 1-13).\\
\hangin Lewandowski, A., Tanaka, H., Schuurmans, D., \& Machado, M.\ C.\ (2023). Curvature Explains Loss of Plasticity. \emph{arXiv preprint arXiv:2312.00246}.\\
\hangin Lyle, C., Rowland, M., \& Dabney, W.\ (2021). Understanding and Preventing Capacity Loss in Reinforcement Learning. \emph{International Conference on Learning Representations}.\\
\hangin Lyle, C., Rowland, M., Dabney, W., Kwiatkowska, M., \& Gal, Y.\ (2022). Learning dynamics and generalization in reinforcement learning. \emph{International Conference on Machine Learning}.\\
\hangin Lyle, C., Zheng, Z., Nikishin, E., Pires, B.\ A., Pascanu, R., \& Dabney, W.\ (2023). Understanding Plasticity in Neural Networks. \emph{International Conference on Machine Learning}.\\
\hangin Lyle, C., Zheng, Z., Khetarpal, K., van Hasselt, H., Pascanu, R., Martens, J., \& Dabney, W.\ (2024). Disentangling the causes of plasticity loss in neural networks. \emph{arXiv preprint arXiv:2402.18762}.\\
\hangin Mnih, V., Kavukcuoglu, K., Silver, D., Rusu, A.\ A., Veness, J., Bellemare, M. G., ... \& Hassabis, D.\ (2015). Human-level control through deep reinforcement learning. \emph{nature}, \emph{518}(7540), 529-533.\\
\hangin Mnih, V., Badia, A.\ P., Mirza, M., Graves, A., Lillicrap, T., Harley, T., ... \& Kavukcuoglu, K.\ (2016). Asynchronous methods for deep reinforcement learning. \emph{International Conference on Machine Learning} (pp.\ 1928-1937).\\
\hangin Michiels van Kessenich, L., De Arcangelis, L., \& Herrmann, H.\ J.\ (2016). Synaptic plasticity and neuronal refractory time cause scaling behaviour of neuronal avalanches. \emph{Scientific reports}, \emph{6}(1), 32071.\\
\hangin Nikishin, E., Oh, J., Ostrovski, G., Lyle, C., Pascanu, R., Dabney, W., \& Barreto, A.\ (2023). Deep reinforcement learning with plasticity injection. \emph{Advances in Neural Information Processing Systems}, 36.\\
\hangin Nikishin, E., Schwarzer, M., D’Oro, P., Bacon, P.\ L., \& Courville, A.\ (2022). The primacy bias in deep reinforcement learning. \emph{International Conference on Machine Learning} (pp.\ 16828-16847).\\
\hangin Sokar, G., Agarwal, R., Castro, P.S., \& Evci, U.\ (2023). The Dormant Neuron Phenomenon in Deep Reinforcement Learning. \emph{International Conference on Machine Learning}.\\
\hangin Sutton, R.\ S.\ \& Barto, A.\ G.\ (2018). Reinforcement Learning: An Introduction. \emph{MIT Press}.\\
\hangin Schulman, J., Wolski, F., Dhariwal, P., Radford, A., \& Klimov, O.\ (2017). Proximal policy optimization algorithms. \emph{arXiv preprint arXiv:1707.06347}.\\
\hangin Todorov, E., Erez, T., \& Tassa, Y.\ (2012). Mujoco: A physics engine for model-based control. International Conference on Intelligent Robots and Systems (pp.\ 5026-5033).\\
\hangin Weng, J., Chen, H., Yan, D., You, K., Duburcq, A., Zhang, M., ... \& Zhu, J.\ (2022). Tianshou: A highly modularized deep reinforcement learning library. \emph{Journal of Machine Learning Research}, \emph{23}(1), 12275-12280.\\
\hangin Wortsman, M., Liu, P.\ J., Xiao, L., Everett, K., Alemi, A., Adlam, B., ... \& Kornblith, S.\ (2023). Small-scale proxies for large-scale Transformer training instabilities. \emph{arXiv preprint arXiv:2309.14322}.\\
\hangin Yoshida, Y., \& Miyato, T.\ (2017). Spectral norm regularization for improving the generalizability of deep learning. \emph{arXiv preprint arXiv:1705.10941}.\\
\hangin Zhang, C., Bengio, S., Hardt, M., Recht, B., \& Vinyals, O.\ (2021). Understanding deep learning (still) requires rethinking generalization. \emph{Communications of the ACM}, \emph{64}(3), 107-115.


\appendix
\section{Proofs}
\label{appendix:proofs}

\subsection{Proof of Theorem \ref{thm:smoothness}}
\begin{proof}
Let us first rewrite our feed-forward neural network function for a given input $\vx$ as follows:
\begin{align*}
f_{\mathcal{W}} &= (\phi_{L} \circ \sigma \dots \sigma \circ \phi_{1})(\vx),
\end{align*}
where $\phi_i$ defines the linear operation given by $\mW_i$ and $\circ$ denotes the composition between two functions. The composition of $k_1$-Lipschitz and $k_2$-Lipschitz function is $k_1k_2$-Lipschitz (see Gouk et al.\ 2021). Thus, the Lipschitz constant of the network is given by
\begin{align*}
    L(f) &\leq \prod^{L}_{i=1} L(\sigma_i) L(\phi_i) = \prod^{L}_{i=1} L(\phi_i),
\end{align*}
where $L(\phi_i) = \sup_{\va\neq 0} \frac{\|\mW_i \va \|_p}{\|\va\|_p}$ is the operator norm of $\mW_i$. Since $L(f)$ depends on the operator norm of weight matrices at each layer, it can change during learning, producing less smooth functions.

Next, we show the Lipschitz constant of the clipped network. For simplicity of the proof, we consider $p=1$. The Lipschitz constant of the clipped network $L(f^\text{Clipped})$ is bounded by:
\begin{align*}
    L(f^\text{Clipped}) &\leq \prod^{L}_{l=1} \sup_{\va\neq 0} \frac{\|\mW_l \va \|_1}{\|\va\|_1} = \prod^{L}_{l=1} \max_j \left(\sum_{i=1}^{m_l} |W_{l,i,j}| \right) \leq \kappa^L  \prod^{L}_{i=1} s_l m_l, 
\end{align*}
where $\mW_l \in \mathbb{R}^{m_l \times n_l}$ and $|W_{l,i,j}| \leq \kappa s_l, \forall l,i,j$.
\end{proof}
\subsection{Proof of Corollary \ref{corollary:boundedness}}
\begin{proof}
For simplicity, let us vectorize $\mathcal{W}$ and combine it with the input vector $\vx$ in a single vector $\boldsymbol\theta \in \mathcal{P}$. Let us consider the two instances $\boldsymbol\theta^{(1)}$ and $\boldsymbol\theta^{(2)}$ corresponding to before and after making an update with the same input, $\boldsymbol\theta^{(2)}=\boldsymbol\theta^{(1)} + \Delta \boldsymbol\theta^{(1)}$. From Theorem \ref{thm:smoothness}, $\exists k\geq 0$ such that the clipped network $f_\mathcal{W}^\text{Clipped}$ is $k$-Lipschitz. 
Thus we can write the following:
\begin{align*}
    \| f^\text{Clipped}(\boldsymbol\theta^{(2)}) - f^\text{Clipped}(\boldsymbol\theta^{(1)}) \|_1 \leq k \| \boldsymbol\theta^{(2)} - \boldsymbol\theta^{(1)} \|_1, \forall \boldsymbol\theta^{(1)},\boldsymbol\theta^{(2)} \in \mathcal{P}.
\end{align*}
Thus, the change in the function $f_\mathcal{W}^\text{Clipped}$ can be written as:
\begin{align*}
    \| \Delta f_\mathcal{W}^\text{Clipped} \|_1 &\leq k\|\boldsymbol\theta^{(2)} - \boldsymbol\theta^{(1)} \|_1 = k \sum_{l=1}^L \sum_{i=1}^{m_l} \sum_{j=1}^{n_l} |W^{(2)}_{l,i,j} - W^{(1)}_{l,i,j} | \leq 2 k \sum^L_{l=1} m_l n_l s_l,
\end{align*}
where $\mW_l^{(1)} \in \mathbb{R}^{m_l \times n_l}$ is the weight matrix at layer $l$ before making the update and $\mW_l^{(2)} \in \mathbb{R}^{m_l \times n_l}$ is the weight matrix at layer $l$ after making the update.
\end{proof}

\newpage
\section{Hyperparameter Search Space}
\label{appendix:hp-search}
In this section, we present the hyper-parameter search space in Table~\ref{tab:hp-search-space} and the best set of hyper-parameter configurations of each method in Table \ref{tab:best-hp-set}.
\begin{table}[ht]
\vspace{0.4cm}
\begin{tabular}{c|c|c|c}
\textbf{Problem} &  & \textbf{Space} & \textbf{Method} \\ \hline
 & & & \\
 & step size $\alpha$ & $\{0.1, 0.01, 0.001, 0.0001\}$ & All \\
\textbf{Input-permuted}  & & & \\
\textbf{MNIST} & noise std. deviation $\sigma$ & $\{0.0, 0.1, 0.01, 0.001, 0.0001\}$ & S\&P  \\ 
 & & & \\
\textbf{Label-permuted} & weight decay factor $\lambda$ & $\{0.0, 0.1, 0.01, 0.001, 0.0001\}$  &  S\&P, L2 Init \\ 
\textbf{EMNIST} & & & \\
 & clipping param $\kappa$ & $\{1,2,3,4,5\}$ & Madam, WC \\
\textbf{Label-permuted} & & & \\
\textbf{\textit{mini}-ImageNet} & Number of Seeds $N$ & $\{20\}$ & All \\
 & & & \\
\hline
 & & & \\
 & step size $\alpha$ & DQN $\{0.0001\}$,  Rainbow $\{0.0000625\}$ & All \\
  & & & \\
\textbf{Atari} & clipping parameter $\kappa$ & $\{1,5\}$ & Adam+WC \\
 \textbf{Environments} & & & \\
& Number of Seeds $N$ & $\{5\}$ & All \\
 & & & \\
 \hline
  & & & \\
 & step size $\alpha$ & $\{0.0001\}$ & All \\
\textbf{MuJoCo} &  &  &  \\
\textbf{Environemnts} & clipping param $\kappa$ & $\{1,3,5\}$ & Adam+WC \\
& & & \\
& Number of Seeds $N$ & $\{30\}$ & All \\
& & & \\
\hline
& & & \\
 & step size $\alpha$ & $\{0.001\}$ & All \\
& & & \\
\textbf{Warm Starting}  & clipping param $\kappa$ & $\{1,2,5,10,20\}$ & SGD+WC \\
 & & & \\
 & Number of Seeds $N$ & $\{10\}$ & All \\
  & & & \\
\end{tabular}
\caption{Search Space for Streaming Learning Experiments.}
\label{tab:hp-search-space}
\end{table}

\begin{table}[ht]
\vspace{0.25cm}
\begin{tabular}{c|c|c}
\textbf{Problem} & \textbf{Method} & \textbf{Best Set} \\
 \hline
 & &  \\
 & SGD & $\alpha=0.001$\\
 & Adam & $\alpha=0.0001$\\
 & SGD + L2 Init & $\alpha=0.001, \lambda=0.01$\\
 & Adam + L2 Init & $\alpha=0.0001, \lambda=0.001$\\
\textbf{Input-permuted} & SGD + S\&P & $\alpha=0.001, \sigma=0.1, \lambda=0.01$\\
\textbf{MNIST}  & Adam + S\&P & $\alpha=0.0001, \sigma=0.1, \lambda=0.001$\\
 &  SGD + WC &  $\alpha=0.001, \kappa=2.0$\\
 & Adam + WC & $\alpha=0.0001, \kappa=1.0$\\
 & Madam & $\alpha=0.01, \kappa=4, $\\
 & &  \\
 \hline
 & &  \\
 & SGD & $\alpha=0.01$\\
 & Adam & $\alpha=0.0001$\\
 & SGD + L2 Init & $\alpha=0.01, \lambda=0.001$\\
 & Adam + L2 Init & $\alpha=0.001, \lambda=0.01$\\
\textbf{Label-permuted} & SGD + S\&P & $\alpha=0.01, \sigma=0.01, \lambda=0.001$\\
\textbf{EMNIST}  & Adam + S\&P & $\alpha=0.001, \sigma=0.001, \lambda=0.01$\\
 &  SGD + WC &  $\alpha=0.01, \kappa=2.0$\\
 & Adam + WC & $\alpha=0.0001, \kappa=3.0$\\
 & Madam & $\alpha=0.01, \kappa=5, $\\
 & &  \\
 \hline
 & &  \\
 & SGD & $\alpha=0.01$\\
 & Adam & $\alpha=0.0001$\\
 & SGD + L2 Init & $\alpha=0.01, \lambda=0.01$\\
 & Adam + L2 Init & $\alpha=0.001, \lambda=0.01$\\
\textbf{Label-permuted} & SGD + S\&P & $\alpha=0.01, \sigma=0.01, \lambda=0.01$\\
\textbf{\textit{mini}-ImageNet}  & Adam + S\&P & $\alpha=0.001, \sigma=0.0, \lambda=0.01$\\
 &  SGD + WC &  $\alpha=0.01, \kappa=1.0$\\
 & Adam + WC & $\alpha=0.0001, \kappa=3.0$\\
 & Madam & $\alpha=0.01, \kappa=5, $\\
 & &  \\
 \hline
 & &  \\
\textbf{MuJoCo} & Adam &  $\alpha=0.0003$  \\
\textbf{Environments} & Adam+WC & $\alpha=0.0003$ and $\kappa=3$ for all except for Ant-v4 ($\kappa=5$)  \\
 & &  \\
 \hline
 & &  \\
\textbf{Atari} & Adam &  $\alpha=0.0001$ for DQN and  $0.0000625$ for Rainbow  \\
\textbf{Environments} & Adam+WC & $\alpha=0.0001$ for DQN and $0.0000625$ for Rainbow with $\kappa=1$ \\
 & &  \\
 \hline
 & &  \\
 & SGD & $\alpha=0.001$\\
\textbf{Warm Starting} & SGD + WC & $\alpha=0.001, \kappa=10$\\
 & SGD + WC@300 & $\alpha=0.001, \kappa=20$\\
 & &  \\
\end{tabular}
\caption{Best hyperparameter set of each method in each problem.}
\label{tab:best-hp-set}
\end{table}

\clearpage
\section{Additional Experimental Results}
\label{appendix:experimental-details}
\subsection{Diagnostic statistics for Label-permuted EMNIST and \textit{mini}-ImageNet}
\begin{figure}[ht]
    \centering
        \begin{subfigure}[b]{\textwidth}
    \includegraphics[width=0.24\textwidth]{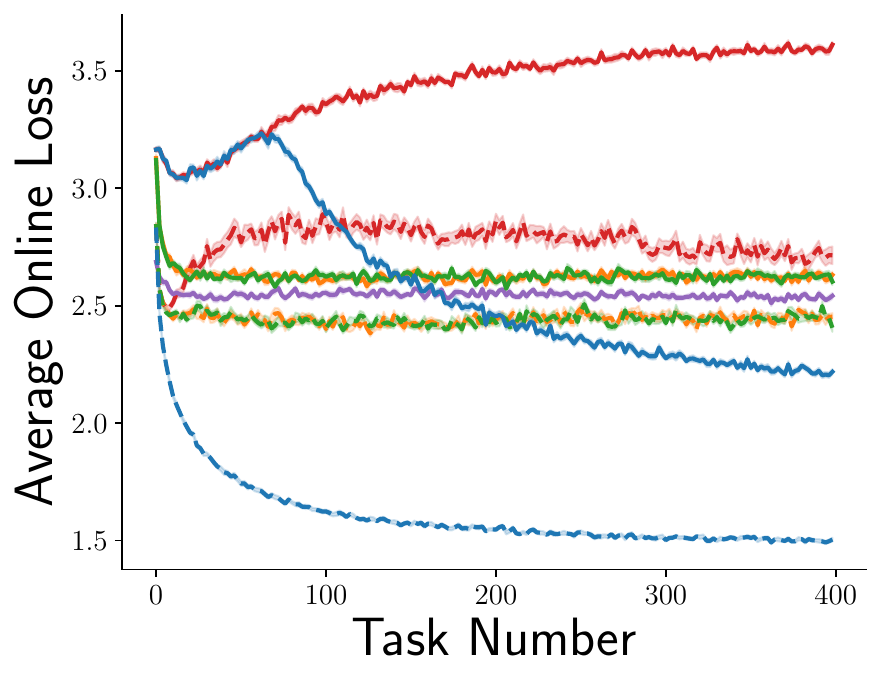}
    \includegraphics[width=0.244\textwidth]{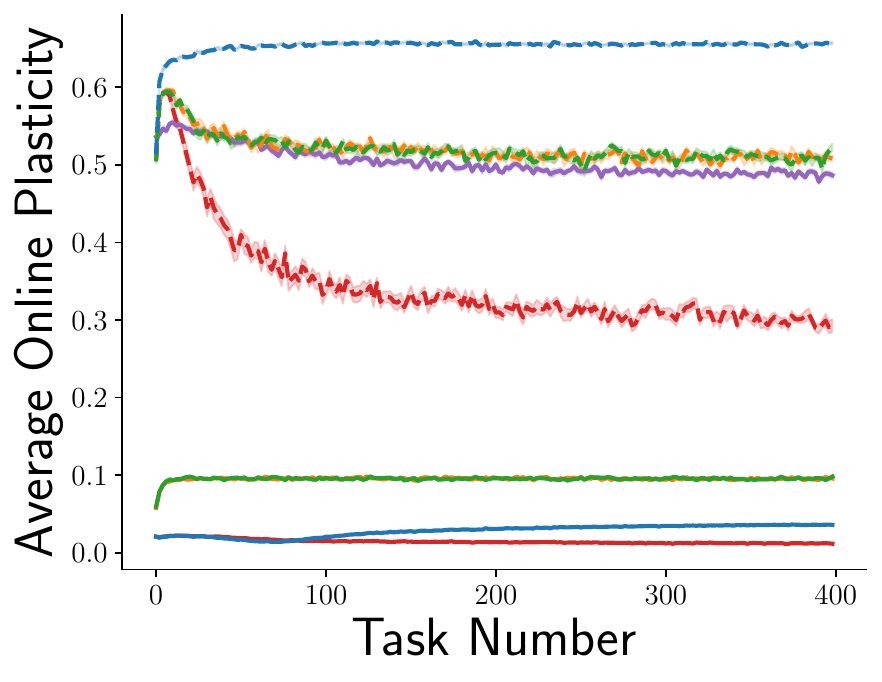}
    \includegraphics[width=0.244\textwidth]{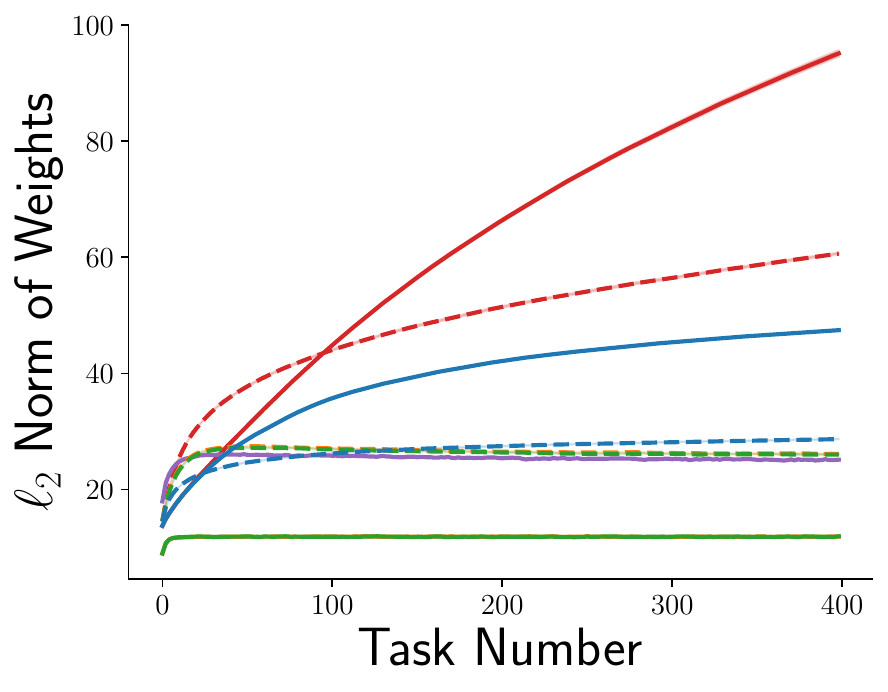}
    \includegraphics[width=0.244\textwidth]{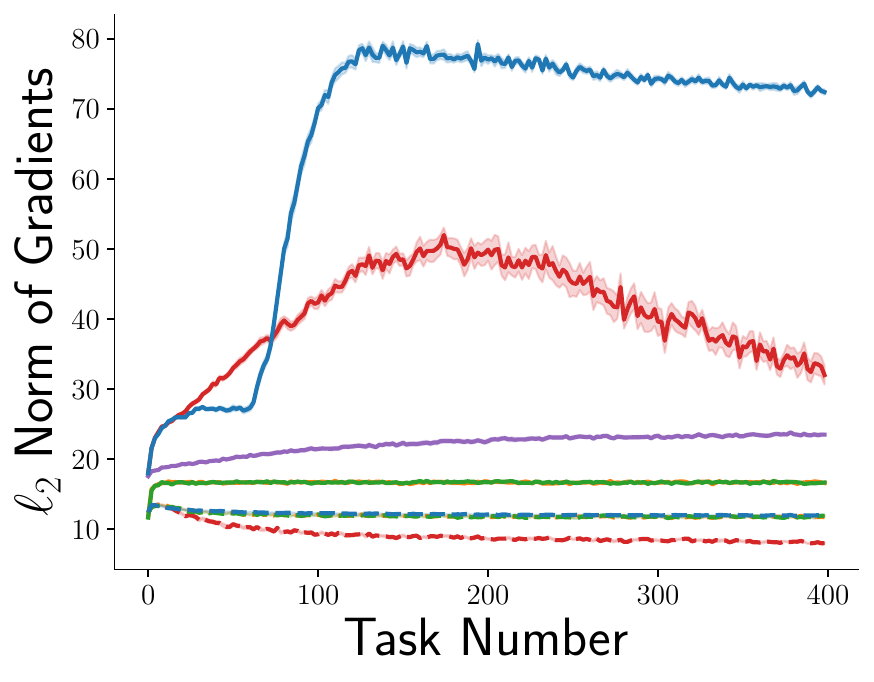}
    \caption{Label-pertmued EMNIST.}
    \end{subfigure}
        \begin{subfigure}[b]{\textwidth}
    \includegraphics[width=0.24\textwidth]{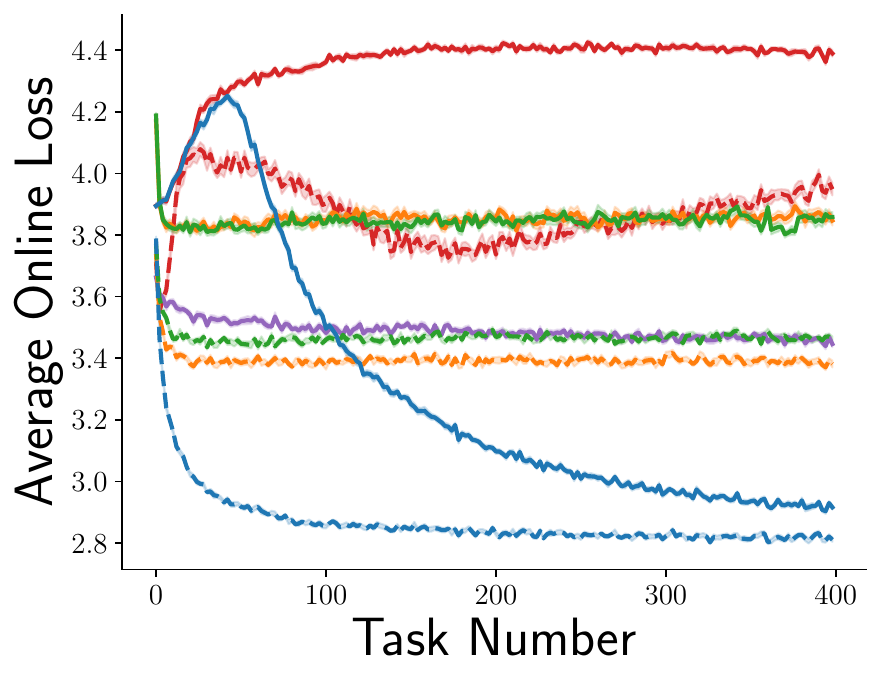}
    \includegraphics[width=0.244\textwidth]{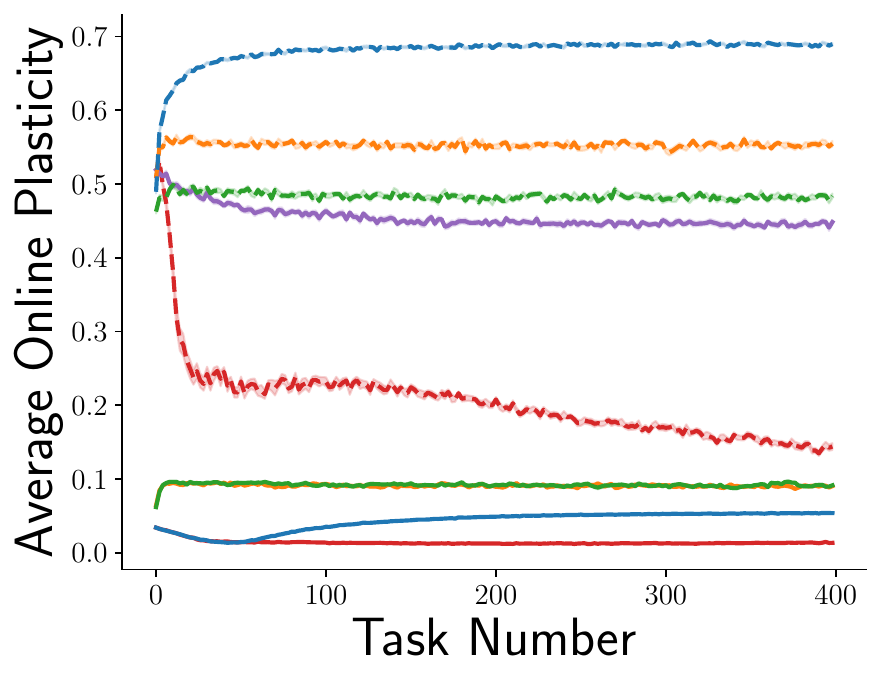}
    \includegraphics[width=0.244\textwidth]{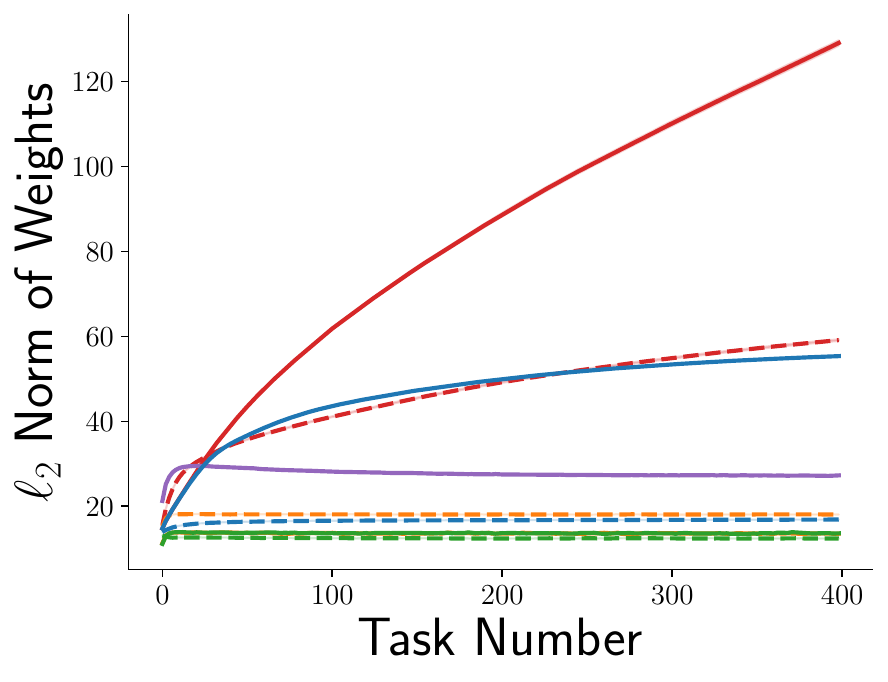}
    \includegraphics[width=0.244\textwidth]{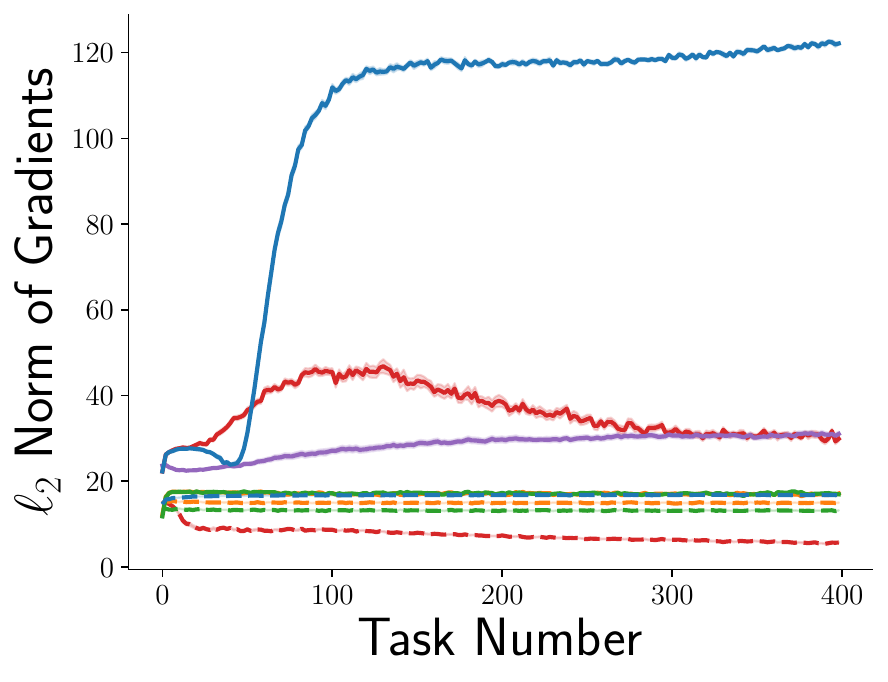}
    \caption{Label-pertmued \emph{mini}-ImageNet.}
    \end{subfigure}
    \includegraphics[width=\textwidth]{figures/SL/wide-legend.pdf}
    \caption{Diagnostic Statistics of different methods in Label-permuted EMNIST and Label-permuted \emph{mini}-ImageNet. We show the online loss, the online plasticity, the $\ell_2$-norm of gradients, and the $\ell_2$-norm of weights.}
    \label{fig:additional-streaming-stats}
\end{figure}

\subsection{Diagnostic statistics of PPO}
\begin{figure}[ht]
    \centering
    \begin{subfigure}[b]{0.49\textwidth}
        \centering
    \includegraphics[width=0.49\textwidth]{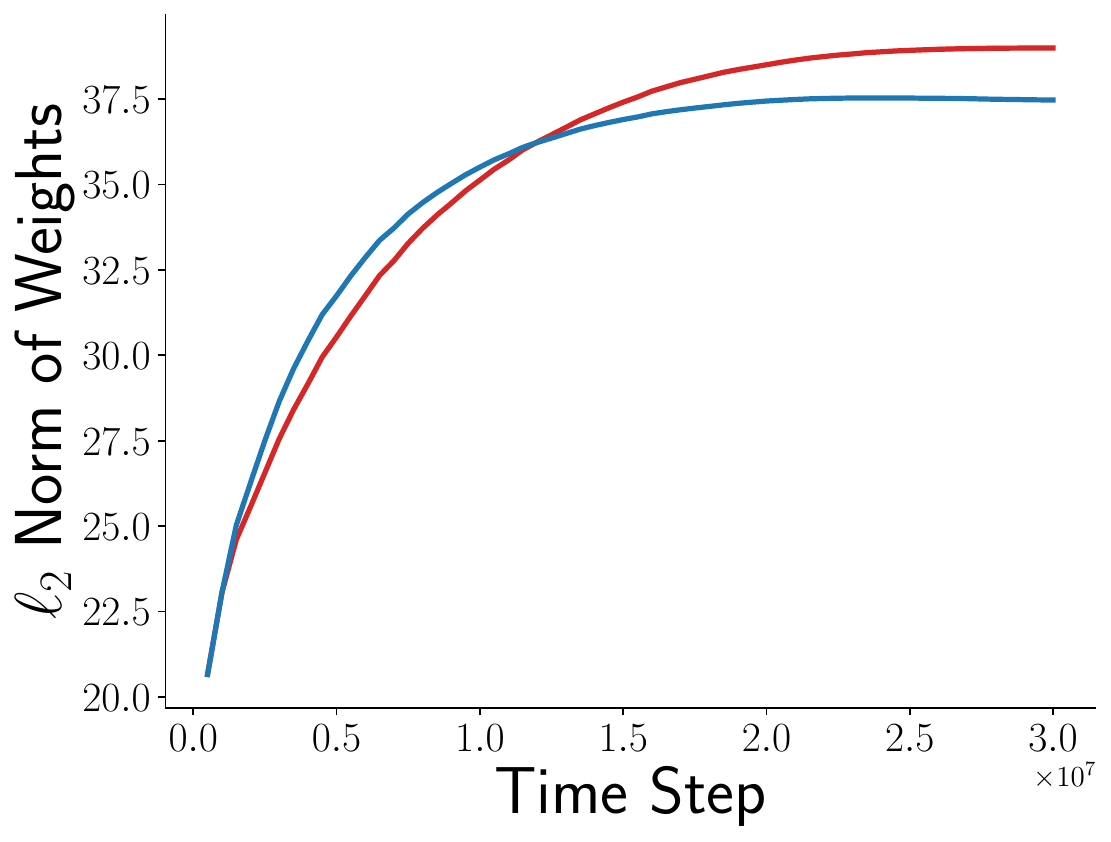}
    \includegraphics[width=0.49\textwidth]{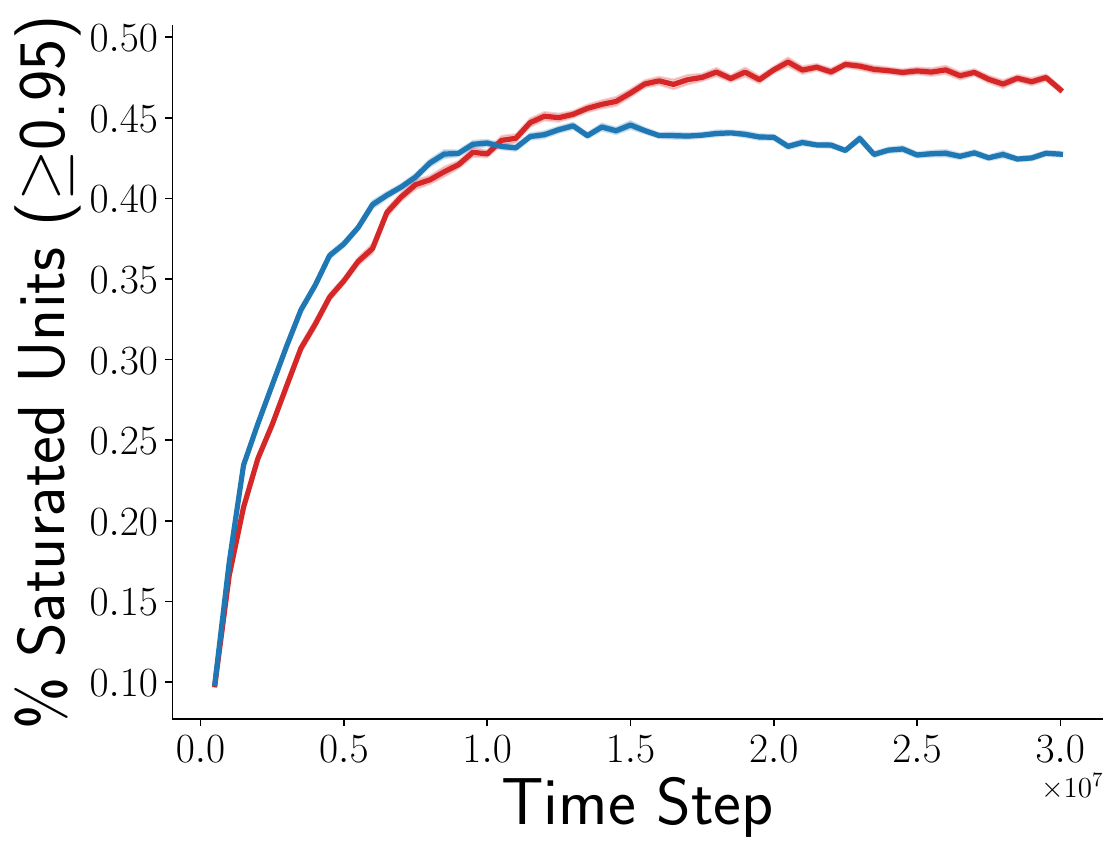}
            \caption{Ant-v4}
        \label{fig:diagnostic-ant-appendix}
    \end{subfigure}
    \hfill
    \begin{subfigure}[b]{0.49\textwidth}
        \centering
    \includegraphics[width=0.49\textwidth]{figures/ppo/Walker2d/Walker2d_weight_l2-v4.pdf}
    \includegraphics[width=0.49\textwidth]{figures/ppo/Walker2d/Walker2d_n_saturated_95-v4.pdf}
            \caption{Walker2d-v4}
        \label{fig:diagnostic-walker2d-appendix}
    \end{subfigure}
    \hfill
    \begin{subfigure}[b]{0.49\textwidth}
        \centering
    \includegraphics[width=0.49\textwidth]{figures/ppo/Humanoid/Humanoid_weight_l2-v4.pdf}
    \includegraphics[width=0.49\textwidth]{figures/ppo/Humanoid/Humanoid_n_saturated_95-v4.pdf}
            \caption{Humanoid-v4}
        \label{fig:diagnostic-humanoid-appendix}
    \end{subfigure}
    \begin{subfigure}[b]{0.49\textwidth}
        \centering
    \includegraphics[width=0.49\textwidth]{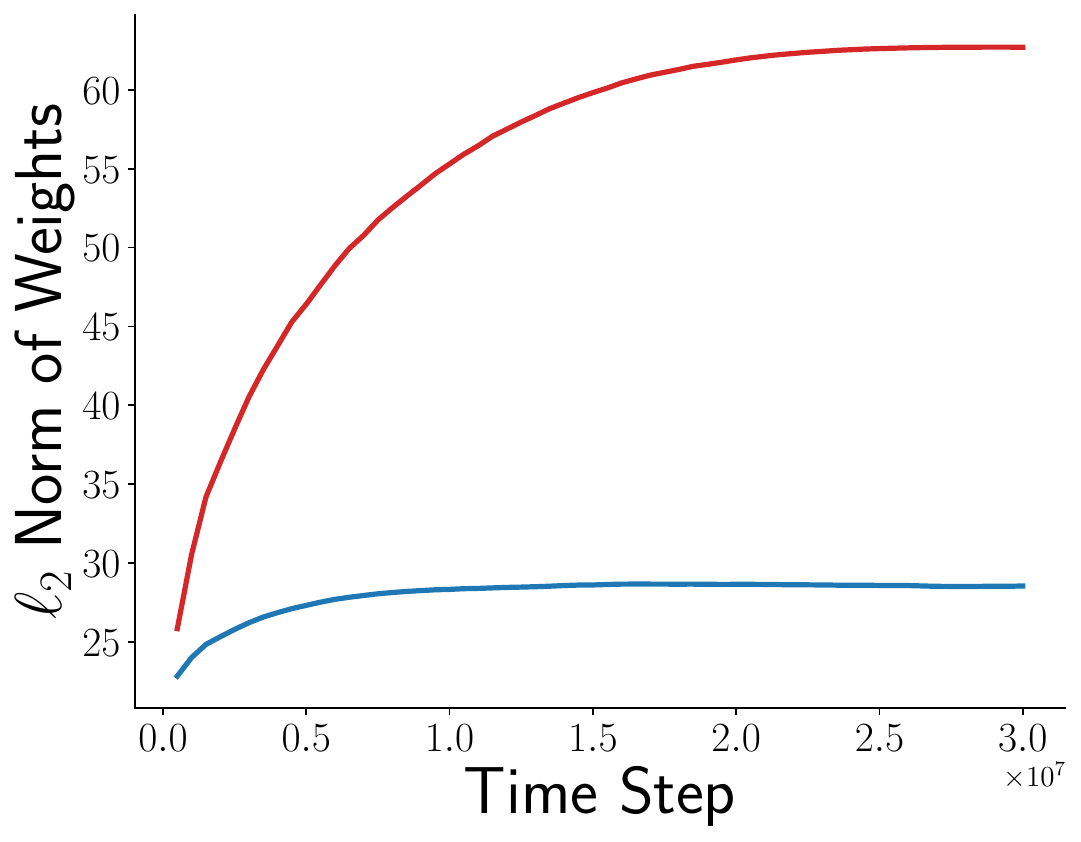}
    \includegraphics[width=0.49\textwidth]{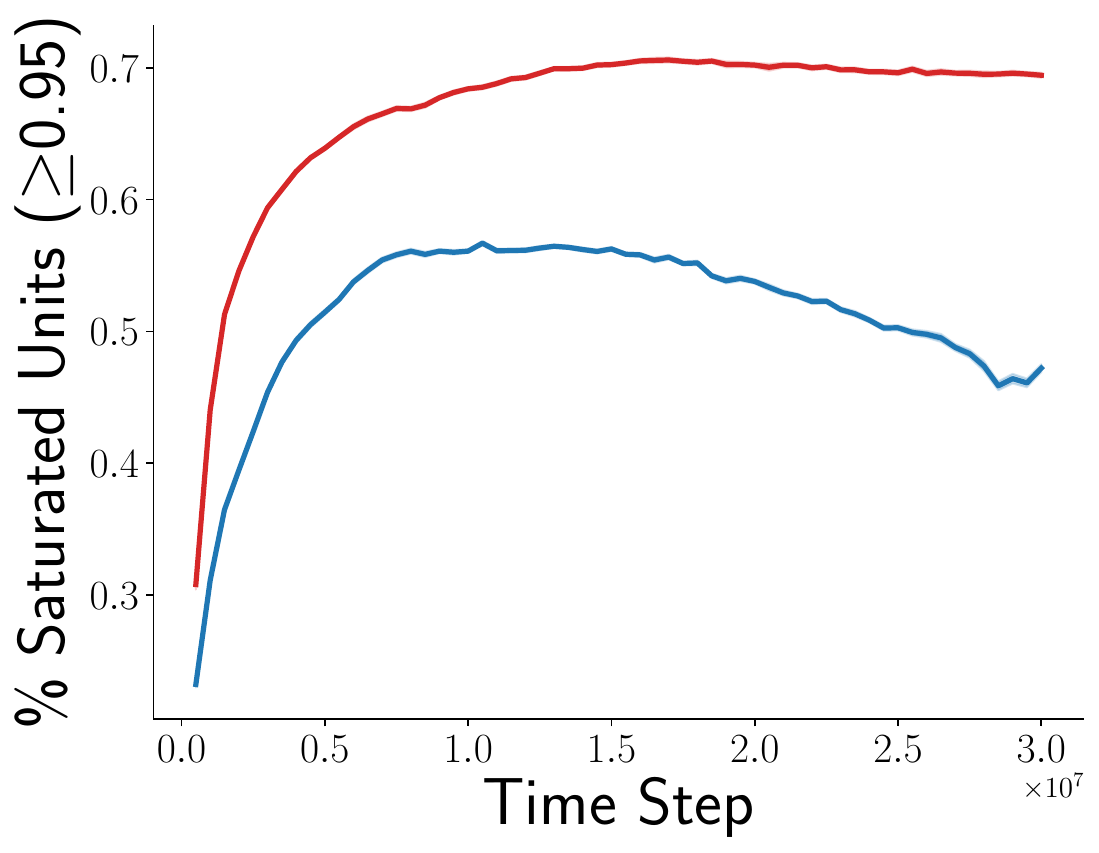}
            \caption{HumanoidStandup-v4}
        \label{fig:diagnostic-humanoidstandup-appendix}
    \end{subfigure}
    \vspace{-0.1cm}
    \includegraphics[width=0.25\textwidth]{figures/ppo/adam-wc.pdf}
    \vspace{-0.1cm}
    \caption{Diagnostic Statistics for policy collapse. The $\ell_2$ norm of the weights and percentage of saturated units are shown. A tanh unit is considered saturated if $|x|\leq 0.95$, where $x$ is its output. In all environments, we use $\kappa=3$ except for Ant-v4, in which we found that $\kappa=5$ performs better. The larger value of $\kappa$ used in Ant-v4 explains the smaller effect of weight clipping on the $\ell_2$ norm of the weights and the percentage of saturated units compared to other environments, although it still has a large effect on reducing the approximate KL (see Fig.\ \ref{fig:ppo-kl}).}
    \label{fig:ppo-diagnostics-all}
\end{figure}

\subsection{The Gradient Covariance Heatmaps of Training DQN and Rainbow in Atari Games}
\label{app:grad}

\begin{figure}[htbp]
\centering
\begin{subfigure}[b]{0.49\textwidth}
\includegraphics[width=\textwidth]{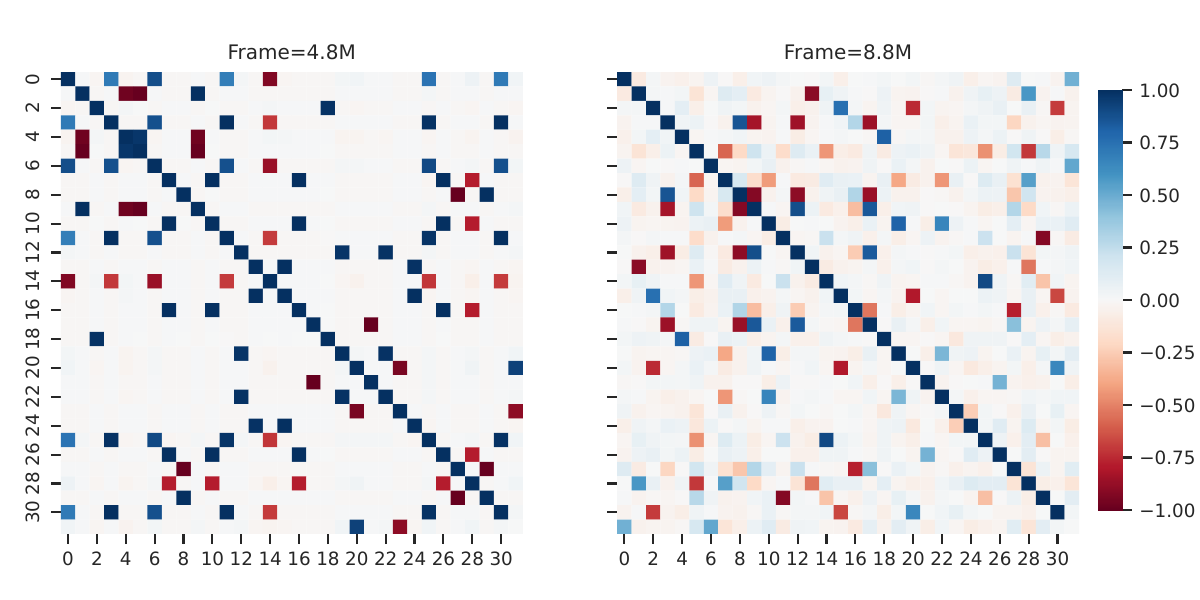}
\caption{DQN in Asterix with Adam}
\end{subfigure}
\begin{subfigure}[b]{0.49\textwidth}
\includegraphics[width=\textwidth]{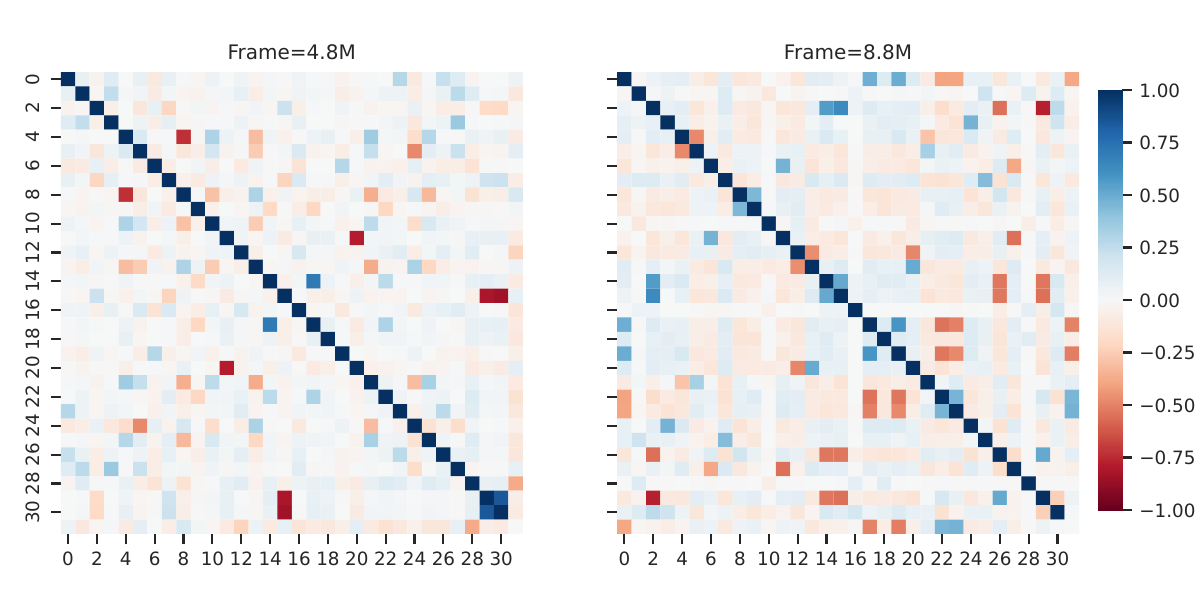}
\caption{DQN in Asterix with Adam+WC}
\end{subfigure}
\hfill
\begin{subfigure}[b]{0.49\textwidth}
\includegraphics[width=\textwidth]{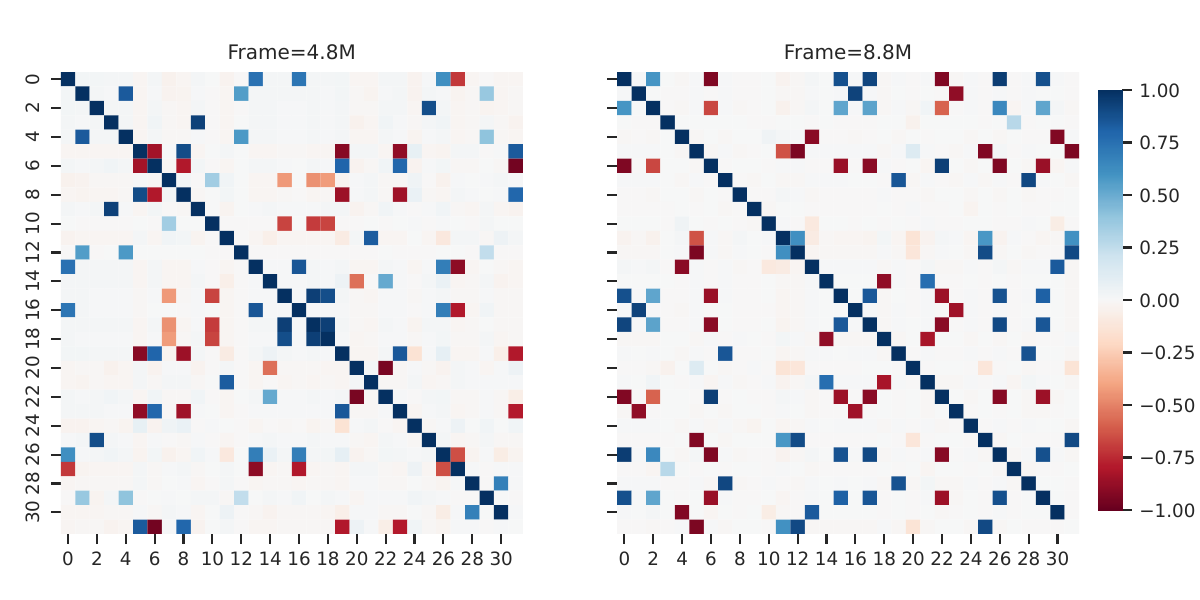}
\caption{DQN in Beam Rider with Adam}
\end{subfigure}
\begin{subfigure}[b]{0.49\textwidth}
\includegraphics[width=\textwidth]{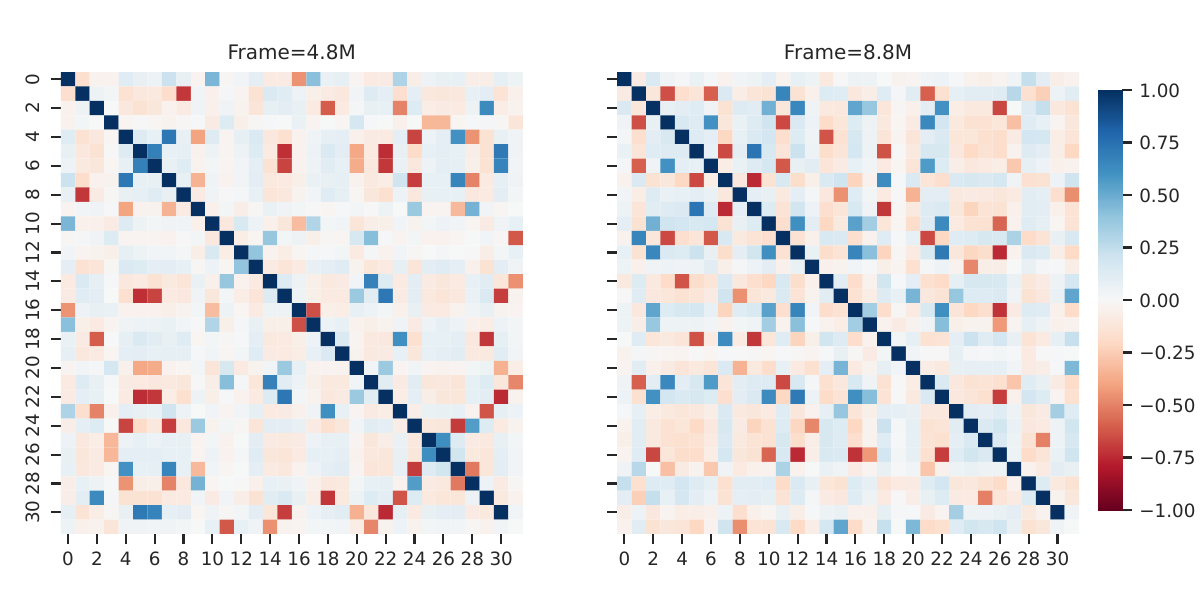}
\caption{DQN in Beam Rider with Adam+WC}
\end{subfigure}
\hfill
\begin{subfigure}[b]{0.49\textwidth}
\includegraphics[width=\textwidth]{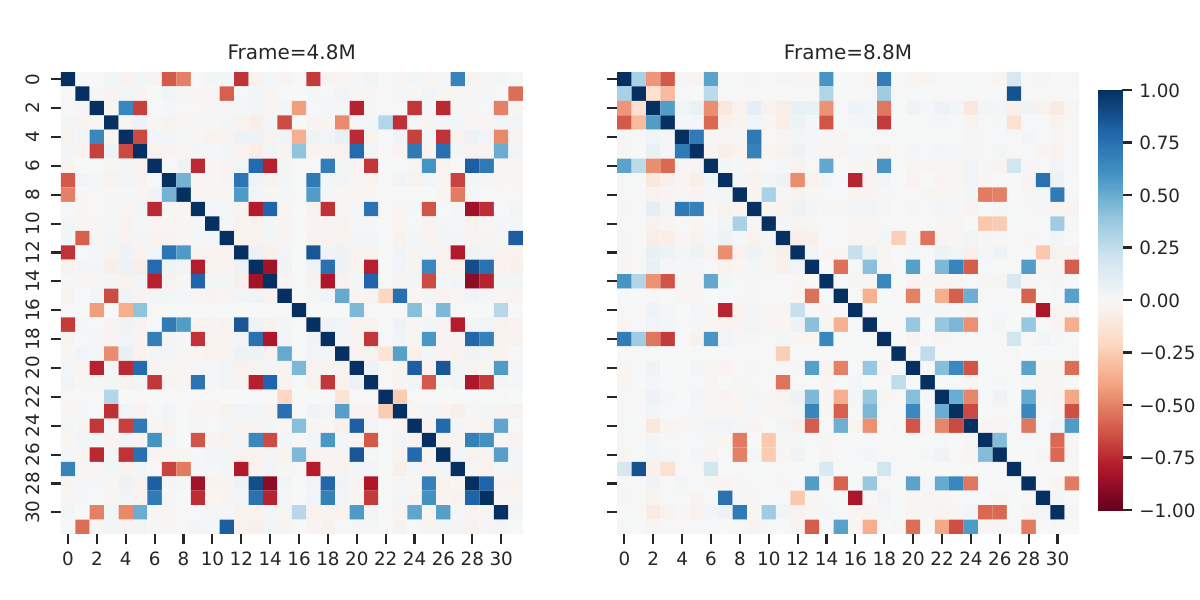}
\caption{DQN in Demon Attack with Adam}
\end{subfigure}
\begin{subfigure}[b]{0.49\textwidth}
\includegraphics[width=\textwidth]{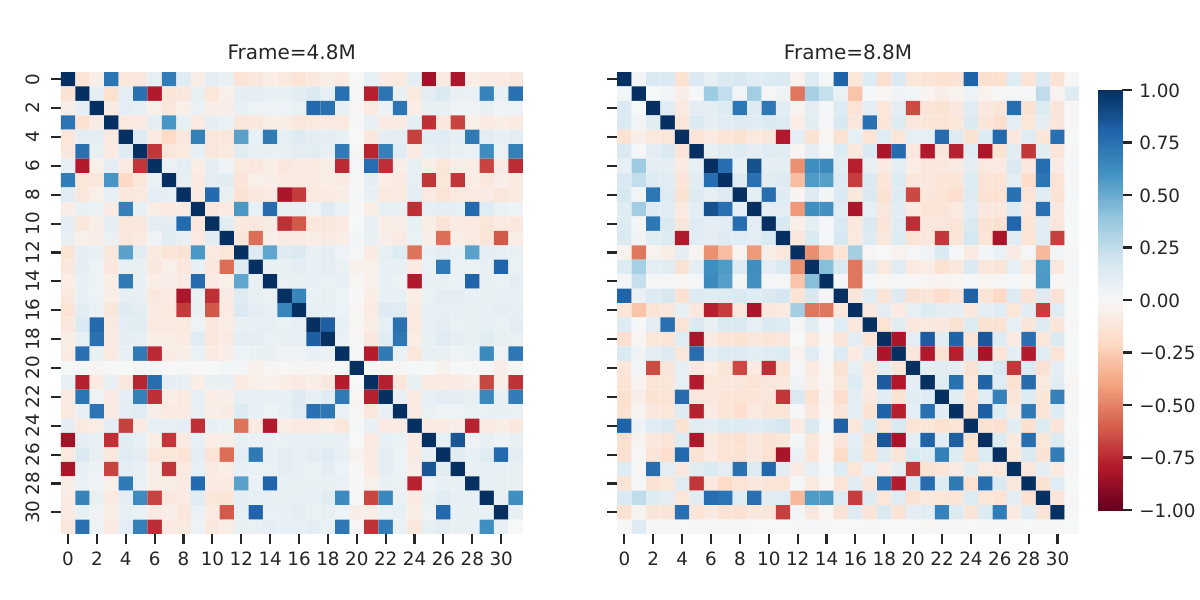}
\caption{DQN in Demon Attack with Adam+WC}
\end{subfigure}
\hfill
\begin{subfigure}[b]{0.49\textwidth}
\includegraphics[width=\textwidth]{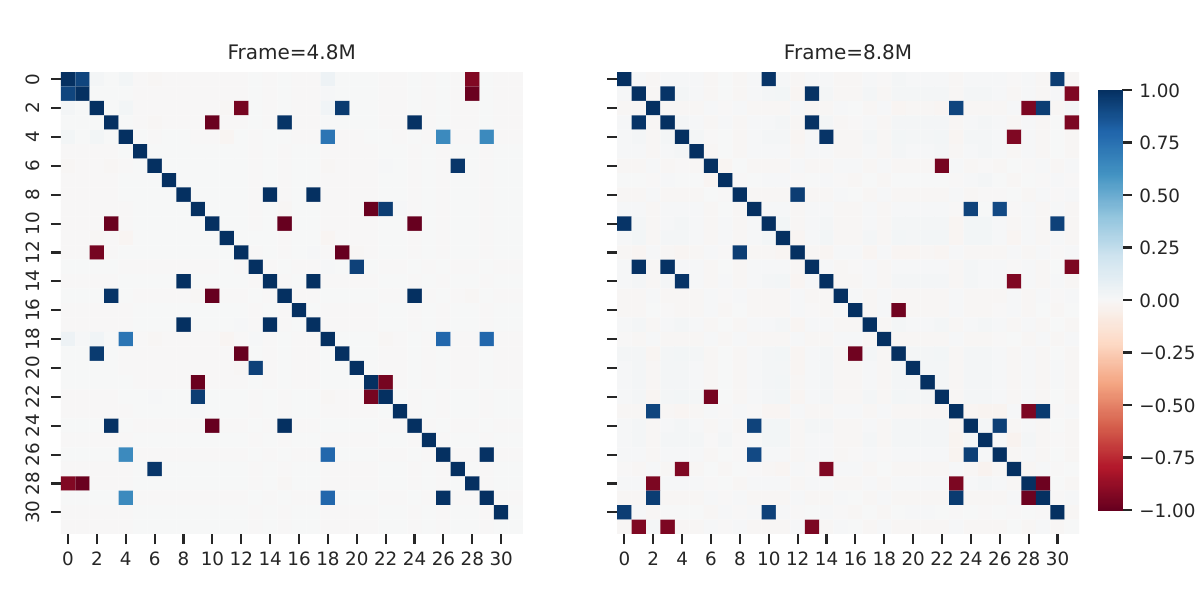}
\caption{DQN in Ice Hockey with Adam}
\end{subfigure}
\begin{subfigure}[b]{0.49\textwidth}
\includegraphics[width=\textwidth]{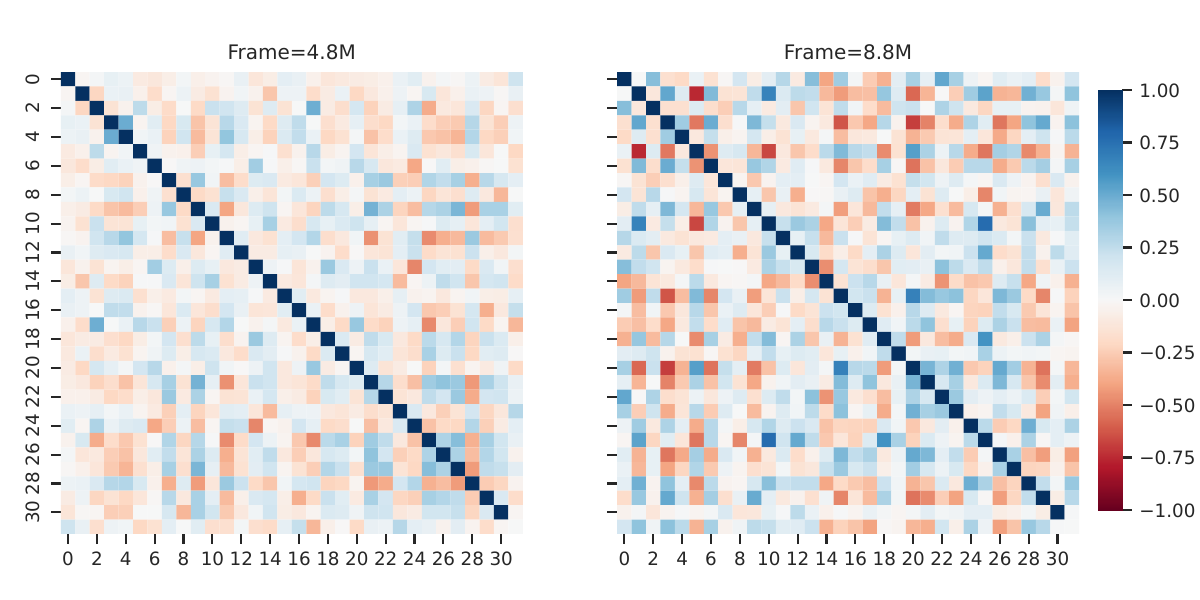}
\caption{DQN in Ice Hockey with Adam+WC}
\end{subfigure}
\caption{The gradient covariance heatmaps of training DQN in Asterix, Beam Rider, Demon Attack, and Ice Hockey, optimized by Adam and Adam+WC, respectively.}
\label{fig:atari_grad_dqn1}
\end{figure}

\begin{figure}[htbp]
\centering
\begin{subfigure}[b]{0.49\textwidth}
\includegraphics[width=\textwidth]{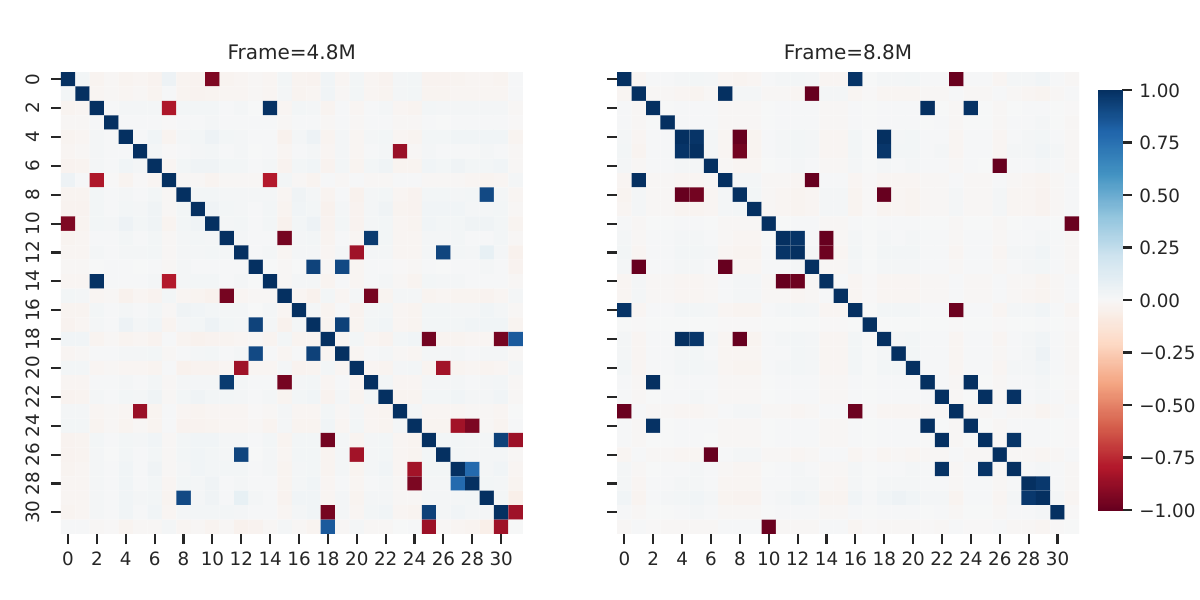}
\caption{DQN in James Bond with Adam}
\end{subfigure}
\begin{subfigure}[b]{0.49\textwidth}
\includegraphics[width=\textwidth]{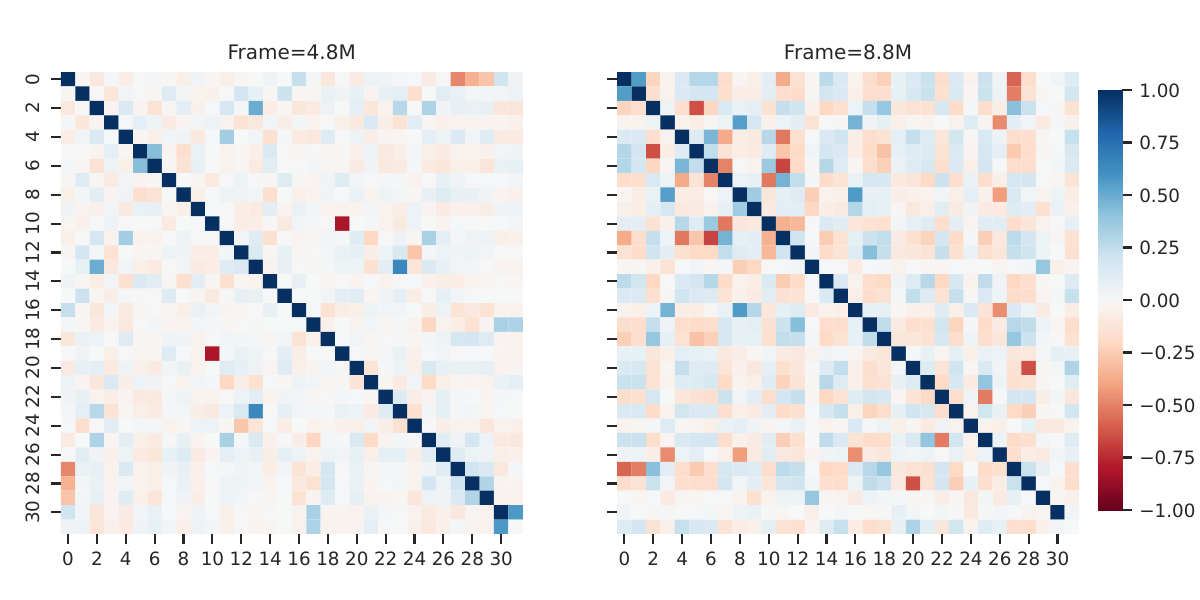}
\caption{DQN in James Bond with Adam+WC}
\end{subfigure}
\hfill
\begin{subfigure}[b]{0.49\textwidth}
\includegraphics[width=\textwidth]{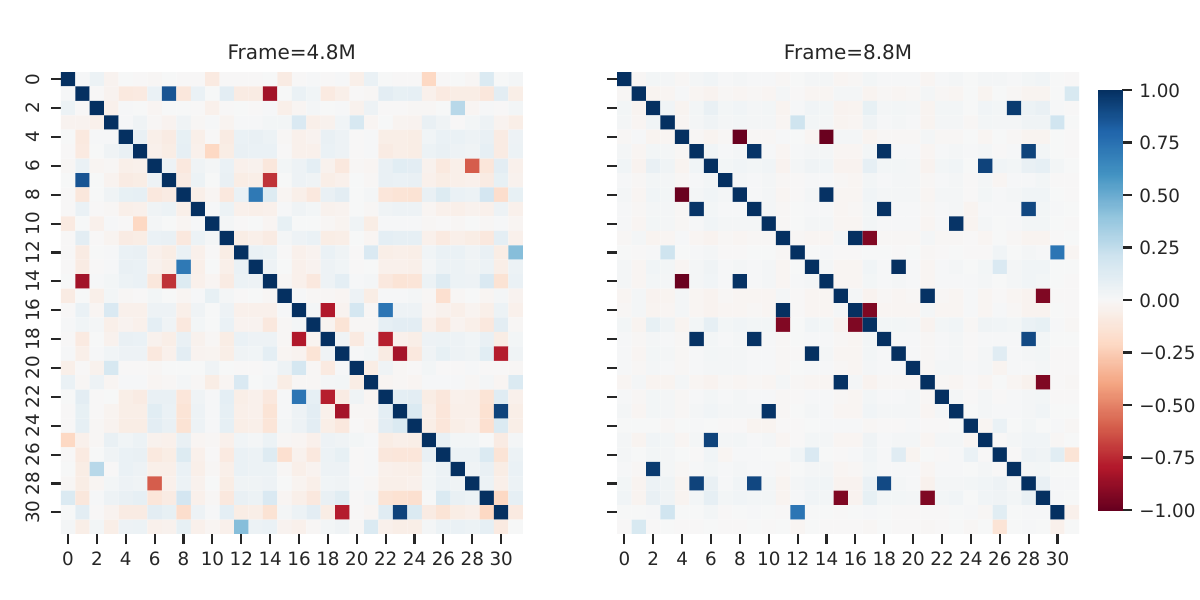}
\caption{DQN in Seaquest with Adam}
\end{subfigure}
\begin{subfigure}[b]{0.49\textwidth}
\includegraphics[width=\textwidth]{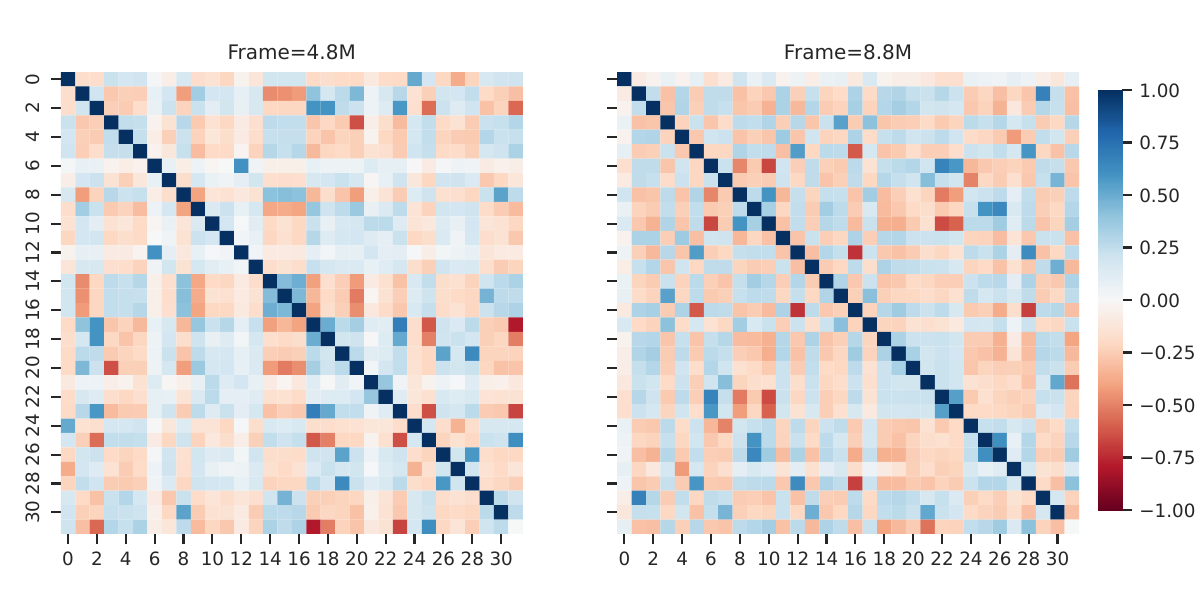}
\caption{DQN in Seaquest with Adam+WC}
\end{subfigure}
\hfill
\begin{subfigure}[b]{0.49\textwidth}
\includegraphics[width=\textwidth]{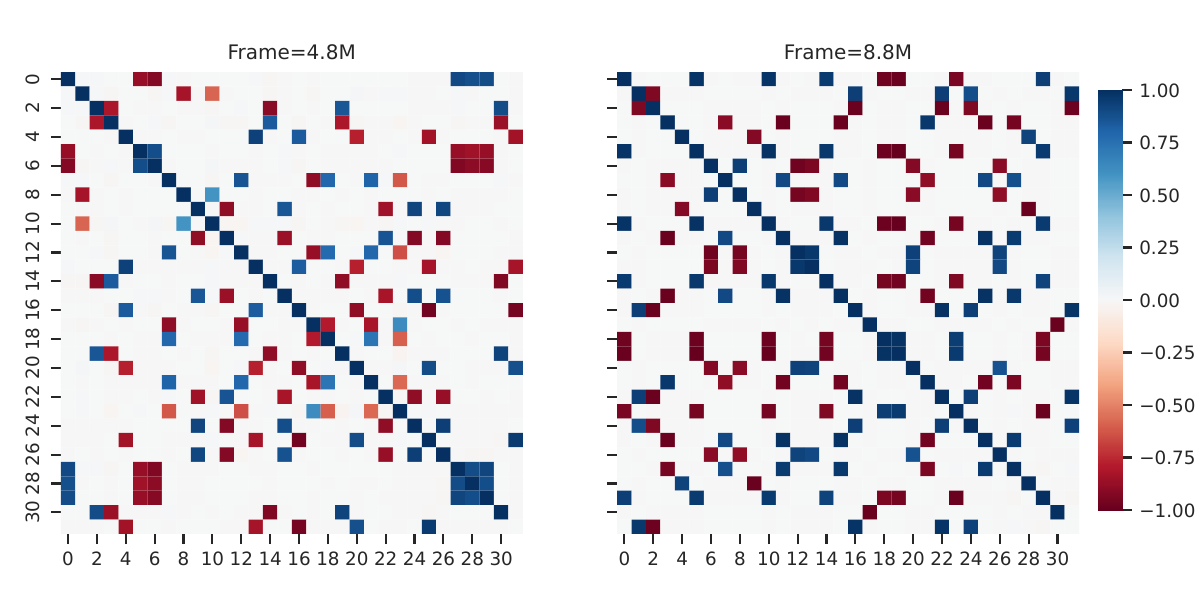}
\caption{DQN in Space Invaders with Adam}
\end{subfigure}
\begin{subfigure}[b]{0.49\textwidth}
\includegraphics[width=\textwidth]{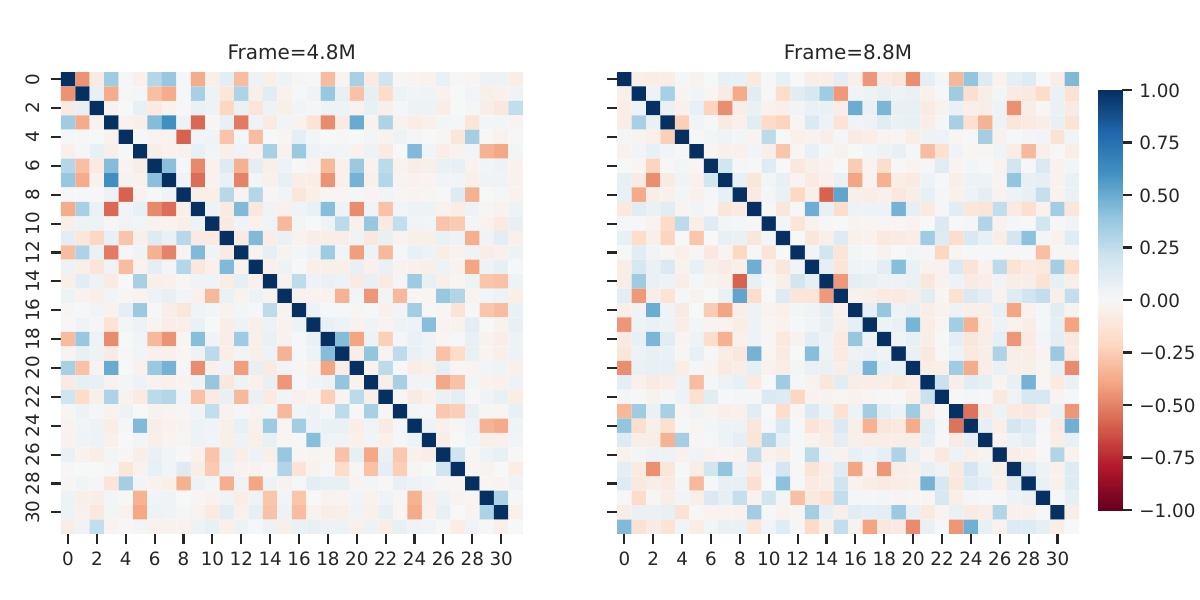}
\caption{DQN in Space Invaders with Adam+WC}
\end{subfigure}
\hfill
\begin{subfigure}[b]{0.49\textwidth}
\includegraphics[width=\textwidth]{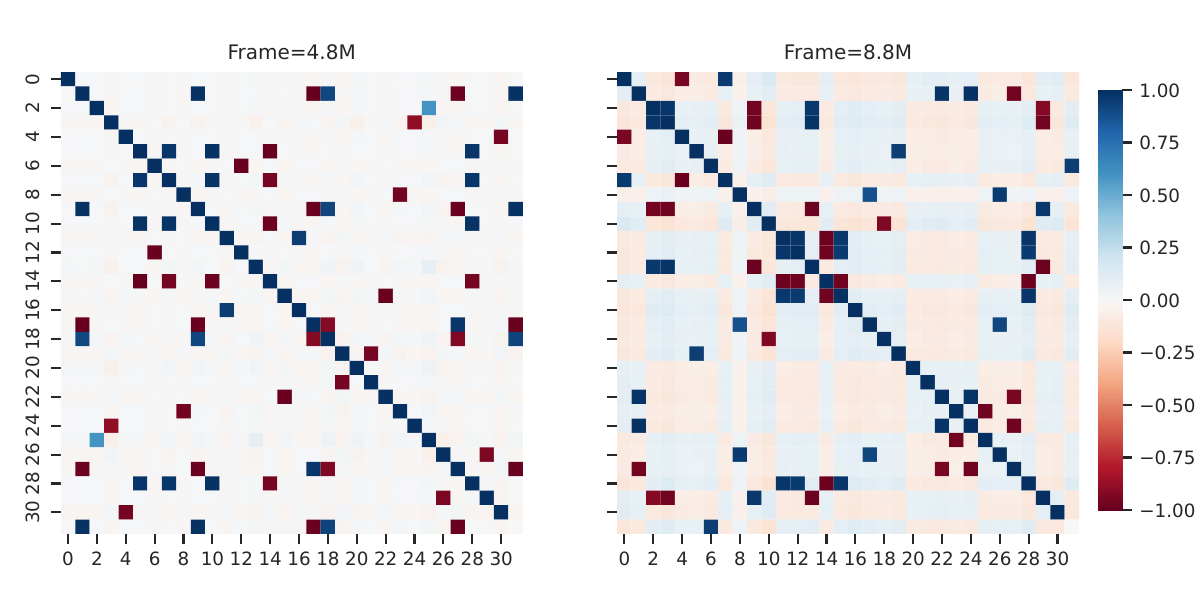}
\caption{DQN in Yars' Revenge with Adam}
\end{subfigure}
\begin{subfigure}[b]{0.49\textwidth}
\includegraphics[width=\textwidth]{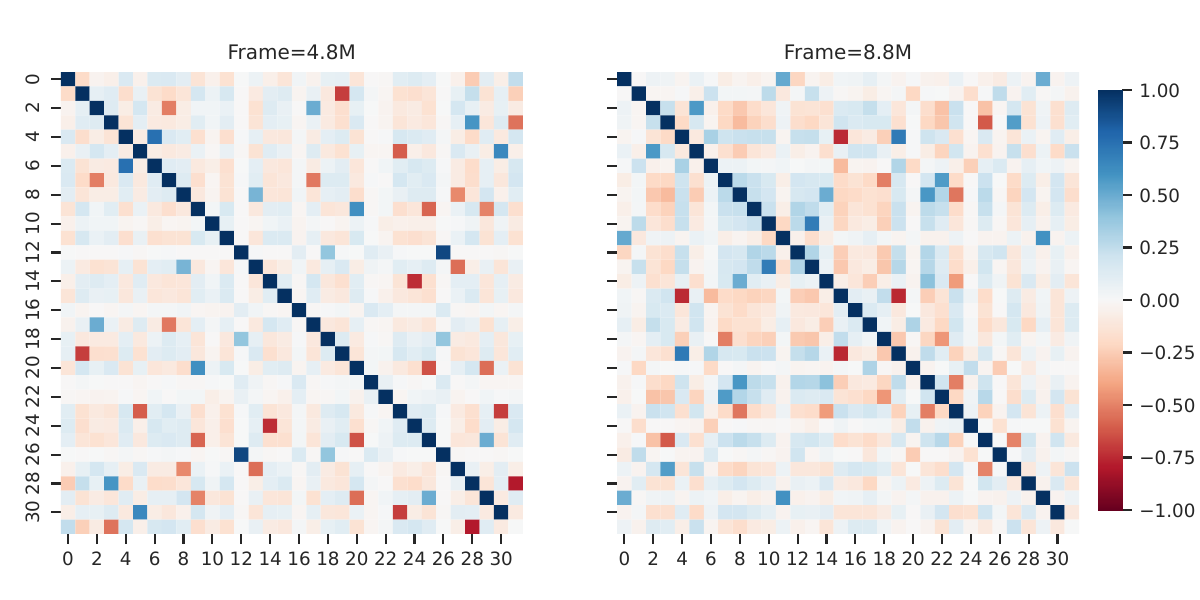}
\caption{DQN in Yars' Revenge with Adam+WC}
\end{subfigure}
\caption{The gradient covariance heatmaps of training DQN in James Bond, Seaquest, Space Invaders, and Yars' Revenge, optimized by Adam and Adam+WC, respectively.}
\label{fig:atari_grad_dqn2}
\end{figure}

\begin{figure}[htbp]
\centering
\begin{subfigure}[b]{0.49\textwidth}
\includegraphics[width=\textwidth]{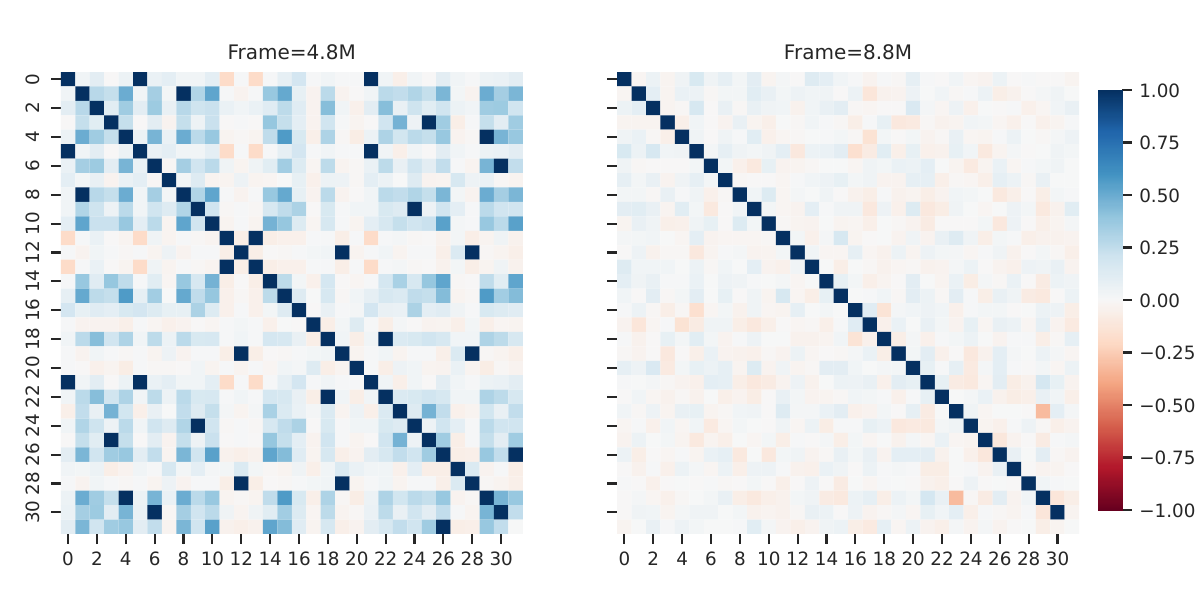}
\caption{Rainbow in Asterix with Adam}
\end{subfigure}
\begin{subfigure}[b]{0.49\textwidth}
\includegraphics[width=\textwidth]{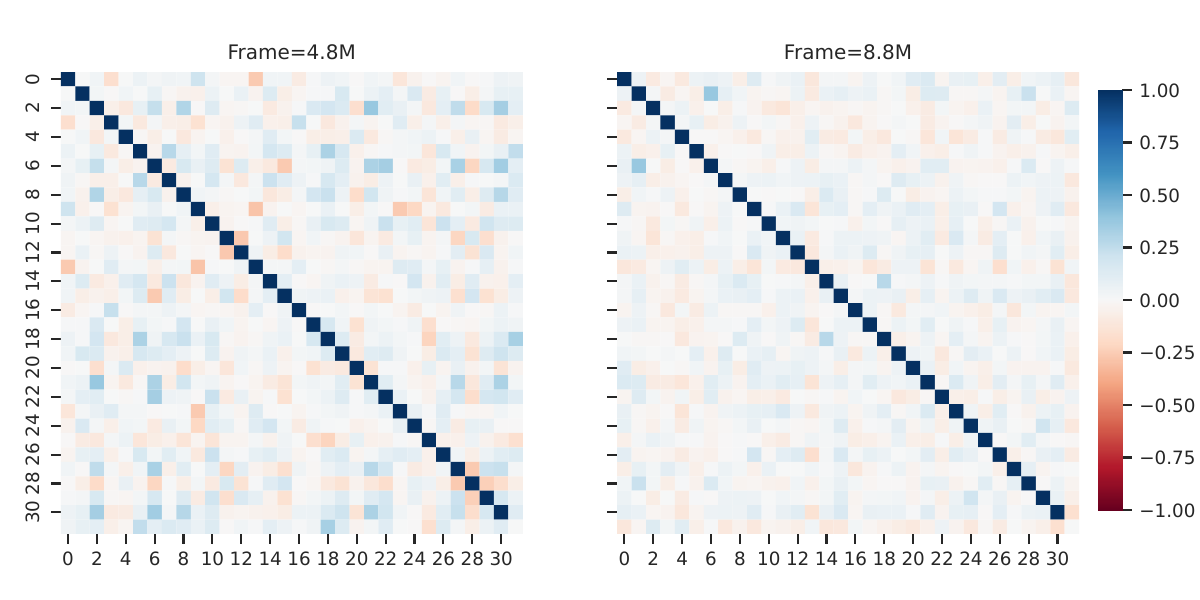}
\caption{Rainbow in Asterix with Adam+WC}
\end{subfigure}
\hfill
\begin{subfigure}[b]{0.49\textwidth}
\includegraphics[width=\textwidth]{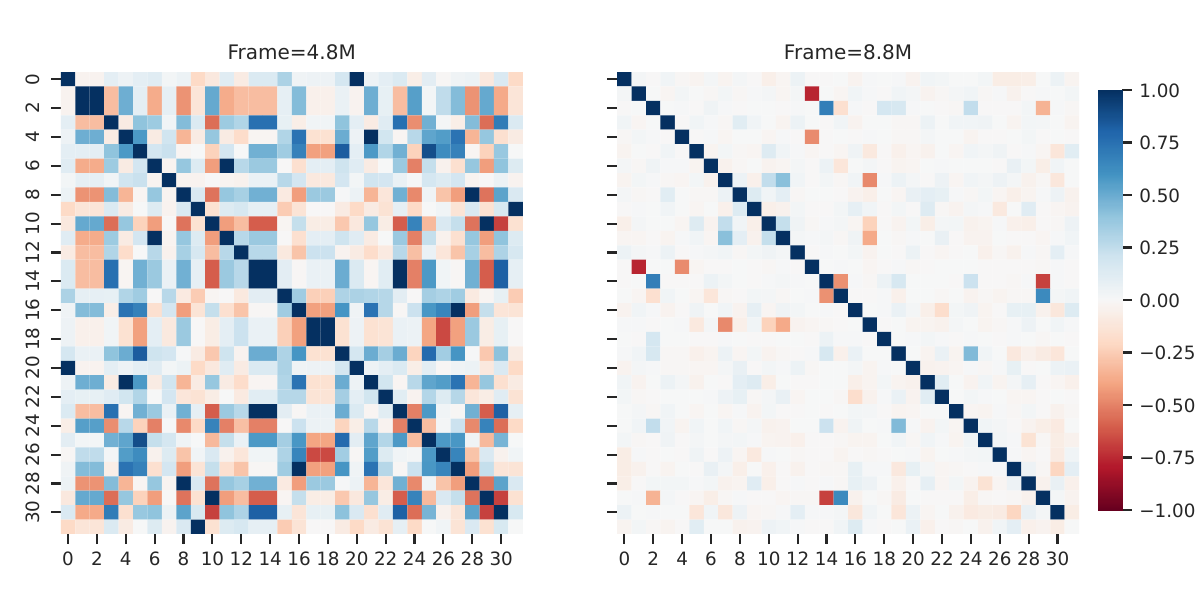}
\caption{Rainbow in Beam Rider with Adam}
\end{subfigure}
\begin{subfigure}[b]{0.49\textwidth}
\includegraphics[width=\textwidth]{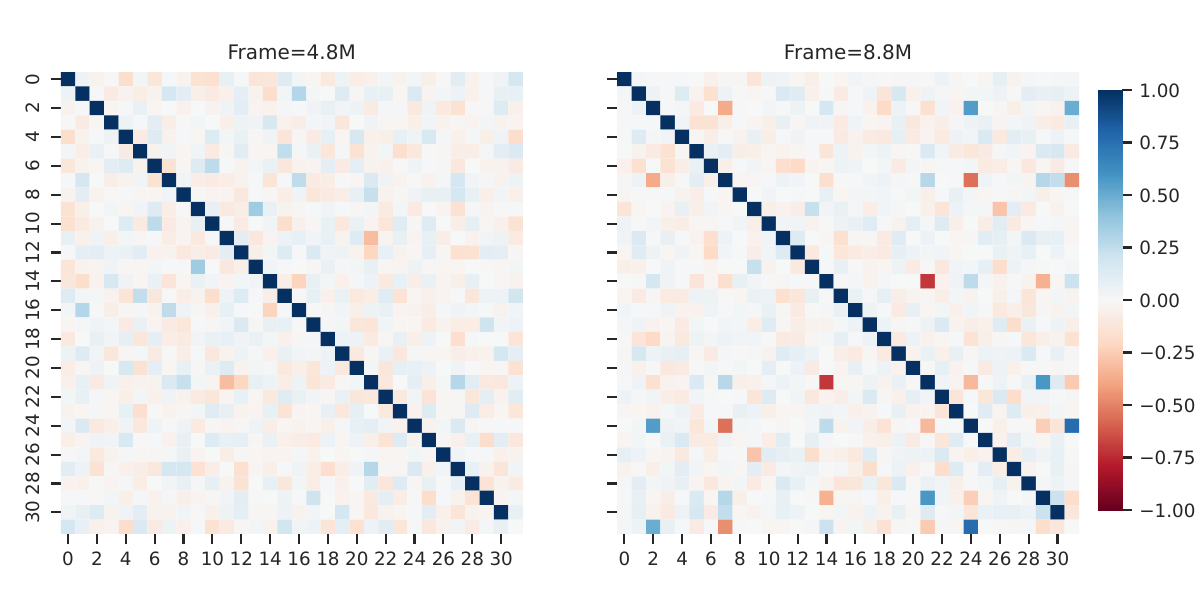}
\caption{Rainbow in Beam Rider with Adam+WC}
\end{subfigure}
\hfill
\begin{subfigure}[b]{0.49\textwidth}
\includegraphics[width=\textwidth]{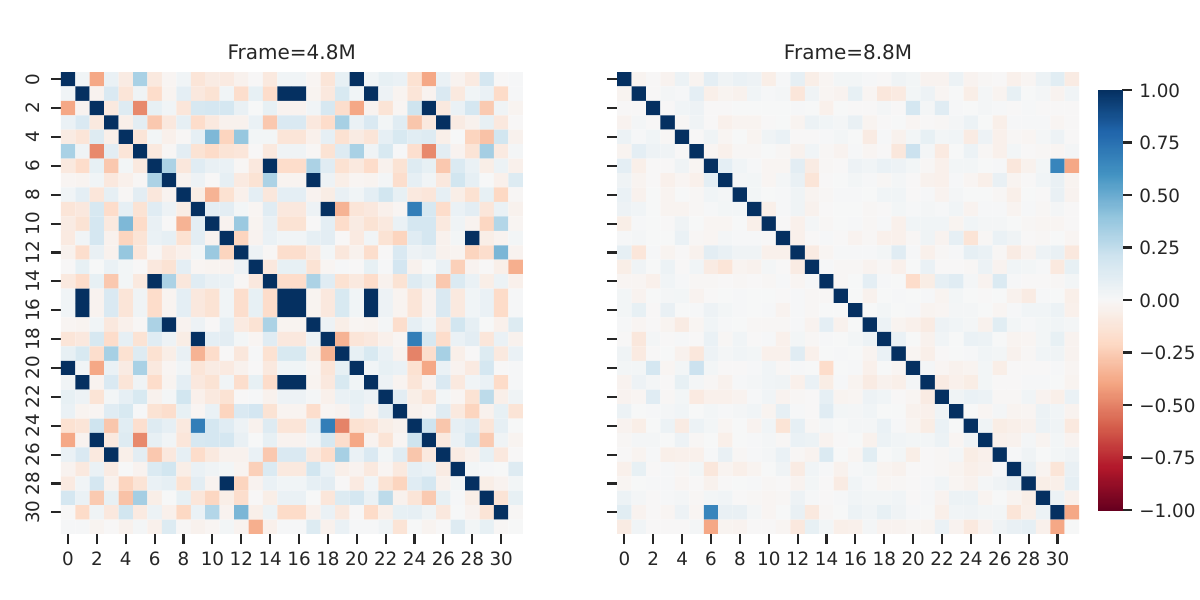}
\caption{Rainbow in Demon Attack with Adam}
\end{subfigure}
\begin{subfigure}[b]{0.49\textwidth}
\includegraphics[width=\textwidth]{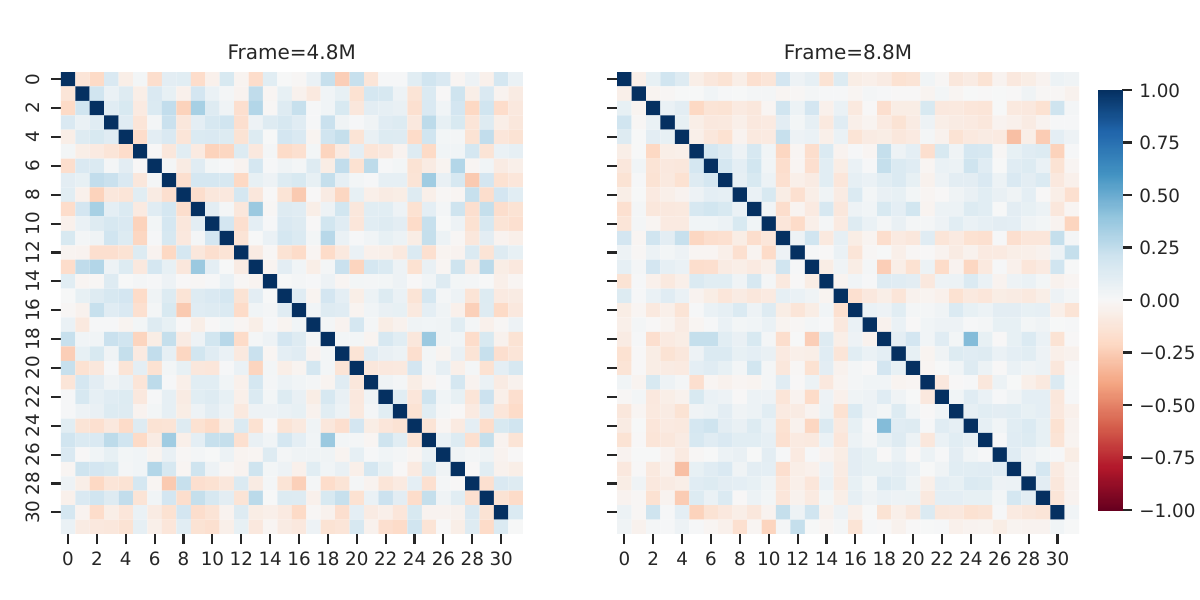}
\caption{Rainbow in Demon Attack with Adam+WC}
\end{subfigure}
\hfill
\begin{subfigure}[b]{0.49\textwidth}
\includegraphics[width=\textwidth]{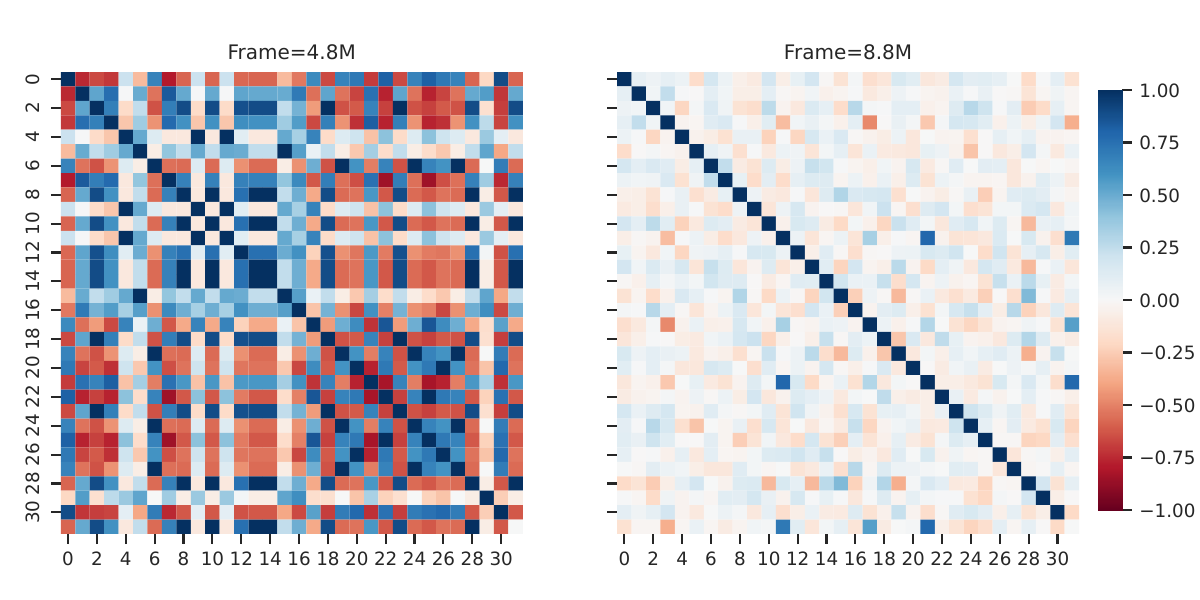}
\caption{Rainbow in Ice Hockey with Adam}
\end{subfigure}
\begin{subfigure}[b]{0.49\textwidth}
\includegraphics[width=\textwidth]{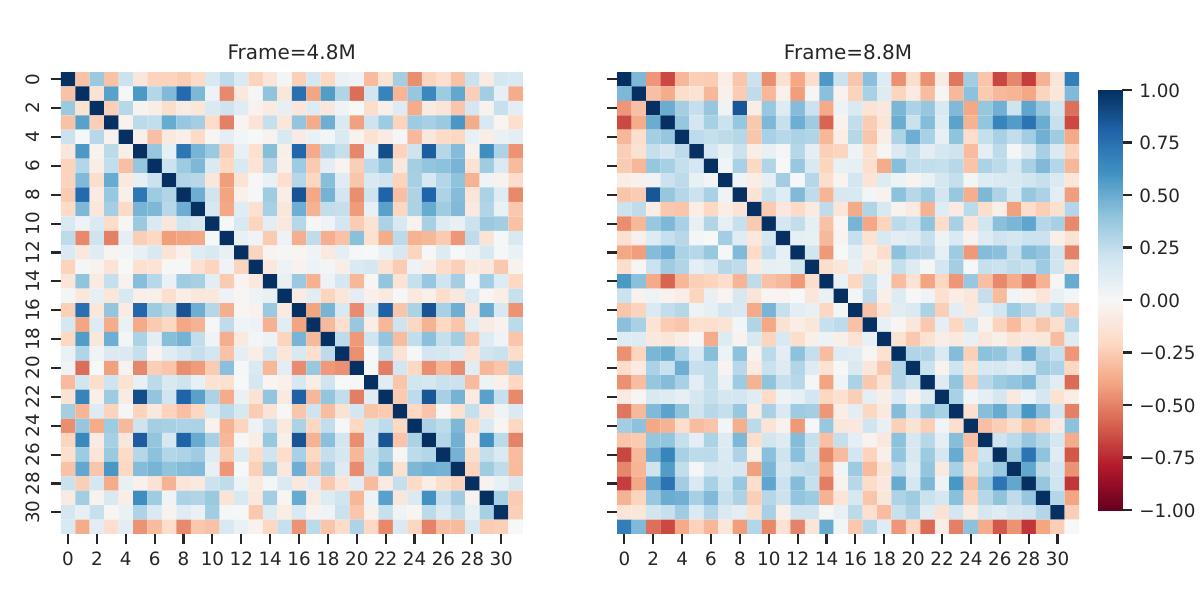}
\caption{Rainbow in Ice Hockey with Adam+WC}
\end{subfigure}
\caption{The gradient covariance heatmaps of training Rainbow in Asterix, Beam Rider, Demon Attack, and Ice Hockey, optimized by Adam and Adam+WC, respectively.}
\label{fig:atari_grad_rainbow1}
\end{figure}

\begin{figure}[htbp]
\centering
\begin{subfigure}[b]{0.49\textwidth}
\includegraphics[width=\textwidth]{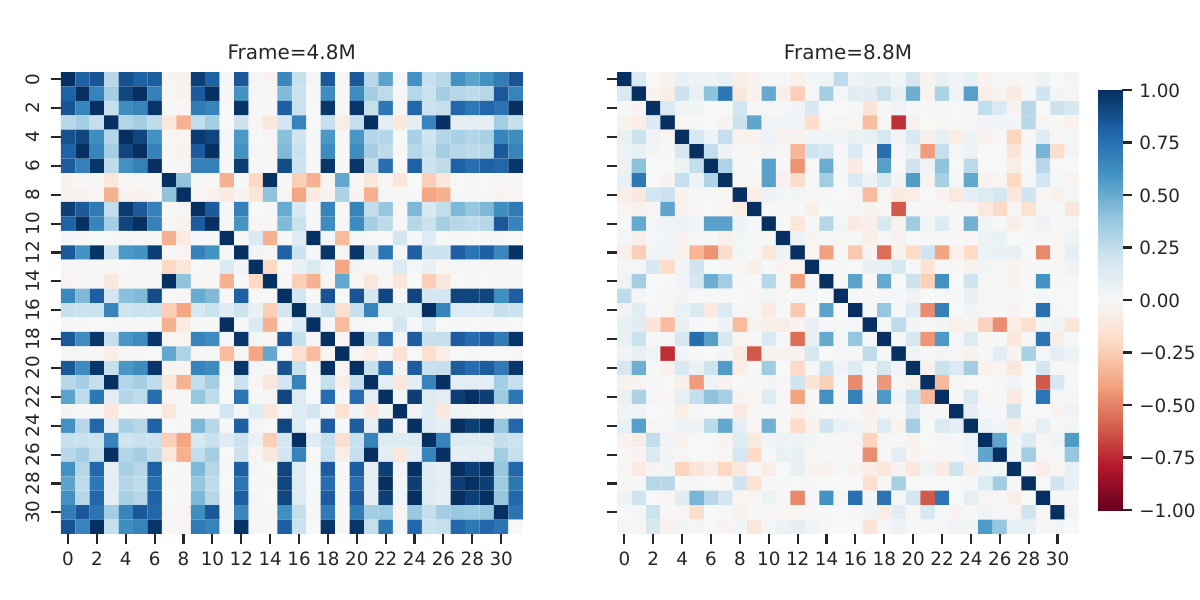}
\caption{Rainbow in James Bond with Adam}
\end{subfigure}
\begin{subfigure}[b]{0.49\textwidth}
\includegraphics[width=\textwidth]{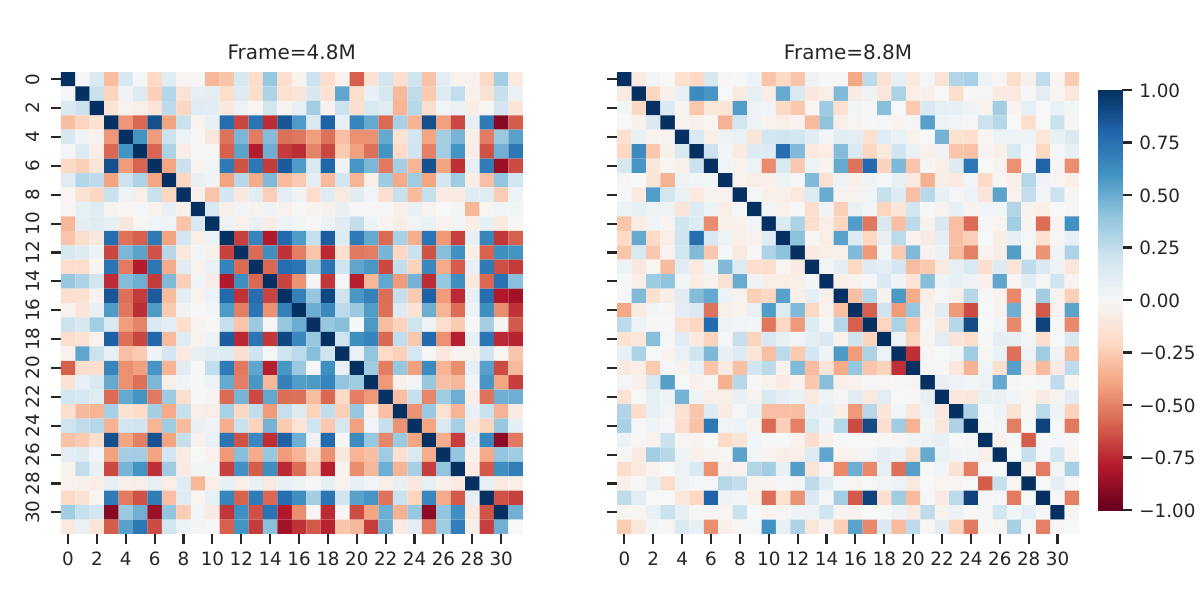}
\caption{Rainbow in James Bond with Adam+WC}
\end{subfigure}
\hfill
\begin{subfigure}[b]{0.49\textwidth}
\includegraphics[width=\textwidth]{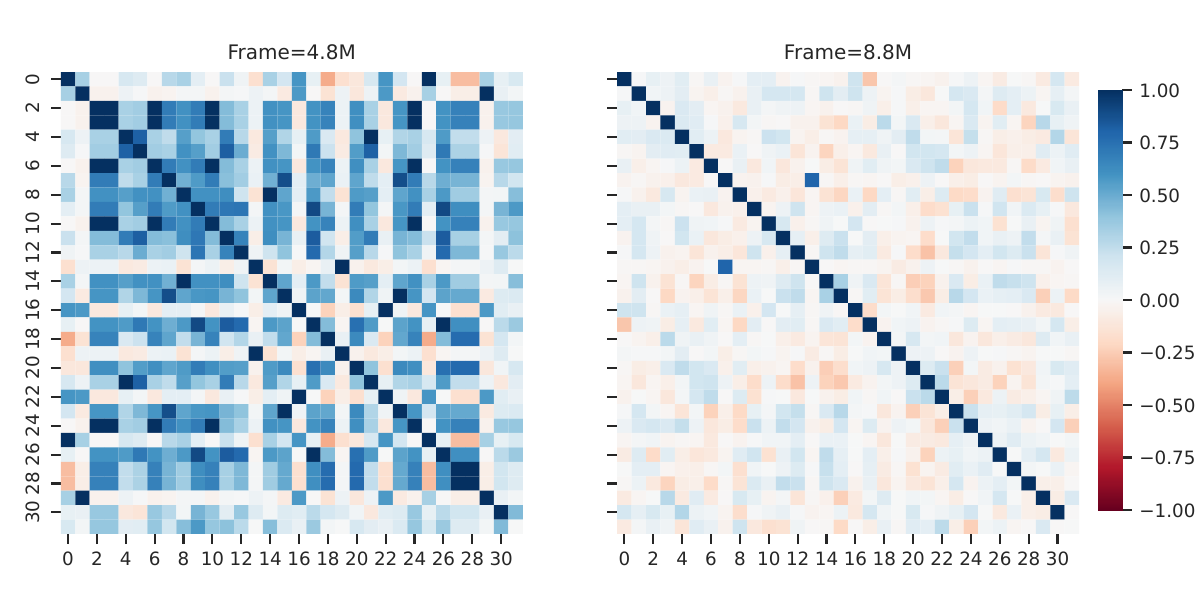}
\caption{Rainbow in Seaquest with Adam}
\end{subfigure}
\begin{subfigure}[b]{0.49\textwidth}
\includegraphics[width=\textwidth]{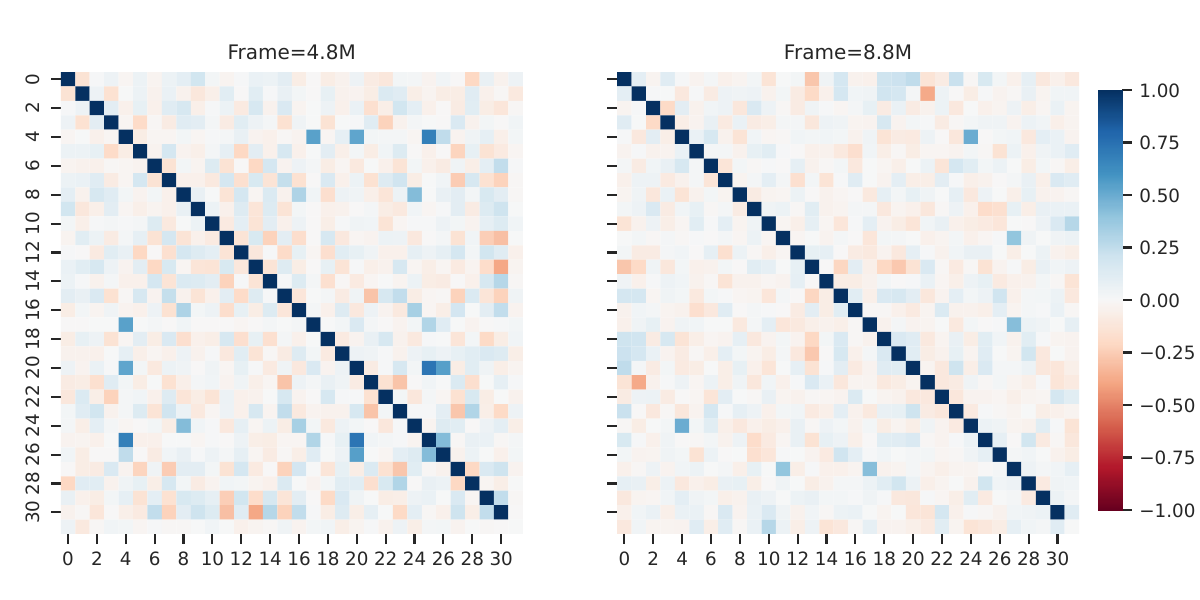}
\caption{Rainbow in Seaquest with Adam+WC}
\end{subfigure}
\hfill
\begin{subfigure}[b]{0.49\textwidth}
\includegraphics[width=\textwidth]{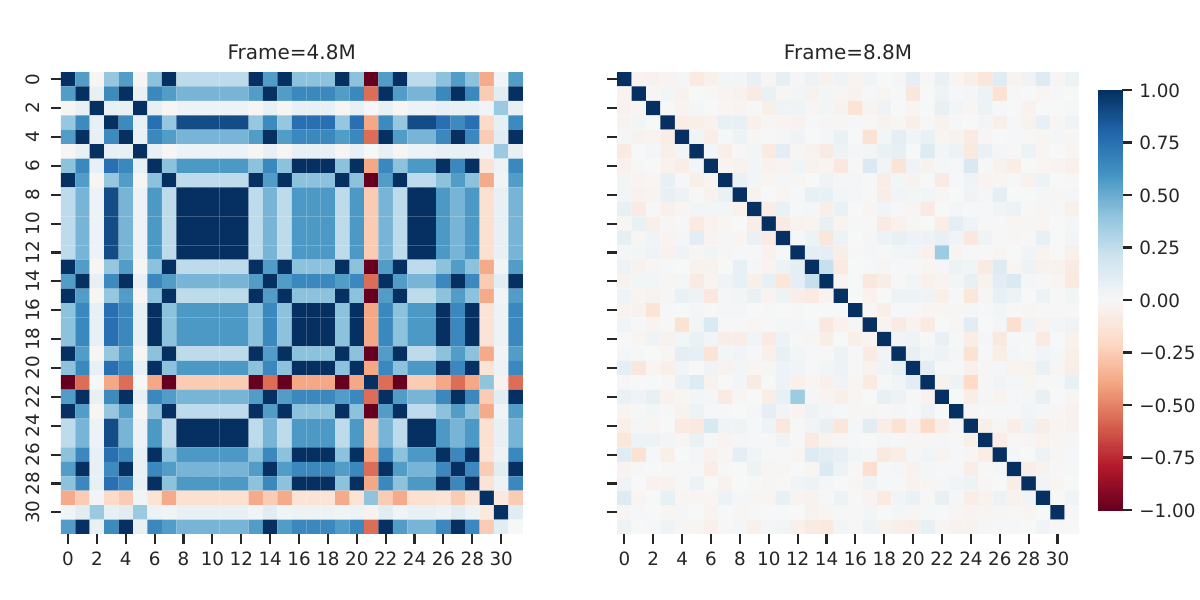}
\caption{Rainbow in Space Invaders with Adam}
\end{subfigure}
\begin{subfigure}[b]{0.49\textwidth}
\includegraphics[width=\textwidth]{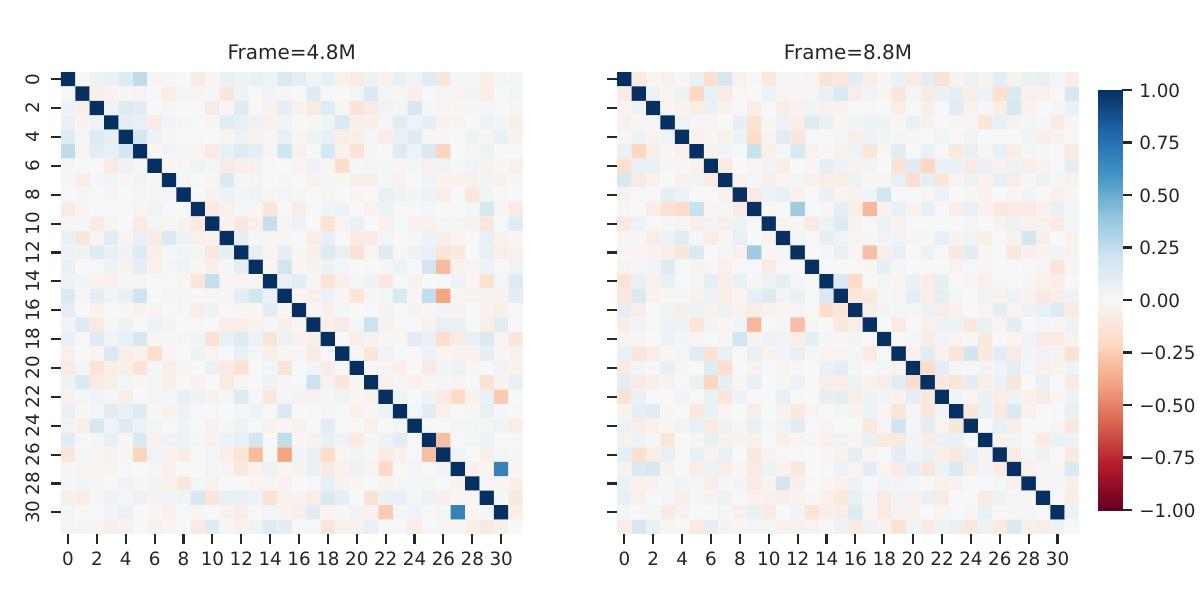}
\caption{Rainbow in Space Invaders with Adam+WC}
\end{subfigure}
\hfill
\begin{subfigure}[b]{0.49\textwidth}
\includegraphics[width=\textwidth]{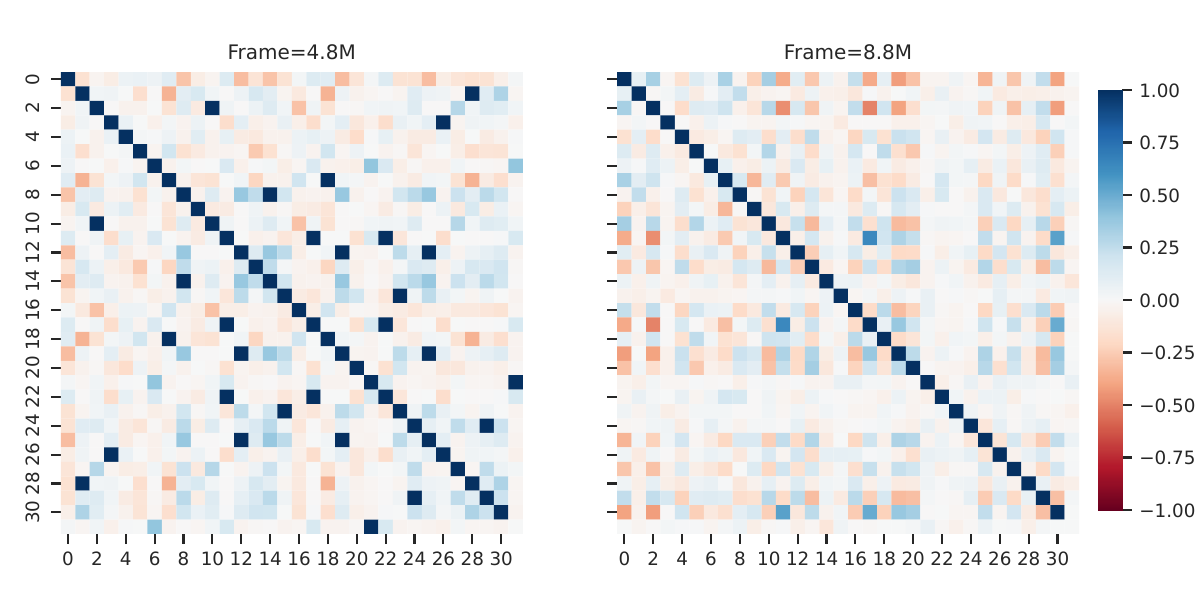}
\caption{Rainbow in Yars' Revenge with Adam}
\end{subfigure}
\begin{subfigure}[b]{0.49\textwidth}
\includegraphics[width=\textwidth]{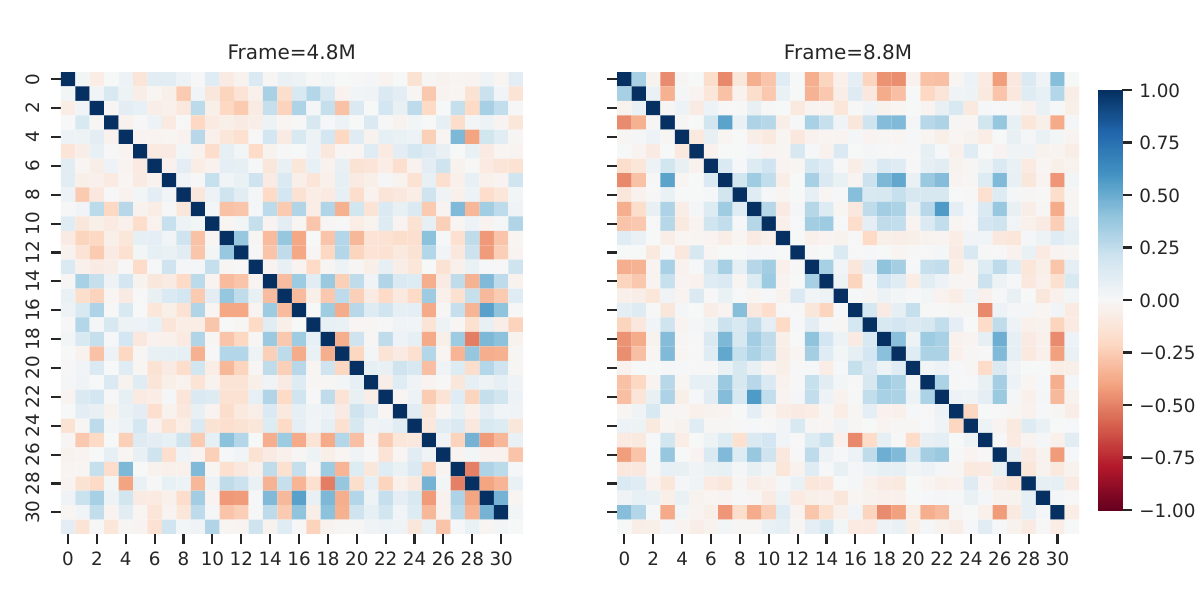}
\caption{Rainbow in Yars' Revenge with Adam+WC}
\end{subfigure}
\caption{The gradient covariance heatmaps of training Rainbow in James Bond, Seaquest, Space Invaders, and Yars' Revenge, optimized by Adam and Adam+WC, respectively.}
\label{fig:atari_grad_rainbow2}
\end{figure}

\end{document}